\documentclass[preprint,5p, times, twocolumn, authoryear]{elsarticle}

\usepackage{amssymb}

\usepackage[round]{natbib}
\usepackage{algorithm, algorithmic}

\usepackage[utf8]{inputenc} 
\usepackage[T1]{fontenc}    
\usepackage[colorlinks=true,linkcolor=blue, citecolor=blue]{hyperref}       
\usepackage{url}            
\usepackage{booktabs}       
\usepackage{amsfonts, bm, amsmath}       
\usepackage{thmtools, thm-restate}
\usepackage{amssymb}
\usepackage{nicefrac}       
\usepackage{xcolor}         

\usepackage{dsfont}
\usepackage{mathrsfs}
\usepackage{placeins}
\usepackage{mathtools}
\usepackage{float}
\usepackage{graphicx}
\usepackage{subfigure}
\usepackage{adjustbox}
\usepackage[capitalize,noabbrev]{cleveref}

\newcommand{\PR}[1]{{\mathbb{P}\left(#1 \right)}} 
 
\newcommand{\V}[1]{{\mathbb{V}\left(#1 \right)}} 
\newcommand{\N}[2]{{\mathcal{N}\left(#1, #2\right)}}
\newcommand{\round}[1]{\ensuremath{\lfloor#1\rceil}}
\newcommand{\sigmad}{\sigma^{2}_{\textnormal{classical}}}
\newcommand{\sigmat}{\sigma^{2}_{\textnormal{optim}}}
\newcommand{\sigmaht}{\hat{\sigma}_{\textnormal{optim}}^{2}}
\newcommand{\sigmahd}{\hat{\sigma}_{\textnormal{classical}}^{2}}

\newtheorem{theorem}{Theorem}[section]

\newtheorem{assumption}[theorem]{Assumption}

\usepackage{float}

\makeatletter
\def\ps@pprintTitle{%
 \let\@oddhead\@empty
 \let\@evenhead\@empty
 \def\@oddfoot{}%
 \let\@evenfoot\@oddfoot}
\makeatother

\begin{document}

\begin{frontmatter}

\title{Confident Neural Network Regression with Bootstrapped Deep Ensembles}

\author{Laurens Sluijterman}
\address{Department of Mathematics, Radboud University, \\P.O. Box 9010-59, 6500 GL, Nijmegen, Netherlands}
\ead{L.Sluijterman@math.ru.nl}
\author{Eric Cator}
\address{Department of Mathematics, Radboud University}
\ead{e.cator@science.ru.nl}
\author{Tom Heskes}
\address{Institute for Computing and Information Sciences, Radboud University}
\ead{Tom.Heskes@ru.nl}

\begin{abstract}
With the rise of the popularity and usage of neural networks, trustworthy uncertainty estimation is becoming increasingly essential. One of the most prominent uncertainty estimation methods is \textit{Deep Ensembles} \citep{lakshminarayanan2017simple}. A classical parametric model has uncertainty in the parameters due to the fact that the data on which the model is build is a random sample. A modern neural network has an additional uncertainty component since the optimization of the network is random. \citet{lakshminarayanan2017simple} noted that Deep Ensembles do not incorporate the classical uncertainty induced by the effect of finite data. In this paper, we present a computationally cheap extension of Deep Ensembles for the regression setting, called \textit{Bootstrapped Deep Ensembles}, that explicitly takes this classical effect of finite data into account using a modified version of the parametric bootstrap. We demonstrate through an experimental study that our method significantly improves upon standard Deep Ensembles.
\end{abstract}
\end{frontmatter}

\section{Introduction}
\noindent There has been an enormous interest in uncertainty quantification for machine learning in the past years. Numerous methods, discussed in the next section, have been developed. Of these methods, Deep Ensembles (DE) \citep{lakshminarayanan2017simple}, which we explain in detail in the following section, is one of the most popular. 

Ensembling methods (see, e.g., \cite{heskes1997practical} for an early example) such as DE accomplish two goals at once: The ensemble average reduces some of the variance and then provides a more accurate prediction than a random member, and the variance between the ensemble members can be used to estimate the uncertainty of the prediction. 

This uncertainty estimate can only be calibrated if the construction of the ensemble members incorporates all relevant random factors. Firstly, we have the classical source of uncertainty: The model is trained on a finite data set that can be considered a random sample drawn from an unknown distribution. We use the term \textit{classical} since for a model with a deterministic fit (such as, e.g., a linear model), the randomness of the data is the only source of variance in the parameter estimates. Secondly, the optimization procedure of neural networks is random due to random batches, initializations, and optimizers. Both the classical and optimization factor must be incorporated in order to have a calibrated uncertainty estimate.

DE are unable to do this. They can only capture the second source of uncertainty, the random optimization procedure, since all ensemble members are trained on the same data. \citet{lakshminarayanan2017simple} were aware of this problem but noted that using a standard bootstrap, where each ensemble member is trained on resampled data, actually decreased performance. \citet{nixon2020bootstrapped} ascribed this decrease in performance to effectively training on less unique data when bootstrapping.

\textit{Contribution}: In this paper, we present an efficient implementation of the parametric bootstrap for a regression setting that incorporates the missing source of uncertainty without affecting accuracy. We demonstrate that this leads to significantly better confidence intervals compared to standard DE and other popular methods. 

\textit{Scope}: We explicitly focus on a regression setting, in which we aim for more accurate confidence and prediction intervals. These are less relevant in a classification setting, where the main challenge is to properly calibrate the network outputs to probabilities. Our approach makes use of the separate estimates for the mean and variance that are given by Deep Ensembles in a regression setting, which does not translate to classification where only a single probability vector is given.

\textit{Organisation:} \cref{Background} describes the uncertainty framework that we use, gives a short overview of related work, and describes DE in more detail. This leads to \cref{Bootstrapped} where we introduce our method, \textit{Bootstrapped Deep Ensembles}. In \cref{Results}, we experimentally demonstrate the significance of the effect of finite data and show that incorporating this improves the confidence intervals significantly, which in turn results in improved prediction intervals.  Finally, \cref{Conclusion} summarises the conclusions and gives possible avenues for future work. 

\section{Background} \label{Background}
\begin{figure*}[tb]
\centering
\includegraphics[width=0.9\textwidth]{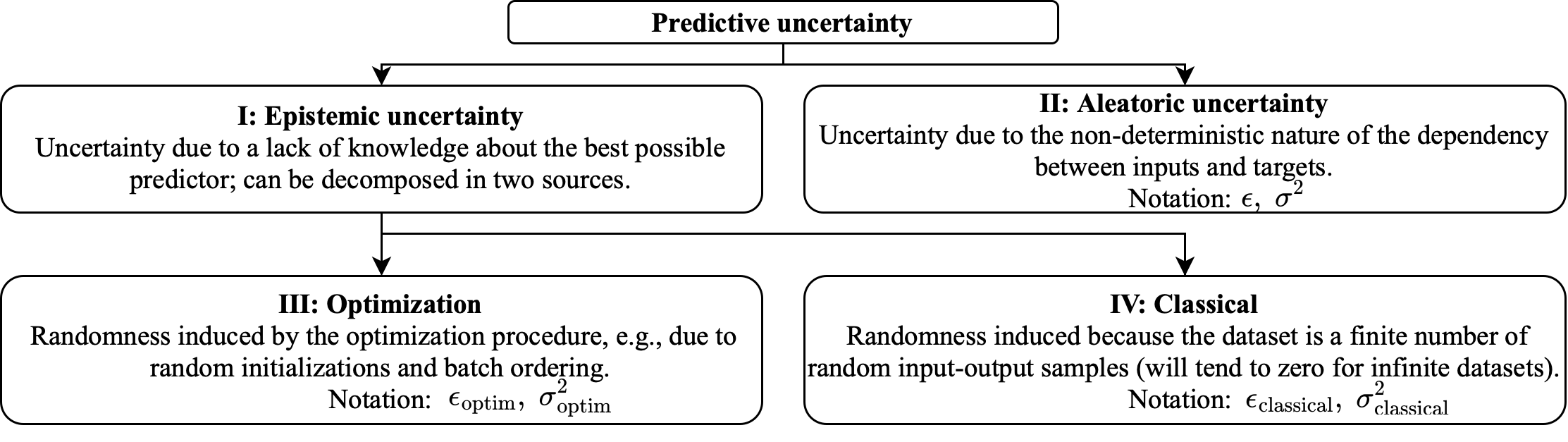}
\caption{The different types of uncertainty.}
\label{uncertaintyfigure}
\end{figure*}
\subsection{Uncertainty framework}
\noindent We consider a frequentist regression setting in which a neural network is trained on a data set $\mathcal{D}$ consisting of $N$ i.i.d.\ observation pairs $(\bm{x}, y)$, with input $\bm{x} \in \mathbb{R}^{d}$ and target $y \in \mathbb{R}$. Given a new input $\bm{x}^{*}$, the neural network outputs a prediction $\hat{f}(\bm{x})$ for the corresponding target $y^{*}$. 

 The uncertainty about this prediction consists of two main components. Firstly, we are unsure about the true underlying function $f(\bm{x})$. This may be the result of a limited amount of, possibly noisy, data or due to randomness in our training procedure, such as random initialisation and batches. This is often referred to as \textit{epistemic uncertainty} \citep{hullermeier2019aleatoric, abdar2021review}. Secondly, even if we were absolutely sure about $f(\bm{x})$, we would still have the uncertainty in our prediction of $y$ as a result of the intrinsic noise, the so-called \textit{aleatoric uncertainty}. Whereas we may be able to reduce the epistemic uncertainty, e.g., by gathering more data or changing our optimization procedures, it is impossible to reduce the aleatoric uncertainty. This distinction is summarised in Figure \ref{uncertaintyfigure}. 

The certainty in our predictions can be expressed via a confidence and prediction interval. Depending on the task, either the confidence or prediction interval may be of more interest. A confidence interval gives the region that is expected to cover the true function value, $f(\bm{x}^{*})$. The prediction interval gives the region that is expected to cover a new observation, $y^{*}$. The confidence interval is determined only by the epistemic uncertainty, whereas the prediction interval also depends on the aleatoric uncertainty. 

In this paper, we adhere to the frequentist interpretation of confidence and prediction intervals. A $(1-\alpha)\%$ confidence interval for $f(\bm{x}^{\ast})$ is a random mapping from $\bm{x}^{\ast}$ to an interval such that if we repeated the entire experiment infinitely many times - that means sampling data, training the network, creating the interval - we would capture the true function value $f(\bm{x}^{\ast})$ in $(1-\alpha)\%$ of the experiments. A confidence interval with a coverage higher than $(1-\alpha)\%$ is called \textit{conservative}. Additionally, if this holds for all values of $\bm{x}$, it is called a \textit{pointwise} conservative confidence interval. A prediction interval is defined similarly but with $y$ instead of $f(\bm{x})$. Bayesian methods typically output a credible interval. Although credible regions have fundamental differences, it is desirable for a credible interval to maintain frequentist properties and it is common to evaluate Bayesian methods frequentistically \citep{ghosal2017fundamentals}.

Throughout this paper, we take the fixed-covariates perspective, meaning that we treat $\bm{x}$ as being fixed and given and $y$ as the realisation of a random variable. The classical uncertainty due to finite data is therefore the uncertainty due to the randomness of the targets. To make this perspective explicit, we will use the term \textit{random targets} instead of \textit{finite data}.

\subsection{Related work and Deep Ensembles}
\noindent Numerous different methods to obtain uncertainty estimates for neural networks have been developed. We refer the reader to \cite{abdar2021review} for an extensive overview and list a few notable contributions here. \textbf{Bayesian Neural Networks} \citep{mackay1992practical, neal2012bayesian} put a prior distribution over the weights of a network and use the posterior to obtain uncertainty estimates. The calculation of the posterior is often intractable. \textbf{Variational Inference} \citep{hinton1993keeping, jordan1999introduction} aims to solve this problem by using a tractable approximation of the posterior. \textbf{Monte-Carlo Dropout} \citep{gal2016dropout, gal2017concrete} is a notable example of variational inference. Since dropout is already used in many neural networks as a regularization technique, it comes at no extra cost at training time. A downside is that the epistemic uncertainty is only influenced by the dropout rate, thus making it impossible to locally tune the uncertainty estimates to have the correct size \citep{sluijterman2021evaluate, osband2016risk}.
 \textbf{Quantile regression} \citep{cannon2011quantile, xu2017composite, clements2019estimating, tagasovska2019single} uses a pinball loss to output quantiles directly without the need of any distributional assumptions. Similarly, direct Prediction Interval (PI) methods use a custom loss function that directly outputs PIs with the goal to capture the correct fraction of data points while being as narrow as possible \citep{pearce2018high}. 
Other methods are focussed more on detecting \textbf{out-of-distribution} (OoD) samples. These are input values that are very different from the training data. A typical approach is to keep track of the pre-activations, the output of the penultimate layer (or sometimes also other layers) times the weight matrix plus the bias vector, and use some distance measure to determine the level of difference of a new datapoint \citep{van2021improving, mukhoti2021deterministic, lee2018simple}. Similarly, \cite{ren2019likelihood} use likelihood ratios directly on the inputs to detect out-of-distribution samples. OoD detection methods do not aim for calibrated prediction or confidence intervals.

\textbf{Deep Ensembles} train an ensemble of $M$ networks, each member receiving the same data but in a different order and having a different initialisation. The architecture of the networks is similar to the mean-variance estimation method by \citet{nix1994estimating}, where each network outputs a mean, $\hat{f}_{i}(\bm{x})$, and variance prediction, $\hat{\sigma}_{i}^{2}(\bm{x})$, for every input. The variance terms $\hat{\sigma}_{i}^{2}(\bm{x})$ estimate the aleatoric uncertainty in box II of \cref{uncertaintyfigure}. The networks are trained by minimizing the negative loglikelihood of a normal distribution, which implies the following assumption.

\begin{restatable}{assumption}{mainA}
\label{assumption1}
 	The targets, $y$, are the sum of a function value $f(\bm{x})$ and normally distributed heteroscedastic noise: 
 	\[
 	y = f(\bm{x}) + \epsilon, 
 	\quad \text{with } \epsilon \sim \N{0}{\sigma^{2}(\bm{x})}.
 	\]
\end{restatable}
Deep Ensembles assume a Gaussian Mixture of the individual models as the predictive model. In this model, the total mean and variance are defined as
\[
\hat{f}_{*}(\bm{x}) = {1 \over M}\sum_{i=1}^{M} \hat{f}_{i}(\bm{x}), \quad \text{and}
\]
\begin{equation} 
\label{sigma}
 \quad \hat{\sigma}^{2}_{*}(\bm{x}) = {1 \over M} \sum_{i=1}^{M} \left(\hat{f}_{i}(\bm{x})^{2} - \hat{f}_{*}(\bm{x})^{2} + \hat{\sigma}_{i}^{2}(\bm{x})\right).
\end{equation}
This results in the prediction interval
\[
\text{PI}_{\text{DE}} =  \hat{f}_{*}(\bm{x}) \pm z_{\alpha / 2}\sqrt{\hat{\sigma}^{2}_{*}(\bm{x})},
\]
where $z_{\alpha / 2}$ is the $\alpha / 2$ quantile of a standard normal distribution. For later comparison, we construct a confidence interval implied by DE by ignoring the aleatoric variance terms $\hat{\sigma}^2_i(x)$ in Equation \eqref{sigma} to arrive at
\[
\text{CI}_{\text{DE}}  = \hat{f}_{*}(\bm{x}) \pm t^{(M-1)}_{\alpha / 2} \sqrt{{1 \over M} \sum_{i=1}^{M} \hat{f}_{i}(\bm{x})^{2} - \hat{f}_{*}(\bm{x})^{2}},
\]
where $t^{(M-1)}_{\alpha / 2}$ is the $\alpha / 2$ quantile of a $t$ distribution with $M-1$ degrees of freedom. 

Deep Ensembles have been shown to clearly outperform Variational Inference and Monte-Carlo Dropout \citep{lakshminarayanan2017simple} and are regarded the state of the art for uncertainty estimation, both in-distribution \citep{ashukha2020pitfalls} and under distributional shift \citep{ovadia2019can}. 

Different explanations for the success of Deep Ensembles have been given. \citet{wilson2020bayesian} relate the method to a form of Bayesian model averaging. They empirically demonstrate that DE are even able to better approximate the predictive distribution than some standard Bayesian approaches. Similarly, \citet{gustafsson2020evaluating} relate the method to sampling from an approximate posterior. Alternatively, \citet{fort2019deep} explain the success via the loss landscape. They argue that the different models are able to explore different local minima, where a Bayesian approximation may only explore a single local minimum.

None of these interpretations fully explains why the obtained intervals would be properly calibrated. In fact, as we will also show in our experiments, by training each ensemble member on the same data, DE ignore a significant part of the epistemic uncertainty that is due to finite data (\cref{uncertaintyfigure} box IV). In the next section, we introduce our method, \textit{Bootstrapped Deep Ensembles}, an easy to implement extension of DE with comparable computational costs that does take this source of epistemic uncertainty into account. 
\section{Bootstrapped Deep Ensembles} \label{Bootstrapped}
\noindent Our method can be summarised in two steps. We first train a regular Deep Ensemble, resulting in the exact same predictor $\hat{f}_{*}(\bm{x})$ and thus identical accuracy. Secondly, we repeat a small part of the training of these members on new data, more on this shortly, in order to capture the uncertainty that we missed by training the ensemble members on identical data.

As previously stated, we model a neural network as a random predictor with an error that decomposes in a part due to the optimization procedure and a part due to random targets. We formalize this in the following assumption. 

\begin{restatable}{assumption}{mainB}
\label{assumption2}
Let $\hat{f}_{i}(\bm{x})$ be the prediction of an ensemble member trained on the same data set $\mathcal{D}$, but with a unique initialization and data ordering, and let $f(\bm{x})$ be the true value, then
\[
\hat{f}_{i}(\bm{x}) = f(\bm{x}) + \epsilon_{\textnormal{classical}} + \epsilon_{\textnormal{optim},i}, \quad \text{with}\] 
\[ 	\epsilon_{\textnormal{classical}} \sim \N{0}{\sigmad(\bm{x})}\;  \text{and} \;\epsilon_{\textnormal{optim},i} \sim \N{0}{\sigmat(\bm{x})},\]
where all $\epsilon$ are independent. The $\epsilon_{\text{classical}}$ term does not have an index $i$ since by definition it is the same for all ensemble members. 
\end{restatable}
The $\epsilon_{\text{classical}}$ and $\sigmad(\bm{x})$ terms relate to the uncertainty in box IV of \cref{uncertaintyfigure} and the terms $\epsilon_{\text{optim}, i}$ and $\sigmat(\bm{x})$ to box III. From \cref{assumption2}, it follows that the average of ensemble members trained on the same data, $f_{*}(\bm{x})$, has variance
\begin{equation} \label{varianceeq}
\mathbb{V}(\hat{f}_{*}(\bm{x})) = \sigmad(\bm{x}) + \frac{\sigmat(\bm{x})}{M}.
\end{equation}
The $\sigmad(\bm{x})$ term does not get divided by $M$ since all ensemble members are trained on the same data. 

One way to estimate the total variance (\cref{varianceeq}) is to train multiple (deep) ensembles. However, this would become extremely expensive. The contribution of our paper is that, with only a modest amount of extra work, we are able estimate the total variance. The key ingredient of our approach is to \textit{separately} estimate the two terms  $\sigmad(\bm{x})$ and $\sigmat(\bm{x})$.

\subsection{Incorporating the missing uncertainty} 
\noindent The estimate for $\sigmat(\bm{x})$ is straightforward. Since the $M$ ensemble members are trained on the same data set, we can take the sample variance of these ensemble members as an estimate of the variance due to the random optimization:
\[
\sigmaht(\bm{x}) = \frac{1}{M-1} \sum_{i=1}^{M}\left(\hat{f}_{i}(\bm{x}) - \frac{1}{M}\sum_{i=1}^{M} \hat{f}_{i}(\bm{x}) \right)^{2}.
\]

Deep Ensembles, however, fail to measure $\sigmad(\bm{x})$. To estimate the missing $\sigmad(\bm{x})$, we propose to use an adapted version of the \textit{parametric bootstrap} \citep{efron1982jackknife}. For a standard parametric model, the parametric bootstrap consists of two steps. A single model is trained on the data, after which $B$ additional models are trained using new data, simulated from the first model. The variance of those extra models is then used to obtain the model uncertainty.

Directly translating the parametric bootstrap to our setup does not work. Training a new network on simulated targets would also capture the variance due to the optimization procedure. As indicated before, a solution could be to estimate the entire variance in Equation (\ref{varianceeq}) directly by training $B$ entire ensembles on simulated data sets, but this would be far too expensive. We therefore propose an approach to \textit{train additional neural networks while eliminating optimization variability.}

\begin{figure}[tb]
\centering
\subfigure[Effect of random optimization]{\includegraphics[width=0.4\textwidth]{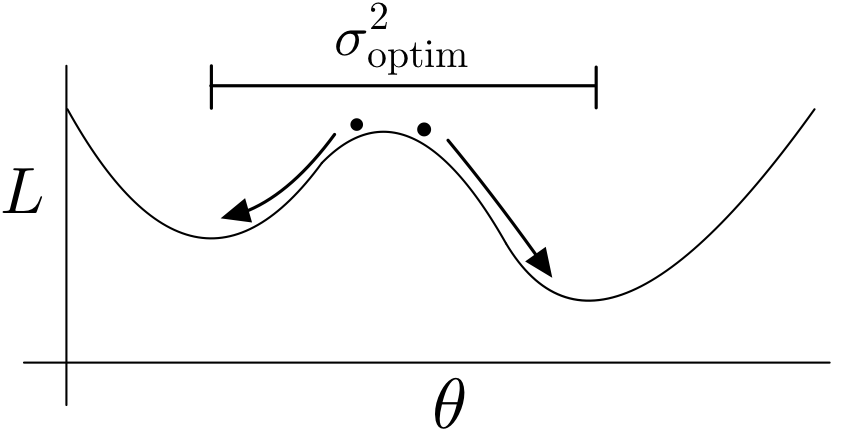}} \hskip 0.25in
\subfigure[Effect of random targets]{\includegraphics[width=0.4\textwidth]{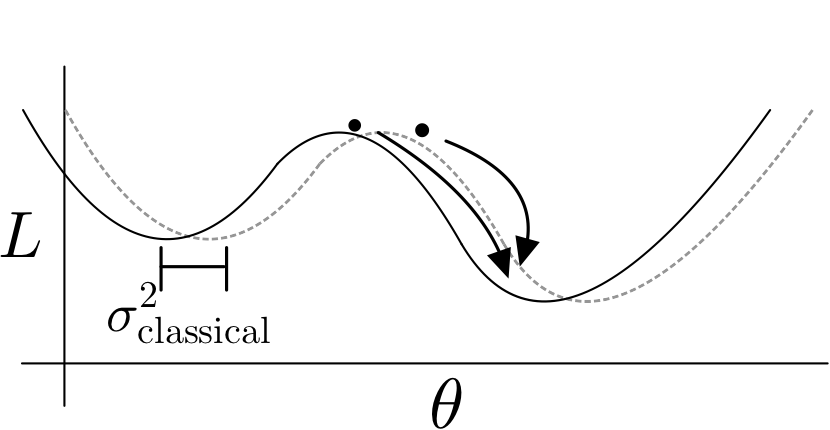}}
\vskip -0.1in
\caption{Figure (a) sketches the effect of the random optimization procedure. Through random initializations and random orderings of the data, different regions of the loss landscape get explored. We denote the variance that arises from this effect with $\sigmat$. With finite training data, our loss landscape itself is subject to randomness. To estimate the resulting uncertainty, we apply the parametric bootstrap, resulting in new targets and a slightly deformed loss landscape. To ensure we end up in the deformed version of the same local minimum, we repeat only a part of the training.}
\label{fig: losslandscape}

\end{figure}

To explain how we do this, we examine the problem from a loss landscape perspective, as sketched in \cref{fig: losslandscape}. The random optimization causes the networks to end up in different local minima, while different targets cause the loss landscape to deform. Starting from a later point in the training cycle - as opposed to starting at initialisation - is much more likely to cause the retrained network to end up in the deformed version of the \textit{same} local minimum, thus eliminating optimization variability. 

In order to estimate $\sigmad(\bm{x})$, we therefore propose the following procedure. During the training of the original ensemble members, we save a copy of the state of the network after $\round{N_{\textnormal{epoch}}\cdot (1-r)}$ epochs, where $N_{\textnormal{epoch}}$ is total amount of training epochs and $r \in [0, 1]$ is the retraining fraction. We then repeat the final $r\cdot N_{\textnormal{epoch}}$ epochs, starting from the saved state, with new targets that are simulated from a  $\mathcal{N}(\hat{f}_{i}(\bm{x}),\hat{\sigma}_{i}^{2}(\bm{x}))$ distribution. We denote this retrained network with $\hat{\hat{f}}_{i}(\bm{x})$.

This retraining is meant to capture solely the variance due to the random targets. A standard assumption of the parametric bootstrap is that the distributions of the difference of the first model and the true model, and the difference of the retrained model and the first model, are similar \citep{efron1982jackknife}. In our case this assumption translates to:
\begin{assumption}
\label{assumption3}
Let $\hat{\hat{f}}_{i}(\bm{x})$ denote the predictions of a retrained ensemble member. The difference between $\hat{f}_{i}(\bm{x})$ and $\hat{\hat{f}}_{i}(\bm{x})$ is normally distributed with zero mean and variance $\sigmad(\bm{x})$: 
 	\[
 	\resizebox{0.48\textwidth}{!}{$\hat{\hat{f}}_{i}(\bm{x}) = \hat{f}_{i}(\bm{x}) + \epsilon_{\textnormal{classical},i} \; \text{with} \; \epsilon_{\textnormal{classical}, i} \sim  \N{0}{\sigmad(\bm{x})}$}.	
 	\]
\end{assumption}
As an estimate for $\sigmad(\bm{x})$, we therefore use
\[
\sigmahd(\bm{x}) = \frac{1}{M}\sum_{i=1}^{M}\left(\hat{f}_{i}(\bm{x}) - \hat{\hat{f}}_{i}(\bm{x})\right)^{2}.
\]

In total, we have made three assumptions. Assumption \ref{assumption1} is a modeling assumption that states that we are dealing with additive Gaussian heteroscedastic noise. This assumption is very standard and made by most works on uncertainty estimation (e.g. Deep Ensembles and Concrete Dropout). Assumption \ref{assumption2} states that we assume the model uncertainty to be normally distributed and to consist of a classical part and an optimization part. This normality is also a very standard assumption. The typical reasoning behind this assumption is that it holds asymptotically for a parametric model and is therefore the most sensible choice, also for finite data and non-parametric models (see \ref{assumptionsjustification} for more details). The same asymptotic normality can be shown for the third assumption, which is a common assumption of the parametric bootstrap \citep{efron1982jackknife}. We observed that these assumptions hold empirically in most of our simulations, as we demonstrate in \ref{assumptionsjustification}. The coverage values that we obtained, given in \cref{Results}, add to the plausibility of these assumptions.

The entire method is summarised in \cref{alg:CIalg}. With relatively little extra effort - we only need to train the equivalent of $M(1+r)$ networks, with $r<1$ - we are able to get uncertainty estimates that translate to confidence and prediction intervals that are better theoretically founded, as is substantiated in the next subsection with a proof that the confidence intervals are guaranteed to be conservative, and empirically result in a better coverage, as is demonstrated in \cref{Results}. 

\begin{algorithm}[tb]
   \caption{Pseudo-code to obtain a confidence interval with Bootstrapped Deep Ensembles}
   \label{alg:CIalg}
\begin{algorithmic}[1]
   \STATE {\bfseries Input:} $M$ - number of ensembles, $N_{\textnormal{epoch}}$ - number of training epochs, $r$ - retrain fraction, $(X,Y)$ - data set   
   \FOR{$i=1$ {\bfseries to} $M$}
   \STATE Train ensemble member $i$ on $(X,Y)$ with random initialisation and data ordering to obtain $\hat{f}_{i}(\bm{x})$ and $\hat{\sigma}_{i}^{2}(\bm{x})$, while saving the model and optimizer state after $N_{\textnormal{epoch}}(1-r)$ training epochs. 
   \STATE Simulate new targets: $Y_{\text{new}} \sim \N{\hat{f}_{i}(X)}{\hat{\sigma}^{2}_{i}(X)}$.
   \STATE Repeat the final $rN_{\textnormal{epoch}}$ training epochs on $(X, Y_{\text{new}})$, obtaining $\hat{\hat{f}}_{i}(\bm{x})$.
    \ENDFOR
   \STATE $\hat{f}_{*}(\bm{x}) := \frac{1}{M}\sum_{i=1}^{M} \hat{f}_{i}(\bm{x})$
   \STATE $\sigmahd(\bm{x}) := \frac{1}{M}\sum_{i=1}^{M}\left(\hat{f}_{i}(\bm{x}) - \hat{\hat{f}}_{i}(\bm{x})\right)^{2}$
   \STATE $\sigmaht(\bm{x}) := \frac{1}{M-1} \sum_{i=1}^{M}\left(\hat{f}_{*}(\bm{x}) - \hat{f}_{i}(\bm{x}) \right)^{2}$
   \STATE Calculate the $1-\alpha$ confidence interval:\\ $\text{CI}^{(\alpha)}({\bm{x}}) = \left[\hat{f}_{*}(\bm{x}) \pm t_{\alpha/2}^{(M-1)} \sqrt{\sigmahd(\bm{x}) + \frac{\sigmaht}{M}}\right]$
\end{algorithmic}
\end{algorithm}

\subsection{Creating confidence and prediction intervals} \label{CIs}
\noindent The following theorem, proven in \ref{theoremproof}, states that, under Assumptions \ref{assumption2} and \ref{assumption3}, the pointwise confidence interval given in \cref{alg:CIalg} is conservative.
\begin{restatable}{theorem}{maintheorem}
\label{maintheorem} Following the notation introduced above, let $\hat{f}_{*}(\bm{x})$ be the average of the $M$ ensemble members with predictions $\hat{f}_{i}(\bm{x})$. Let $\hat{\hat{f}}_{i}(\bm{x})$ be the prediction of ensemble member $i$ after a part of the training is repeated with newly simulated targets. Define $\sigmaht(\bm{x})$ and $\sigmahd(\bm{x})$ as in \cref{alg:CIalg}. Under Assumptions \ref{assumption2} and \ref{assumption3}, with probability at least $(1-\alpha)\cdot 100 \%$:
\begin{equation}	
f(\bm{x}) \in \hat{f}_{*}(\bm{x}) \pm t_{\alpha/2}^{(M-1)} \sqrt{\sigmahd(\bm{x}) + \frac{\sigmaht(\bm{x})}{M}},
\end{equation}
where $t_{\alpha /2}^{(M-1)}$ is the critical value of a $t$-distribution with $M-1$ degrees of freedom.
\end{restatable}

The confidence interval from \cref{maintheorem} can be easily extended to a prediction interval. The prediction interval combines the aleatoric uncertainty, governed by a normal distribution with variance $\sigma^2(\bm{x})$, with the epistemic uncertainty, a scaled student distribution with $M-1$ degrees of freedom. \cref{alg: PIalg}  describes a simple Monte-Carlo sampling procedure to quickly estimate empirical quantiles of the resulting distribution.
\begin{algorithm}[tb]
   \caption{Pseudo-code to obtain a prediction interval with Bootstrapped Deep Ensembles}
   \label{alg: PIalg}
\begin{algorithmic}[1]
   \STATE $\hat{\sigma}^{2}(\bm{x}) = \frac{1}{M} \sum_{i=1}^{M}\hat{\sigma}_{i}^{2}(\bm{x})$
   \FOR{$j=1$ {\bfseries to} $N_{t}$}
   \STATE $t_{j} \sim t(M-1)$
   \STATE $\mu_{j}(\bm{x}) = \hat{f}_{*}(\bm{x}) + t_{j} \sqrt{\sigmahd(\bm{x}) + \frac{\sigmaht(\bm{x})}{M}}$
   \STATE $y_{j} \sim \N{\mu_{j}(\bm{x})}{\hat{\sigma}^{2}(\bm{x})}$
  \ENDFOR
  \STATE Take the $(1-\alpha / 2)$ and $\alpha/2$ empirical quantiles of all $y_{j}$ as the bounds of the PI 
\end{algorithmic}
\end{algorithm}

\section{Experimental results} \label{Results}
\noindent In this section, we empirically examine the quality of our confidence and prediction intervals. We first explain why and how we simulated data for our experiments. We then go through our three experiments that - 1 - show that the obtained confidence and prediction intervals have a typically better, or in some cases at least similar, coverage compared to other popular methods, - 2 - demonstrate the significant effect of random targets on the total uncertainty to underline the importance of incorporating this effect, and - 3 - show that our method is able to correctly estimate the separate variances due to random optimization and targets.

In the appendix we provide additional experimental results. Specifically, we also test the effect of differently distributed noise, a different simulation method, and different retraining fractions. We observe that our method works well for a variety of datasets using retraining fractions between 0.2 and 0.4, meaning that we do not need to tune it and can simply pick a default value.

\subsection{Simulating data} \label{testingmethodology}
\noindent In order to compare confidence intervals, it is necessary to know the true function values. A simple toy experiment would meet this requirement but is likely not representative for a real-world scenario. To overcome this, we created simulations based on the regression benchmark data sets used in \citet{hernandez2015probabilistic}. These data sets were also used in other works on uncertainty estimation \citep{gal2016dropout,lakshminarayanan2017simple,mancini2020prediction,liu2016stein,salimbeni2017doubly, khosravi2011comprehensive, pearce2020uncertainty, su2018tight} and have become the standard benchmark data sets for regression uncertainty quantification. We take one of these  real-world data sets, for instance Boston Housing, and train a random forest to predict $y$ given $\bm{x}$, and we use this model as the true function $f(\bm{x})$. We then train a second forest to predict the residuals squared $(y - f(\bm{x}))^{2}$ and use this forest as the true variance $\sigma^{2}(\bm{x})$. Using these $f(\bm{x})$ and $\sigma^{2}(\bm{x})$, we can simulate new targets from a $\N{f(\bm{x})}{\sigma^{2}(\bm{x})}$ distribution. We used random forests with 100 trees and a max depth of 3. The simulating procedure is summarized in Algorithm \ref{alg: simulationalgo}. 
\begin{algorithm}[tb]
   \caption{Pseudo-code to simulate data}
   \label{alg: simulationalgo}
\begin{algorithmic}[1]
   \STATE Train a random forest on $\mathcal{D}$ and use this forest as the true function $f(\bm{x})$
   \STATE Calculate the residuals, $(Y - f(X))$ 
   \STATE Train a second random forest on the squared residuals and use this function for the true variance $\sigma^{2}(\bm{x})$
   \STATE Simulate new targets: $y_{\text{new}} \sim \N{f(\bm{x})}{\sigma^{2}(\bm{x})}$
   \STATE \textbf{Return:} $\mathcal{D}_{\text{new}} = (X, Y_{\text{new}})$
\end{algorithmic}
\end{algorithm}

\begin{table*}[tb]
\caption{Results of the comparison of our method Bootstrapped Deep Ensembles (BDE) to Deep Ensembles (DE), the Naive Bootstrap (NB), Concrete Dropout (CD) and Quality Driven Ensembles (QDE). A total of 100 simulated data sets were used to calculate the metrics. On each new data set, new ensembles were trained, and new confidence and prediction intervals were constructed. Brier-CI80 denotes the Brier score of the CICF (Equation \eqref{CICF}) of an 80$\%$ confidence interval. Brier-PI80 denotes the equivalent quantity for the PICF (Equation \eqref{PICFequation}). We observe comparable Brier scores for most prediction intervals and significantly lower scores for all but two confidence intervals. The confidence intervals of BDE are wider but this is justified by the superior coverage. BDE and DE have identical RMSE since they use the same predictor. The RMSE value of CD was in general extremely similar. NB had a notably larger RMSE (often around 20$\%$). This is to be expected since each ensemble member is effectively being trained on less data due to the resampling. All RMSE values, as well as a further comparison to DE, including training without regularisation, with a different base simulation model, and differently distributed noise, can be found in \ref{extraresults}. Bold values are used to indicate the overall best brier score and underscored values are used to compare Bootstrapped Deep Ensembles and regular Deep Ensembles.}
\label{benchresults}
\begin{center}
\begin{small}
\begin{sc}
\vskip -0in	
\begin{adjustbox}{width=0.99\textwidth}
\begin{tabular}{lcccc|cccc|cccc|cccc}
\toprule
 \textbf{SIMULATION} & \multicolumn{4}{c}{\textbf{Brier-CI80} \textcolor{green}{$\downarrow$}}& \multicolumn{4}{c}{\textbf{Brier-PI80} \textcolor{green}{$\downarrow$}} & \multicolumn{4}{c}{\textbf{Width CI80}} & \multicolumn{4}{c}{\textbf{Width PI80}}  \\
 & BDE & DE & NB & CD & BDE & DE & QDE & CD & BDE & DE & NB & CD & BDE & DE & QDE & CD\\
  & \multicolumn{4}{c|}{$\times 10^{-2}$}&\multicolumn{4}{c|}{$\times 10^{-2}$} && \\
\midrule	
Boston   & \underline{\textbf{7.8}} & 11 & 18 & \textbf{7.8} &\underline{2.4} & 3.5 & 3.1 & \textbf{1.5} & 3.4 & 3.0 & 4.0 & 3.7 & 7.6 & 7.2 & 14.8 & 8.6 \\
Concrete  & \underline{\textbf{2.6}} & 8.2 & 33 & 15 & \underline{\textbf{0.80}} & 1.5 & 1.5 & 2.6 & 8.7 & 6.4 & 7.9 & 6.0  &  22.7 & 21.1 & 39.3 & 20.1 \\
Energy & \underline{4.0} & 6.7 & 37 & \textbf{1.5} & \underline{0.80} & 1.3 & 1.7 & \textbf{0.59} & 2.3 & 1.8 & 2.0 & 2.1 & 5.6 & 5.1 & 17.7 & 5.4 \\
Kin8nm & 0.78 & \underline{\textbf{0.63}} & 62 & 51 & \underline{\textbf{0.13}} & 0.25 & 0.93 & 3.1 & 0.17 & 0.14 & 0.12 & 0.024 &  0.53 & 0.52 & 0.69 & 0.44 \\
Naval & \underline{\textbf{2.1}} & 3.6 & 59 & 30 & 0.24 & \underline{0.19} & 1.1 & \textbf{0.11} & 2.2$^{1}$ &  1.4$^{1}$ & 1.2$^{1}$ & 0.7$^{1}$ & 0.03 & 0.03 & 0.03 & 0.03 \\
Power & \underline{\textbf{8.5}} & 12 & 33 & 36 & 0.08 & \underline{\textbf{0.07}} & 2.3 & 0.21 & 3.1 & 2.2 & 2.8 & 0.78&  10.7 & 10.4 & 19.2 & 9.9 \\
Yacht & \underline{\textbf{11}} & 14 & 22 & \textbf{11} & \underline{\textbf{2.8}} & 4.3 & 12 & 3.7 & 2.7 & 2.4 & 3.5 & 7.3 & 2.7 & 2.4 & 3.5 & 7.3 \\
Wine & \underline{\textbf{0.89}} & 2.6 & 8.6 & 26 & \underline{0.95} & 1.5 & \textbf{0.80} & 6.2& 0.50 & 0.45 & 0.68 & 0.27& 1.4 & 1.4 & 2.1 & 1.1 \\ 
\bottomrule 
\tiny{${}^{1}\times 10^{-3}$} 
\end{tabular}
\end{adjustbox}
\end{sc}
\end{small}
\end{center}
\vskip -0.3in
\end{table*}

\subsection{Training procedure}
\noindent We used neural networks with three hidden layers having 40, 30, and 20 units respectively, ReLU activations functions in these hidden layers, and a linear activation function in the final layer. To ensure positivity of $\hat{\sigma}$ we used an exponential transformation and added a minimum value of 1e-3 for numerical stability for the ensemble networks. Each network was trained for 80 epochs with a batchsize of 32 using the ADAM optimizer. We used the same the train\_test\_split function from the scikit learn package with random seed 1 for our train/test splits.

This setup is very similar to those used in \cite{gal2016dropout} and \cite{hernandez2015probabilistic} with the exception that we use more than one hidden layer. We do this because we observed a much larger bias when using only one layer. We stress that these are typical architectures and data sets for work on uncertainty quantification in a regression setting. Uncertainty estimation methods for regression are typically evaluated on smaller data sets and with smaller architectures than classification methods.

 We used $r=30\%$ and $M=5$. We used $l_{2}$ regularisation with a standard constant of $1 / (\# \text{Training Samples})$. The networks used for the Concrete Dropout and Quality Driven Ensembles methods were trained for 240 epochs each since we found that these networks needed longer to converge. All training data was standardized to have zero mean and unit variance before training. All testing was done on the original scales. 

 The code for all experiments has been made available at \url{https://github.com/LaurensSluyterman/Bootstrapped_Deep_Ensembles/tree/master}.

%

\subsection{Experiment 1: Simulations based on benchmark data sets} 
\noindent Our first experiment compares the coverage of our confidence and prediction intervals with different popular methods. Our prediction intervals are compared to DE, Concrete Dropout (CD) \citep{gal2017concrete}, and Quality-Driven Ensembles (QDE) \citep{pearce2018high}. Our confidence intervals are compared to DE, the Naive Bootstrap (NB) \citep{efron1982jackknife, heskes1997practical}, and Concrete Dropout. 

We test the coverage by using 100 simulated data sets instead of the popular practice of using a single test set. The latter tests empirical coverage of the intervals on a previously unseen part of a single real-world data set \citep{khosravi2011comprehensive, pearce2020uncertainty, su2018tight}. However, it has been argued that it is not sufficient to  evaluate the quality of prediction and confidence intervals in this manner on a test set. This only checks the overall coverage, which is relatively easy to tune, and not the quality of the pointwise confidence intervals \citep{sluijterman2021evaluate}. The proposed alternative is to use simulated data and calculate the coverage per $\bm{x}$-value over a large number of simulations. This gives rise to the Confidence Interval Coverage Fraction (CICF):
\begin{equation} \label{CICF}
\text{CICF}(\bm{x}) := \frac{1}{n_{\text{sim}}} \sum_{j = 1}^{n_{\text{sim}}}\mathds{1}_{f(\bm{x}) \in [LC^{(j)}(\bm{x}), RC^{(j)}(\bm{x})]},
\end{equation}
where $n_{\text{sim}}$ is the number of simulations, $f(\bm{x})$ is the true function value, and $LC^{(j)}(\bm{x}), RC^{(j)}(\bm{x})$ are the lower and upper limit of the CI of $f(\bm{x})$ in simulation $j$. If our CI has the correct pointwise coverage, the CICF($\bm{x}$) should be close to $1-\alpha$ for each value of $\bm{x}$. The following Brier score -- with a perfect score of 0 meaning a CICF of $1-\alpha$ for each individual value of $\bm{x}$ -- captures this:
\begin{equation}
\text{BS} = \frac{1}{n_{\text{test}}}\sum_{i = 1}^{n_{\text{test}}}\left(\text{CICF}(\bm{x}_{i}) -(1-\alpha)\right)^{2}.
\label{BS}	
\end{equation}

Evaluating the quality of the prediction interval is done similarly. We define the Prediction Interval Coverage Fraction:
\begin{equation} \label{PICFequation}
\text{PICF}(\bm{x}) := \frac{1}{n_{\text{sim}}} \sum_{j = 1}^{n_{\text{sim}}}\PR{y \in [L^{(j)}(\bm{x}), R^{(j)}(\bm{x})]},
\end{equation}
and report the resulting Brier scores. Additionally, we report the average widths of the intervals. In case of a comparable Brier score, we favor the method with smaller intervals. In summary, we create 100 new data sets, create 100 prediction and confidence intervals for each $\bm{x}$-value in the test set, and check how often the intervals contain the true value. 

\begin{figure*}[]
\centering
  \begin{minipage}{0.48\textwidth}
    \centering
   \hskip 0.38in \textbf{Confidence Intervals}
  \end{minipage}
  \begin{minipage}{0.48\textwidth}
    \centering
    \hskip -0.2in
	\textbf{Prediction Intervals}
  \end{minipage}
  \vskip 0.2in
  \includegraphics[width=0.95\textwidth]{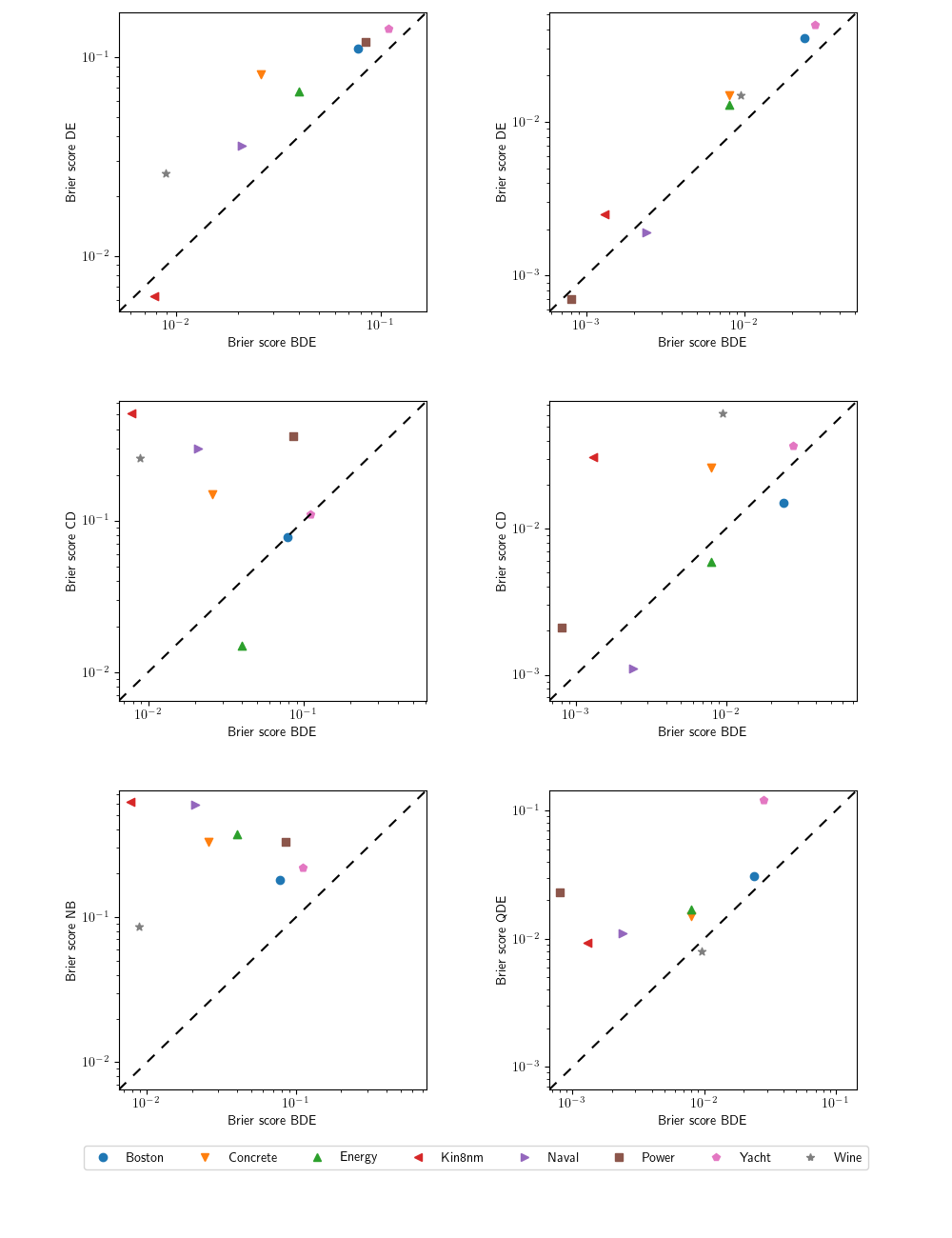}
\vskip -0.5in
\caption{Comparison of our method Bootstrapped Deep Ensembles (BDE) to Deep Ensembles (DE), the Naive Bootstrap (NB), Concrete Dropout (CD) and Quality Driven Ensembles (QDE). The Brier scores (see Table \ref{benchresults}) of BDE are plotted against those of the other methods. Our method has superior performance for all instances above the dotted diagonal. Additionally, the distance to the dotted line gives an indication of the difference in model performance. BDE clearly improves upon DE and is generally very robust.}
\label{fig: experiment1figure}
\vskip 0.2in
\end{figure*}

The results are presented in Table \ref{benchresults}, with Figure \ref{fig: experiment1figure} offering a visual summary of the various methods' performance.  The Brier scores of all data sets are plotted against the other methods. BDE perform better for all instances above the dotted diagonal line. Additionally, the distance to the diagonal gives an indication about the difference in performance. 

\begin{figure}[tb]
  \centering
  \includegraphics[width=0.8\linewidth]{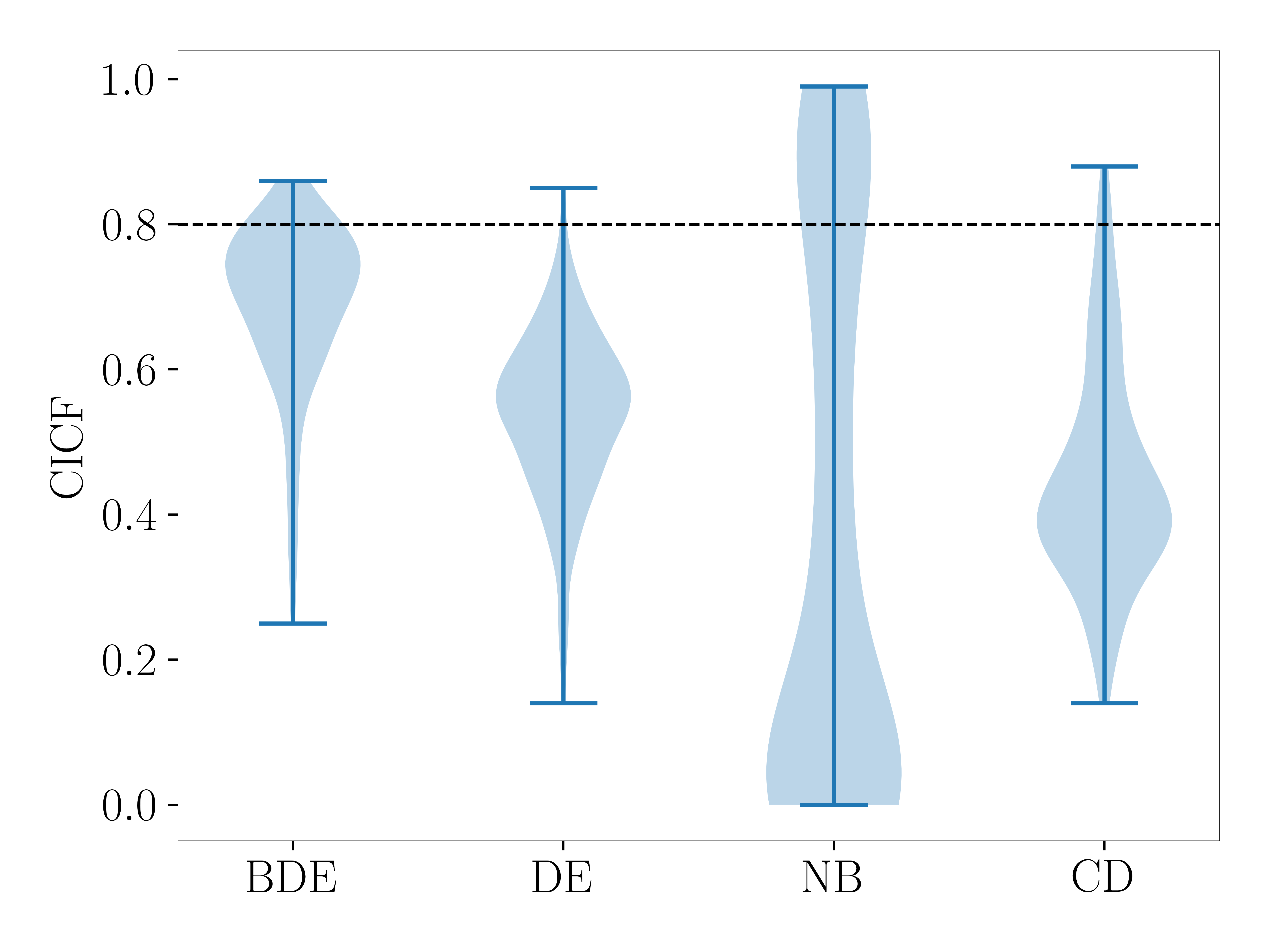}
  \caption{Violin plot of the individual CICF($\bm{x}$) values for 80$\%$ CIs, calculated on the test set of the \textit{Concrete} simulation. Each CICF value was obtained using 100 simulations. Violin plots of all other simulations, including for the PICF values, can be found in \ref{extraresults}. Note that perfect coverage would correspond to a sharp peak at $1-\alpha$. It can be seen that the confidence intervals for Deep Ensembles (DE) tend to be too optimistic, those for Concrete Dropout (CD) are often far too optimistic, and those for the Naive Bootstrap (NB) are all over the place.}
  \label{fig:test1}
  \vskip -0.1in
  \end{figure}

Several trends emerge from these results. Notably, our method clearly improves upon Deep Ensembles, producing superior Brier scores for the confidence intervals in seven of the eight data sets and for the prediction intervals in six of the eight data sets. For the other three data sets, the performance is very similar. The performance gain is often substantial. For example, on the Concrete data set, the Brier scores for the CICF range from 0.026 (BDE) to 0.33 (NB). To illustrate the significance of these differences, we visualised the individual CICF scores as a violin plot in \cref{fig:test1}. We observe that for most values of $\bm{x}$ the BDE confidence intervals are close to the correct size. 

 Moreover, our method is very robust. While other methods sometimes perform very poorly, as indicated by the large deviation from the diagonal line in Figure \ref{fig: experiment1figure}, our method does not. In particular, our confidence intervals are almost always better than those generated by the other methods and in the few cases that they are not, they still perform almost as well. The confidence intervals of Concrete dropout, for instance, perform similar on one data set and slightly better on two data sets, but dramatically underperform on the remaining five. 

\begin{figure}[h!]
\centering
\subfigure[Data point 13 in the BostonHousing test set]{\includegraphics[width=0.4\textwidth]{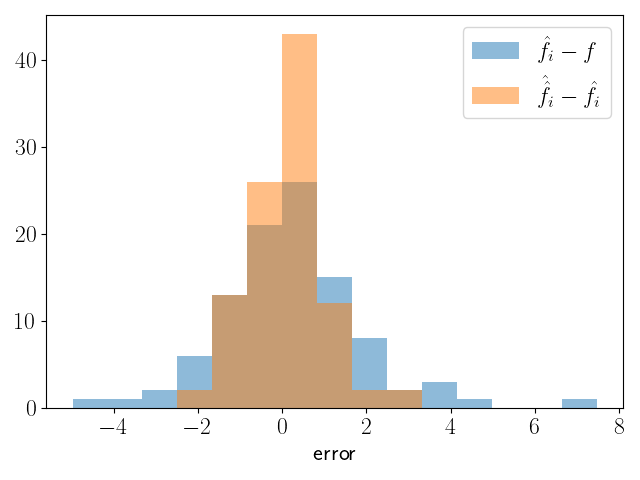}} 
\subfigure[Data point 15 in the BostonHousing test set]{\includegraphics[width=0.4\textwidth]{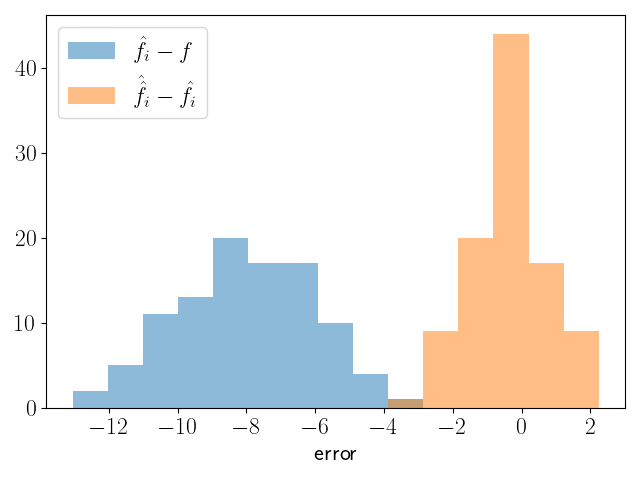}} 
\vspace{.3in}
\caption{Empirical example of Assumptions \ref{assumption2} and \ref{assumption3}. Each histogram is made by evaluating $\hat{f}_{1}(\bm{x}) - f(\bm{x})$ and $\hat{\hat{f}}_{1}(\bm{x}) - \hat{f}_{1}(\bm{x})$ for a single value of $\bm{x}$ on 100 simulated data sets. In (a) we see that the assumptions appear to hold. We have normally distributed errors with a slightly smaller variance after retraining. This should be the case since the aim of the retraining is to only capture the variance due to the random targets and not the random training. In (b), however, we see that for some values of $\bm{x}$, a large bias can occur.}
\label{fig: assumptionemperical}
\end{figure}

To check our assumptions, we kept track of all the predictions of the first ensemble member before and after retraining in each of the 100 simulations of the Boston Housing part of experiment 1. Figure \ref{fig: assumptionemperical} shows two typical scenarios. For most values of $\bm{x}$, we found that the errors are indeed normally distributed and that the variance of $\hat{f}_{i}(\bm{x}) - f(\bm{x})$ is slightly larger than the variance of $\hat{\hat{f}}_{i}(\bm{x}) - \hat{f}_{i}(\bm{x})$. This is expected, since the former also has variance due to the randomness of the training. We also see, however, that for some values of $\bm{x}$ we get a large bias term, violating Assumption \ref{assumption2}. Figure \ref{fig: moreerrorplots} in \ref{assumptionsjustification} shows these plots for the first 28 data points in the Boston Housing test set.

This violation of the \cref{assumption2} explains that some  of the BDE confidence intervals have a low coverage. We note that Deep Ensembles have the same problem, and that also on points with high bias, our method has better CICF values. 

\subsection{Experiment 2: Relative effect of random targets}
\noindent To further motivate the benefit of our method, we compared the relative contributions of random optimization and random data to the total variance of a neural network. As argued above, a neural network can be seen as a random predictor. This randomness is partially a consequence of the random optimization procedure and initialisation, which is captured well by Deep Ensembles. However, especially for small training sets, the classical variance due to random targets is also a significant part of the total variance. 

 We set up a simulation based on the large Protein-tertiary-structure data set. We trained two sets of 50 networks. The first set was trained with different targets for each network and the second was trained with the same targets for each network. The random targets were simulated using the two random forests. We subsequently examined the average variance of the networks with fixed and random targets on 5000 previously unseen test points. The average variance of the 50 networks trained on random targets gives an estimate of $\sigmat + \sigmad$, and the average variance of the 50 networks trained on the same targets gives an estimate of $\sigmat$. We took the difference of the two as an estimate of $\sigmad$. This process was repeated multiple times on an increasing number of data points $N$. 
 
\begin{figure}[tb]
		\centering
		\includegraphics[width=0.9\linewidth]{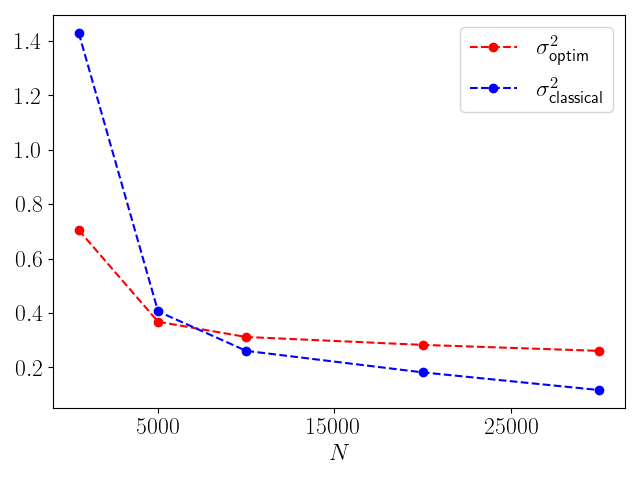}
		\vskip -0.1in
		\caption{The effect of random targets and random optimization on the total variance of the predictor $\hat{f}(\bm{x})$, trained on the \textit{Protein} data set. For small $N$, the effect of random targets is dominant, and even for quite large $N$, it is still significant. The variance due to random optimization, $\sigmat$, is obtained by examining the variance of 50 networks trained on the same data. Another 50 networks are trained on 50 different data sets, each with newly simulated targets. The variance of the second group is an estimate of $\sigmat + \sigmad$.}
		\vskip -0.1in
		\label{fig: Learning Curve}
\end{figure}

 Figure \ref{fig: Learning Curve} shows that with less than 5000 data points, the classical variance due to random targets is the dominant part of the total variance, and even with a lot of training data, the effect of random targets is still significant. This significant effect is not taken into account by standard DE, which explains the subpar coverage found in the first experiment (see Table \ref{benchresults} and Figure \ref{fig: experiment1figure}).
 
\subsection{Experiment 3: examining the separate estimates} 
\noindent Our method gives separate estimates for $\sigmad(\bm{x})$ and $\sigmat(\bm{x})$.
To test how well BDE can capture both components, we trained a BDE with $M=50$ and $r=30\%$ on a single data set, simulated with the two random forests. The variance of these 50 networks before retraining gives an estimate for $\sigmaht(\bm{x})$. Through our retraining step, we get an estimate for $\sigmahd(\bm{x})$. Subsequently, we trained 50 networks using newly simulated targets for each network. These targets were simulated using the random forests. The variance of the predictions of those second 50 networks gives an estimate of the ground truth $\sigmad(\bm{x}) + \sigmat(\bm{x})$. If our assumptions are correct, our estimate $\sigmahd(\bm{x})$ should be roughly the difference between  $\sigmad(\bm{x}) + \sigmat(\bm{x})$ and $\sigmat(\bm{x})$. 

Table \ref{experiment3table} shows that our separate estimates for the variances due to random targets and due to random training sum correctly - within approximately $10\%$ -  to the true variance when training with random targets. This shows that our approach of repeating a part of the training on new targets is an effective method to incorporate the uncertainty due to random targets without affecting accuracy. 

\begin{table}[b]
\vskip -0.2in
\caption{The quality of the average estimate for $\sigmad(\bm{x})$ on 4 different simulated data sets. The first column gives the ground truth of $\sigmad + \sigmat$, obtained by training with random targets. The sum of the estimates $\sigmaht$ and $\sigmahd$  match the ground truth within roughly $10\%$.}
\vskip -0in
\label{experiment3table}
\begin{center}
\begin{adjustbox}{width=0.48\textwidth}
\begin{tabular}{lcccr}
\toprule
\textbf{Simulation} & $\sigmad + \sigmat$& $\sigmaht$ & $\sigmahd$ &$\sigmahd + \sigmaht$  \\
		\midrule
		Boston   & 2.44 & 1.52 & 0.91 & 2.43\\
		Concrete  & 13.03 & 6.45 & 6.02  & 12.47\\
		Energy &  0.92 & 0.50 & 0.45 & 0.95\\
		kin8nm & 2.7e-3 & 1.7e-3 & 1.6e-3 & 3.3e-3 \\
		\bottomrule
\end{tabular}
\end{adjustbox}
\end{center}
\vskip -0.1in
\end{table}

\subsection{Limitations}
\noindent We end the results section by noting some of the limitations of our method.

 Most notably, it is important to realize that our assumptions will not always hold. The goal of our method was to incorporate the classical uncertainty that is a consequence from the fact that we are training on a random data set. In order to translate the predictions of the ensemble members to a confidence interval, we must make some distributional assumptions. Our assumptions are theoretically motivated by asymptotic analysis for parametric models (see \ref{assumptionsjustification}), but are not guaranteed to always hold in practice.

In particular, the unbiasedness assumption will not always hold. This is a problem with ensembling in general, and not specific to our method. If all ensemble members have a certain bias, then the corresponding confidence interval can easily have a very low coverage. Similarly, if the additive noise is not Gaussian, then the coverage can be imperfect. We investigated this in more detail in \ref{extraresults} and found this effect to be noticeable but much less substantial than the effect of a large bias.

Our method is 30$\%$ more expensive than regular Deep Ensembles. While our tailor-made bootstrapping approach is far more efficient than the alternative of training an ensemble of ensembles, this 30$\%$ may be significant depending on the application. Additionally, at inference time, a total of $2M$ forward passes needs to be made, in comparison to $M$ for regular Deep Ensembles.
\section{Conclusion} \label{Conclusion}
\noindent In this paper, we presented our uncertainty estimation method Bootstrapped Deep Ensembles. The BDE confidence intervals have much better coverage than those obtained with standard DE or other popular methods, at a price of just $30\%$ more training time. BDE improves upon DE because it incorporates the epistemic uncertainty due to the randomness of the training targets, where DE only captures the randomness of the optimization procedure. Our simulations show that the randomness of the training targets is substantial, even for larger data sets. 

Where, based on asymptotic statistical theory, one would expect this variance to be inversely proportional to the number of data points, we observed a slower decay, closer to $1 \over \sqrt{N}$. It would be interesting to study the (asymptotic) behavior of these two components in more detail, also to be able to judge when one can indeed be neglected compared to the other. As a potential bonus, to be investigated in more detail in future work, our method appears to better detect overfitting than standard DEs. Arguments and initial empirical evidence can be found in \ref{overfittingappendix}.

In regions with relatively little data, confidence intervals tend to get larger. In general, however, we would like to discourage the use of confidence intervals for out-of-distribution detection: confidence intervals may get wider for quite different reasons, in particular when we allow for heteroscedastic noise, and, perhaps more importantly, they simply cannot be trusted in regions with little training data, since the underlying assumptions on which they are based are doomed to be violated. A more promising  avenue for future work is to combine our method with an orthogonal approach, specifically for OoD detection (such as, for example, \citet{ren2019likelihood}). 

\FloatBarrier
\section*{References}
\bibliographystyle{apalike}
\bibliography{../../references}

\newpage

\appendix
\onecolumn

\begin{center}
\section*{APPENDIX}
\end{center}

\noindent This appendix consists of four parts. In \ref{theoremproof} we provide the proof of Theorem \ref{maintheorem}. In \ref{assumptionsjustification}, we motivate the assumptions on which Bootstrapped Deep Ensembles rely. We provide additional experimentation in \ref{extraresults} and briefly investigate the possibility to detect overfitting in \ref{overfittingappendix}.

\section{Proof of Theorem \ref{maintheorem}} \label{theoremproof}
\paragraph{Proof} The result follows by evaluating 
\begin{equation}\label{statistic}
T^{2} = \frac{(f(\bm{x}) - \hat{f}_{*}(\bm{x}))^{2}}{\sigmahd(\bm{x}) + \frac{\sigmaht(\bm{x})}{M}}.
\end{equation}
We recall that our estimates for $\sigmat(\bm{x})$ and $\sigmad(\bm{x})$ are given by:
\[
\sigmaht(\bm{x}) = \frac{1}{M-1} \sum_{i=1}^{M}\left(\hat{f}_{i}(\bm{x}) - \frac{1}{M}\sum_{i=1}^{M} \hat{f}_{i}(\bm{x}) \right)^{2},
\]
and
\[
\sigmahd(\bm{x}) = \frac{1}{M}\sum_{i=1}^{M}\left(\hat{f}_{i}(\bm{x}) - \hat{\hat{f}}_{i}(\bm{x})\right)^{2}.
\]
Assumption \ref{assumption2} tells us that
\[
(f(\bm{x}) - \hat{f}_{*}(\bm{x}))^{2} =\left(\sigmad(\bm{x}) +\frac{\sigmat}{M}\right)\epsilon_{0}^{2}, \quad \text{with} \quad   \epsilon_{0} \sim \N{0}{1},
\]
and
\[
\sigmaht(\bm{x}) = \frac{\sigmat(\bm{x})}{M-1} \zeta_{\text{o}}, \quad \text{with} \quad \zeta_{\text{o}} \sim \chi^{2}(M-1).
\]
Assumption \ref{assumption3} implies
\[
\sigmahd(\bm{x}) = \frac{\sigmad(\bm{x})}{M} \zeta_{\text{c}},\quad \text{with} \quad\zeta_{\text{c}} \sim \chi^{2}(M).
\]
This enables us to rewrite equation \eqref{statistic} to
\begin{equation} \label{T2rewritten}
T^{2} = \frac{\epsilon_{0}^{2}}{ \gamma W_{\text{o}} + (1-\gamma) W_{\text{c}}},
\end{equation}
with 
\[
W_{\text{o}} = \frac{\zeta_{\text{o}}}{M-1}, \quad \text{and}, \quad  W_{\text{c}} = \frac{\zeta_{\text{c}}}{M}, 
\]
and
\[
\gamma = \frac{\frac{\sigmat(\bm{x})}{M}}{\sigmad(\bm{x}) + \frac{1}{M}\sigmat(\bm{x})},
\]
where we dropped the dependence of $\gamma$ on $\bm{x}$ to simplify notation. Our goal is to bound the following probability:
\begin{equation}\label{probtobound}
\mathbb{P}\left(\frac{\epsilon_{0}^{2}}{ \gamma W_{\text{o}} + (1-\gamma) W_{\text{c}}} > F_{1-\alpha}(1, M-1) \right), 
\end{equation}
which we can rewrite as
\[
\resizebox{0.46\textwidth}{!}{
$\int\int\left(\mathbb{P}\left(\frac{\epsilon_{0}^{2}}{ \gamma w_{\text{o}} + (1-\gamma) w_{\text{c}}} > F_{1-\alpha}(1, M-1) \right)  \right) \\
dG_{\text{o}}(w_{\text{o}})dG_{\text{c}}(w_{\text{c}}),$}
\]
where $G_{\text{o}}(w_{\text{o}})$ is the cumulative distribution function of $W_{\text{o}}$ and $G_{\text{c}}(w_{\text{c}})$ is the cumulative distribution function of $W_{\text{c}}$. We define the conditional probability in the integral as $\phi(\gamma)$: 
\begin{equation}\label{phi}
\phi(\gamma) := \mathbb{P}\left(\frac{\epsilon_{0}^{2}}{ \gamma w_{\text{o}} + (1-\gamma) w_{\text{c}}} > F_{1-\alpha}(1, M-1) \right).
\end{equation}
The next step is to show that $\phi(\gamma)$ is convex. Let $H$ be the CDF of $\epsilon_{0}^{2}$, which has a $\chi^{2}(1)$ distribution, then $h=H'$ is strictly decreasing, which implies $h'<0$. 
We can rewrite $\phi(\gamma)$ as
\[
\phi(\gamma) = 1 - H\big((\gamma w_{\text{o}} + (1-\gamma) w_{\text{c}} ) F_{1-\alpha}(1, M-1) \big),
\]
which gives
\[
\phi'(\gamma) = - (w_{\text{o}} - w_{\text{c}})F_{1-\alpha}(1, M-1)h\big((\gamma w_{\text{o}} + (1-\gamma) w_{\text{c}} ) F_{1-\alpha}(1, M-1) \big)
,\]
and 
\begin{align*}
\phi''(\gamma) &= - (w_{\text{o}} - w_{\text{c}})^{2}F_{1-\alpha}(1, M-1)^{2} \\
& \quad \cdot h'\big((\gamma w_{\text{o}} + (1-\gamma) w_{\text{c}} ) F_{1-\alpha}(1, M-1) \big) >0. 
\end{align*}
This means that $\phi(\gamma)$ is convex, which implies that equation \eqref{probtobound} is convex. The maximum of equation \eqref{probtobound} is therefore either at $\gamma=0$ or $\gamma=1$. Evaluating equation \eqref{T2rewritten} shows that taking $\gamma=0$ gives $T^{2}$ an $F(1, M)$ distribution and taking $\gamma=1$ gives $T^{2}$ an $F(1, M-1)$ distribution. Since $F_{\alpha}(1, M) < F_{\alpha}(1, M-1)$ for all $\alpha$, we get
\[
\mathbb{P}\left(\frac{(f(\bm{x}) - \hat{f}_{*}(\bm{x}))^{2}}{\sigmahd(\bm{x}) + \frac{\sigmaht(\bm{x})}{M}} > F_{1-\alpha}(1, M-1)\right) \leq \alpha. \]
$\hfill \square$

\section{Motivation of assumptions}\label{assumptionsjustification}
\noindent Our method relies on three assumptions. We will first provide a theoretical motivation of these assumptions and then provide some additional empirical support. 

\subsection{Theoretical motivation of assumptions}

\noindent Assumption \ref{assumption1} is a common modeling assumption that may or may not hold depending on the data. The second assumption claims that the output of the neural network is normally distributed. This normality is a very standard assumption. The typical reasoning is that in a deterministic parametric model without regularisation this assumption holds asymptotically. The same asymptotic normality can be shown for the third assumption, which is a common assumption of the parametric model. 

We now provide the proof of these statements for a parametric model. Here, there is no variance due to training and we need to show - 1 - that the output of the model is normally distributed and - 2 - that if we train the model again on simulated targets, that the output will still be normally distributed with roughly equal variance. We stress that this is not a proof that our assumptions hold, which is impossible to prove for a neural network, but a proof of the result for a parametric model which motivates the assuptions.

 Let $\theta$ be the parameters that parametrize our network. Let $p_{\theta}(\mathcal{D})$ be the likelihood of the data given $\theta$. Our setup corresponds to finding the $\theta$ that maximizes $p_{\theta}(\mathcal{D})$:
\[
\hat{\theta} = \text{arg}\max_{\theta} p_{\theta}(\mathcal{D}),
\]
and subsequently finding $\hat{\hat{\theta}}$ that maximises $p_{\theta}(\mathcal{D}_{\text{new}})$:
\[
\hat{\hat{\theta}} = \text{arg}\max_{\theta} p_{\theta}(\mathcal{D}_{\text{new}}),
\]
Furthermore, we define
\begin{equation} \label{Fisher}
I(\theta_{0}) := \text{Cov}_{\theta_{0}} \left. \frac{\partial}{\partial \theta} \log(p_{\theta}(\bm{x}_{1},y_{1})) \right\rvert_{\theta_{0}}.\end{equation}

Under certain consistency and regularity conditions it is possible to show that 
\[
\sqrt{n}(\hat{\theta} - \theta_{0}) \rightarrow \N{0}{I(\theta_{0})^{-1}},
\]
and 
\[
\sqrt{n}(\hat{\hat{\theta}} - \hat{\theta}) \rightarrow \N{0}{I(\hat{\theta})^{-1}}.
\]

For a proof and clarification of the assumed consistency and regularity see \citet{van2000asymptotic} or \citet{nonlinear2003}.

With $\hat{f}_{i}$ and $\hat{\hat{f}}_{i}$ we denote the output of the mean prediction of an ensemble member before and after repeating part of the training. The delta method gives the variance of $\hat{f}_{i}$ and $\hat{\hat{f}}_{i}$:

 \[
\V{\hat{f}_{\hat{\theta}}(\bm{x})} = \left. \frac{\partial}{\partial \theta} f_{\theta}(\bm{x}) \right\rvert_{\theta_{0}} \V{\hat{\theta}} \left( \left. \frac{\partial}{\partial \theta} f_{\theta}(\bm{x}) \right\rvert_{\theta_{0}}\right)^{T},
\]
and 
\[
\V{\hat{\hat{f}}_{\hat{\hat{\theta}}}(\bm{x})} = \left. \frac{\partial}{\partial \theta} f_{\theta}(\bm{x}) \right\rvert_{\hat{\hat{{\theta}}}} \V{\hat{\hat{\theta}}} \left( \left. \frac{\partial}{\partial \theta} f_{\theta}(\bm{x}) \right\rvert_{\hat{\hat{\theta}}}\right)^{T}.
\]
Under the assumed consistency, $\hat{\theta}$ and $\hat{\hat{\theta}}$ will be close and thus $\left. \frac{\partial}{\partial \theta} f_{\theta}(\bm{x}) \right\rvert_{\hat{\theta}}$ will be close to $ \left. \frac{\partial}{\partial \theta} f_{\theta}(\bm{x}) \right\rvert_{\hat{\hat{\theta}}}$. By the same consistency, $I(\theta_{0})$ and $I(\hat{\theta})$ will be close. 

\subsection{Empirical assessment of assumptions}
\begin{figure}[h!]
\centering
\vskip 0in
{\includegraphics[width=.24\textwidth]{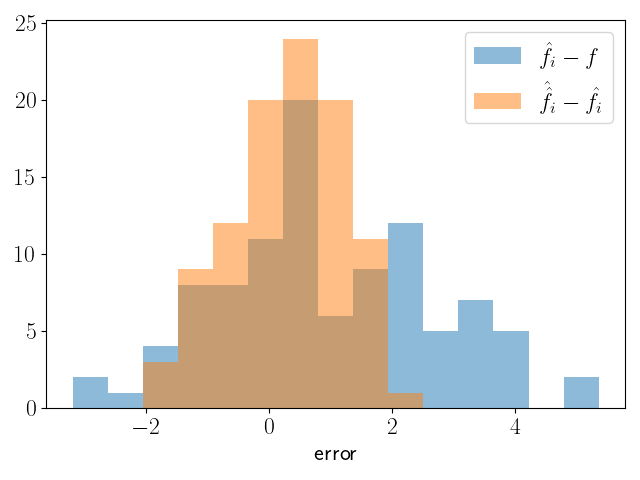}}
{\includegraphics[width=.24\textwidth]{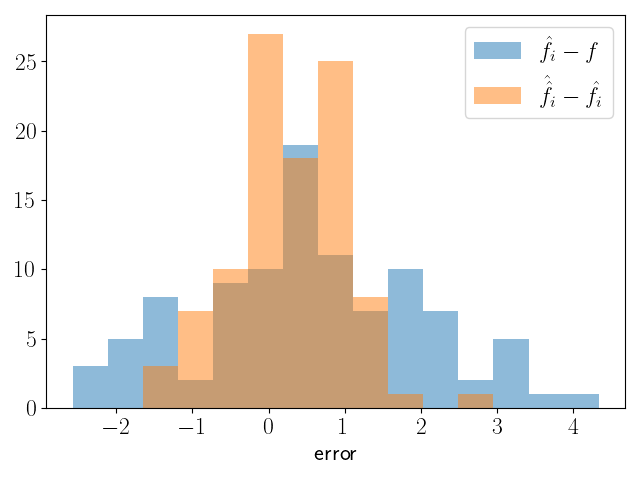}}
{\includegraphics[width=.24\textwidth]{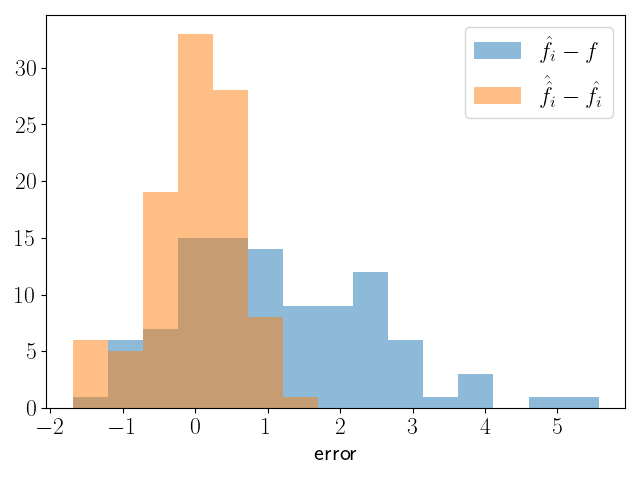}}
{\includegraphics[width=.24\textwidth]{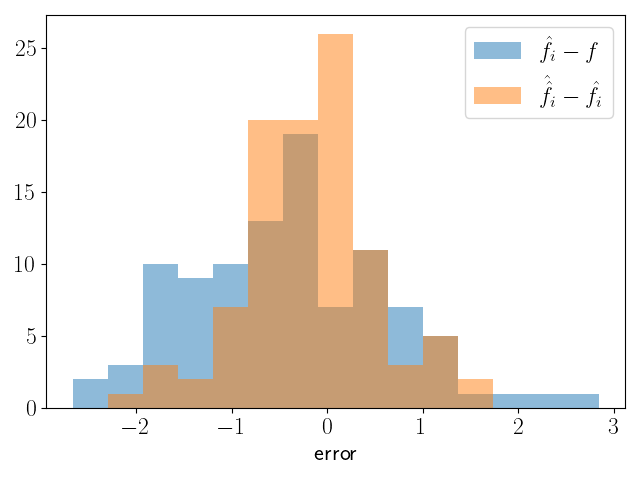}}
{\includegraphics[width=.24\textwidth]{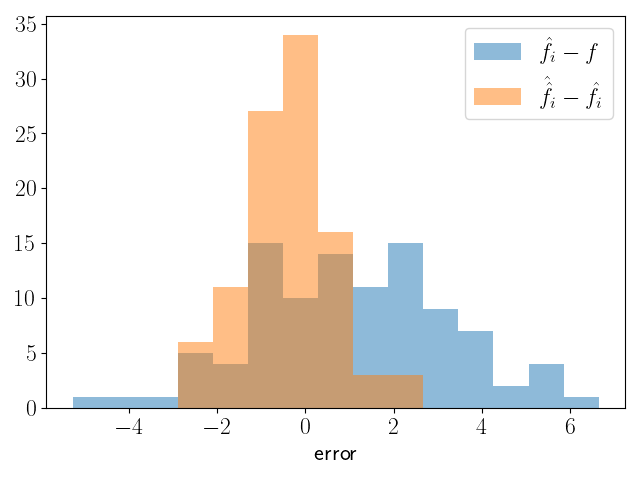}}
{\includegraphics[width=.24\textwidth]{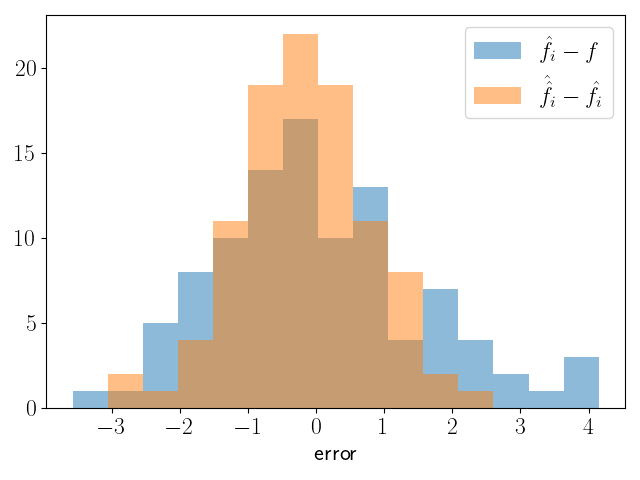}}
{\includegraphics[width=.24\textwidth]{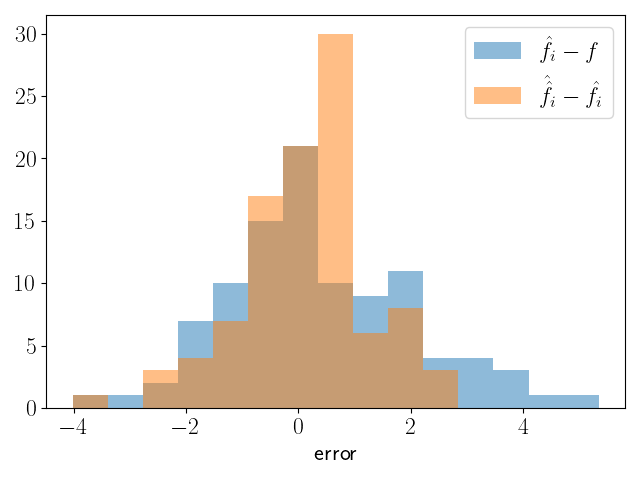}}
{\includegraphics[width=.24\textwidth]{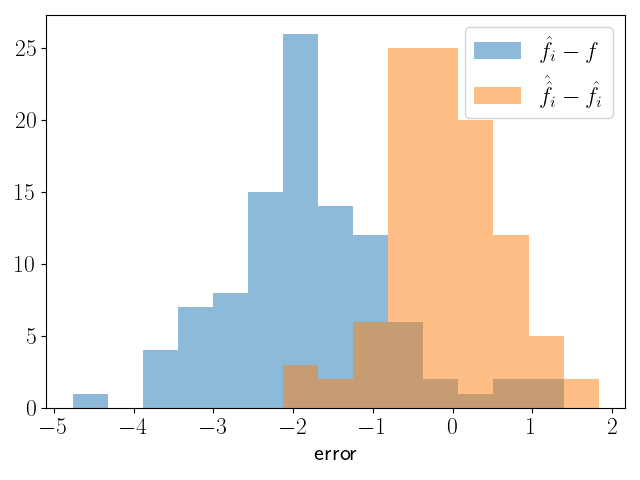}}
{\includegraphics[width=.24\textwidth]{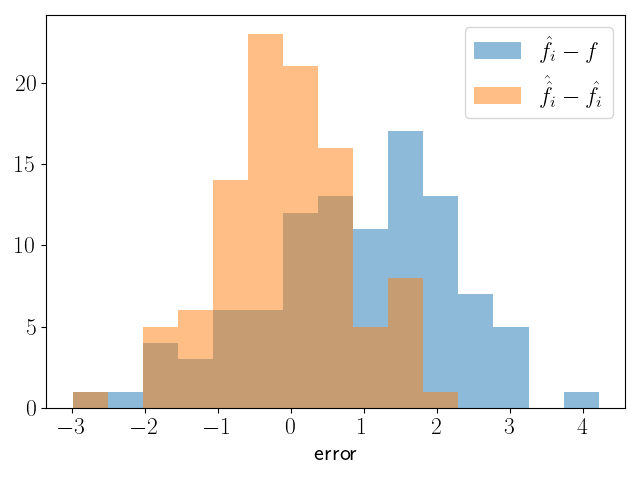}}
{\includegraphics[width=.24\textwidth]{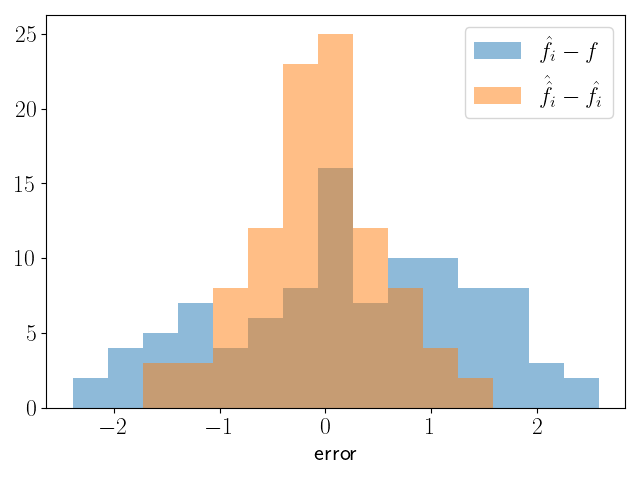}}
{\includegraphics[width=.24\textwidth]{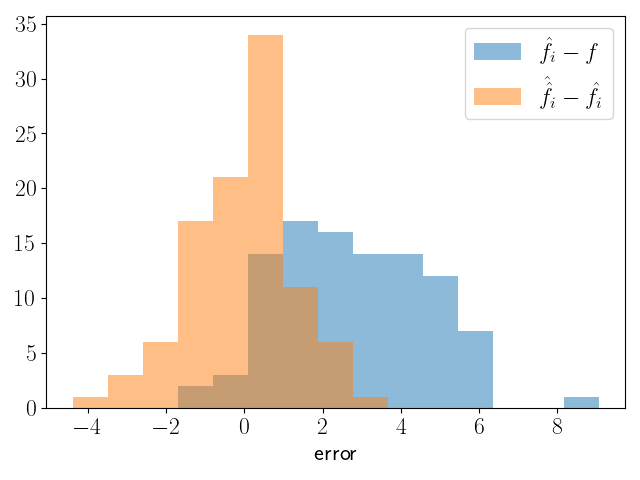}}
{\includegraphics[width=.24\textwidth]{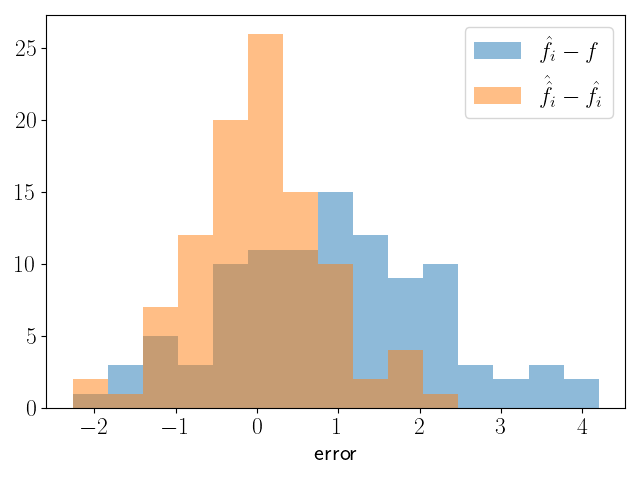}}
{\includegraphics[width=.24\textwidth]{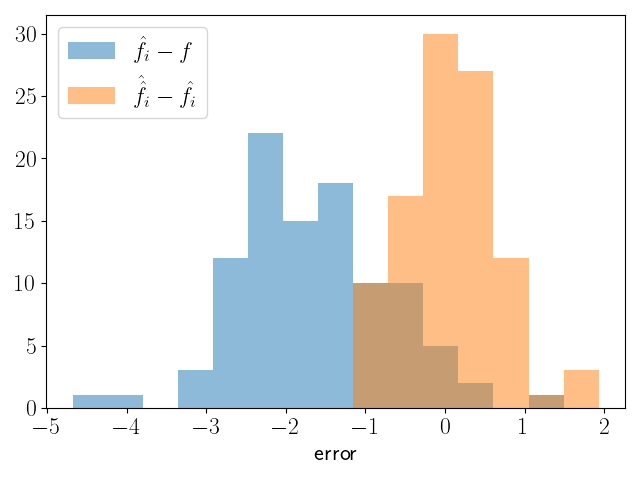}}
{\includegraphics[width=.24\textwidth]{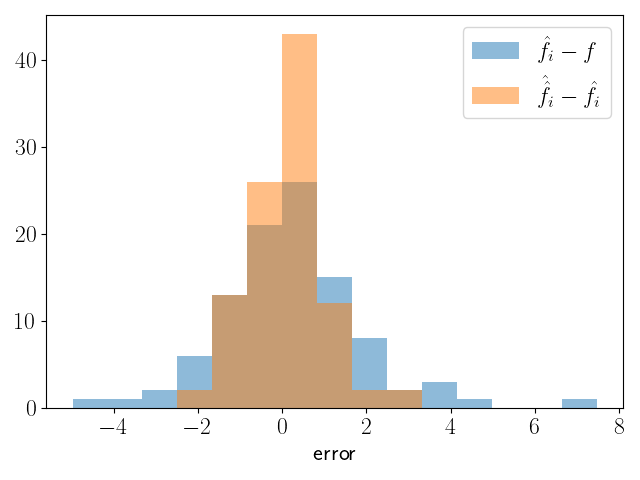}}
{\includegraphics[width=.24\textwidth]{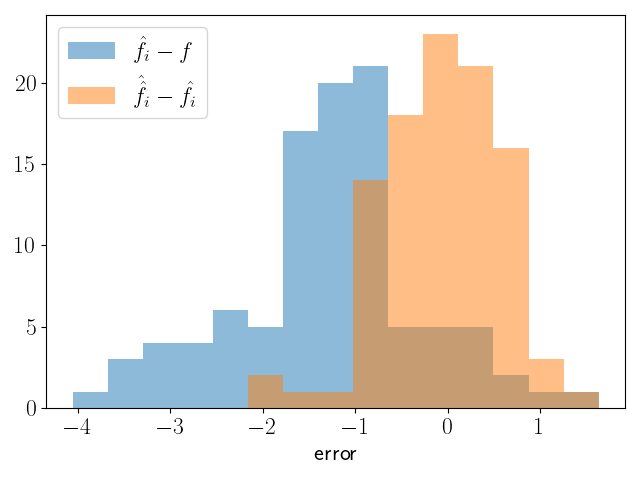}}
{\includegraphics[width=.24\textwidth]{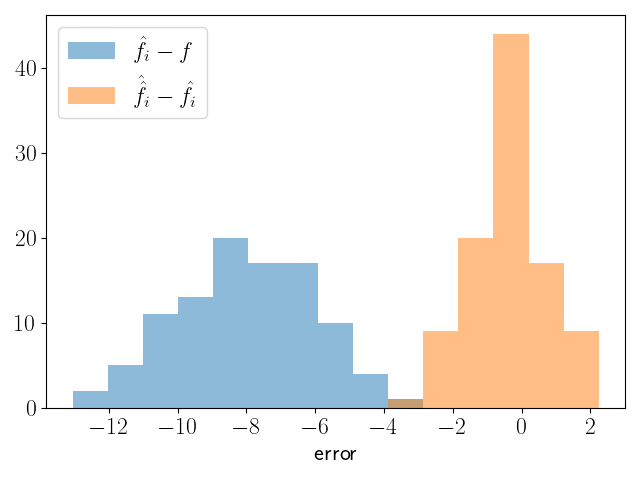}}
{\includegraphics[width=.24\textwidth]{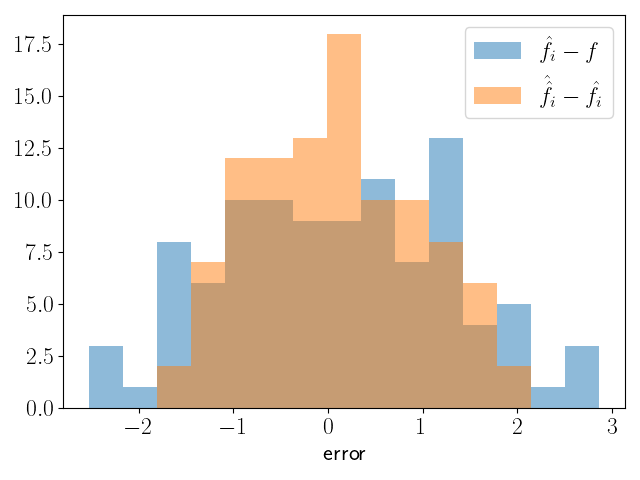}}
{\includegraphics[width=.24\textwidth]{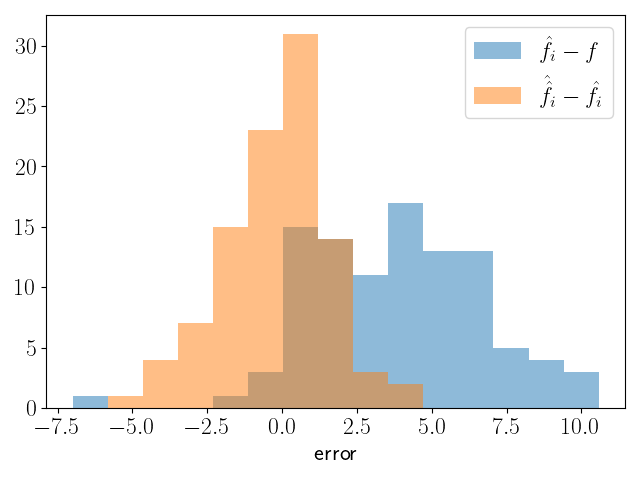}}
{\includegraphics[width=.24\textwidth]{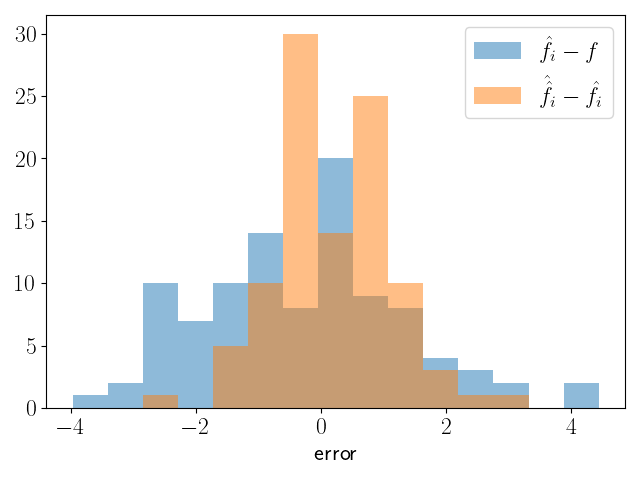}}
{\includegraphics[width=.24\textwidth]{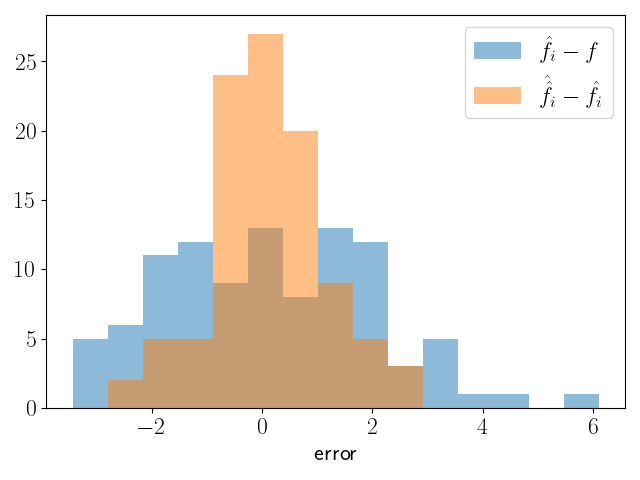}}
{\includegraphics[width=.24\textwidth]{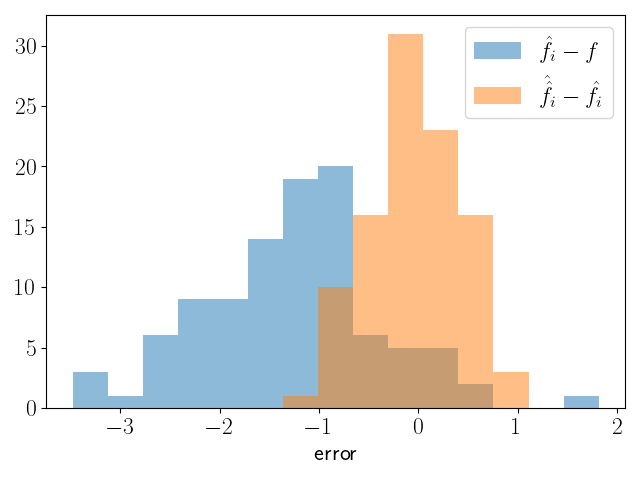}}
{\includegraphics[width=.24\textwidth]{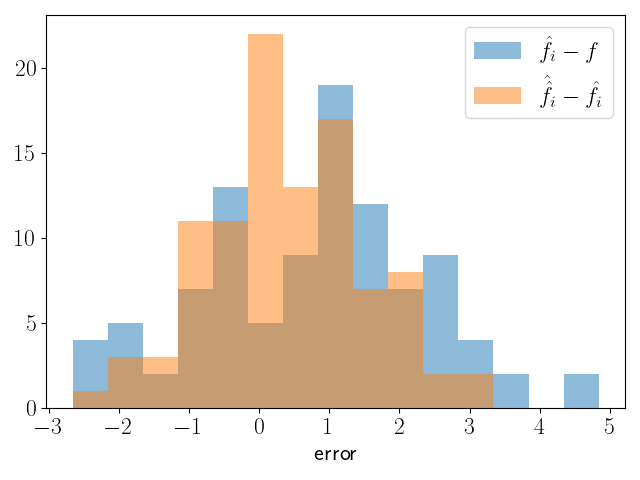}}
{\includegraphics[width=.24\textwidth]{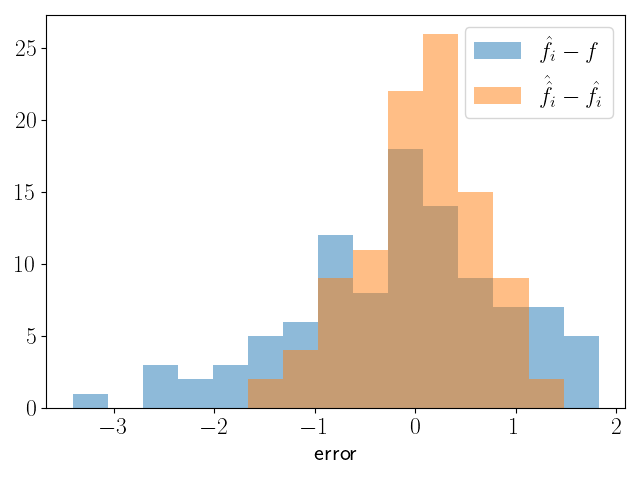}}
{\includegraphics[width=.24\textwidth]{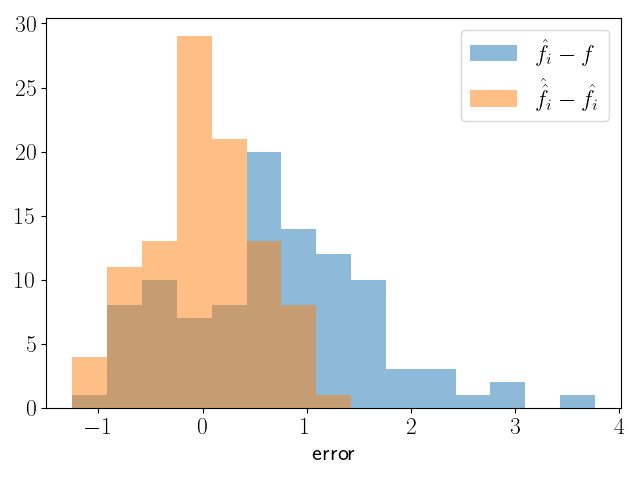}}
{\includegraphics[width=.24\textwidth]{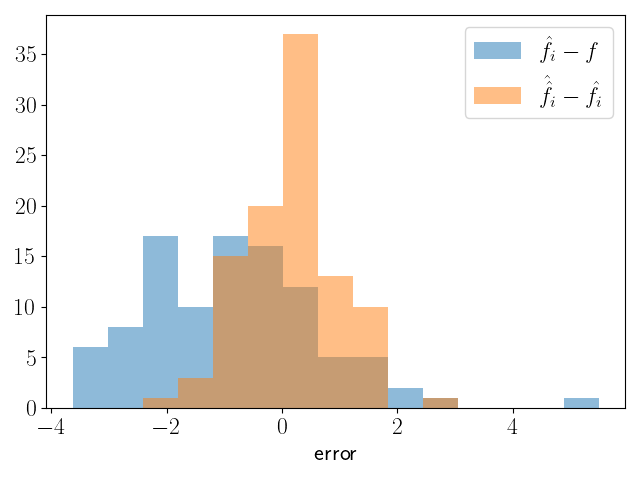}}
{\includegraphics[width=.24\textwidth]{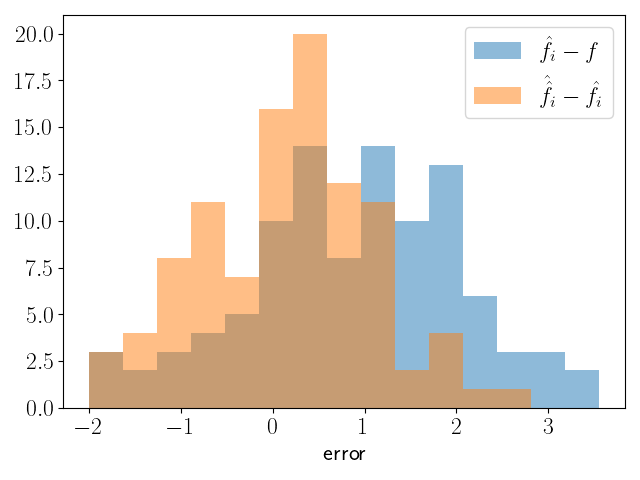}}
{\includegraphics[width=.24\textwidth]{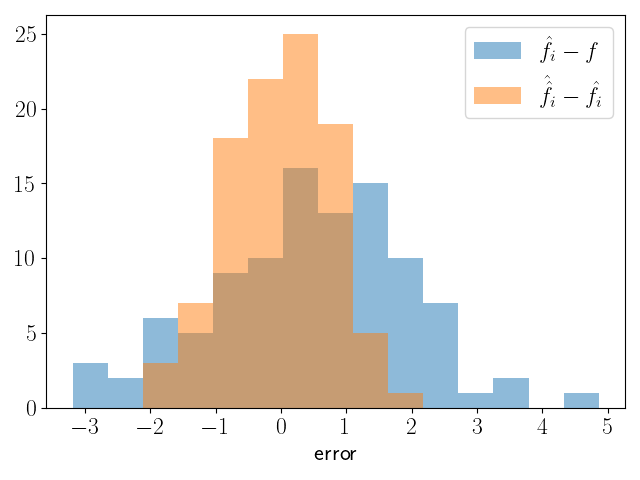}}
{\includegraphics[width=.24\textwidth]{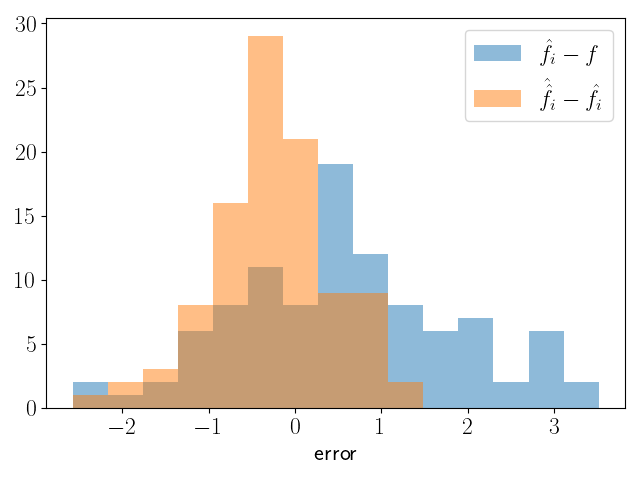}}
\caption{The same error plots as in \cref{fig: assumptionemperical} for the first 28 data points in the Boston Housing test set. The assumed normality seems to hold well. However, the original predictor is often biased.}
\vskip -0.6in
\label{fig: moreerrorplots}
\end{figure}

During the Boston Housing part of experiment 1, we kept track of all the predictions of the first ensemble member before and after retraining in each of the 100 simulations. For most values of $\bm{x}$, we found that the errors are indeed normally distributed and that the variance of $\hat{f}_{i}(\bm{x}) - f(\bm{x})$ is slightly larger than the variance of $\hat{\hat{f}}_{i}(\bm{x}) - \hat{f}_{i}(\bm{x})$. This is expected, since the former also has variance due to the randomness of the training. We also see, however, that for some values of $\bm{x}$ we get a large bias term, violating Assumption \ref{assumption2}. Figure \ref{fig: moreerrorplots} shows these plots for the first 28 data points in the Boston Housing test set.

\section{Additional experimentation} \label{extraresults}

\noindent In this section, we provide additional experimental results. We repeated parts of Experiment 1 in \cref{Results} with the following alterations:

\begin{enumerate}
	\item We used differently distributed noise -- violating Assumption \ref{assumption1} -- in order to see how this affects the confidence intervals.
	\item We removed all regularisation in the neural networks.
	\item We used a different simulation model. Recall that we used a random forest that was trained on real-world data sets to be able to simulate data for our experiments. We replaced the random forest with a neural network as the true function, $f(\bm{x})$.
	\item We used different retraining fractions. 
\end{enumerate} 
We also give the RMSE values from Experiment 1 and provide the violin plots for all the prediction and confidence intervals.

\subsection{Differently distributed noise}
\noindent Here we study what would happen if we misspecified our model. In order to test this, we simulated data with additive $t(3)$ and $\Gamma(1/10, \sqrt{10})$ distributed noise, denoted with $\epsilon$.  In order to make the experiments comparable to the earlier ones, we used the variance $\sigma^{2}(\bm{x})$ from the random forest and a scaling factor, $C$, to obtain a comparable size heteroscedastic variance:
\[
y = f(\bm{x}) + C\sigma(\bm{x}) \epsilon
\]
For the $t(3)$ distribution we have $C = \sqrt{1/3}$ and for the $\Gamma(1/10, \sqrt{10})$ distribution we have $C =1$.

\begin{figure}[h]
\centering
\subfigure[$\alpha = 0.2$]{\includegraphics[width=0.4\textwidth]{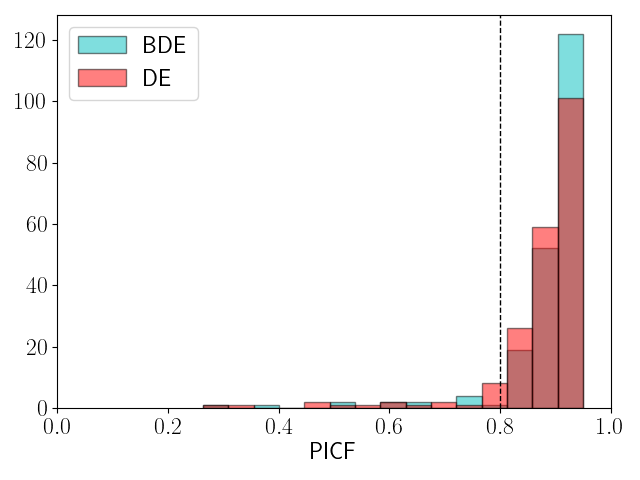}} 
\subfigure[$\alpha = 0.3$]{\includegraphics[width=0.4\textwidth]{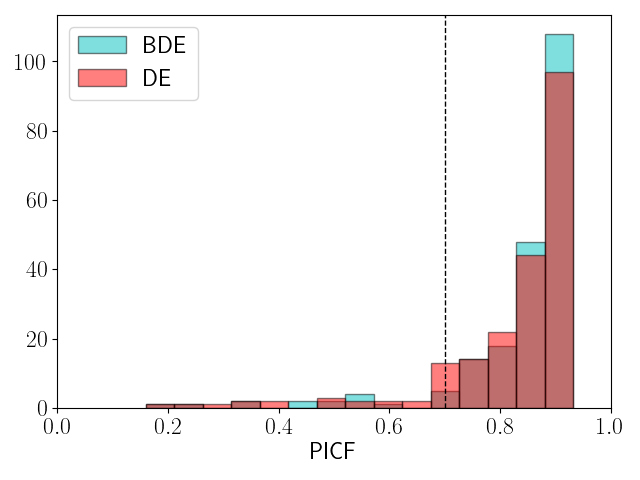}} 

\caption{Histogram of the individual PICF($\bm{x}$) values calculated on the test set of the Concrete simulation. Each point in the histogram represents the fraction of times the true function value $f(\bm{x})$ was inside the confidence interval calculated over 100 simulations. The skewness of the gamma distribution shows in the error of the PICF.}
\label{fig: PICFgamma}
\end{figure}

Table \ref{bench-different-noise} illustrates that BDE still produce better confidence intervals than DE, but that the performance is affected by the violation of Assumption \ref{assumption1}. In the case of the, purposely very skewed, gamma distribution, we also see that the prediction intervals are no longer calibrated, as is illustrated in Figure \ref{fig: PICFgamma}.

\begin{table}[h]
\caption{Results of the comparison of our method Bootstrapped Deep Ensembles (BDE) to Deep Ensembles (DE) on the Concrete simulation using differently distributed additive noise. A total of 100 simulated data sets were used to calculate the metrics. Brier-CI$A$ denotes the Brier score of the CICF of an $A\%$ confidence interval.}
\label{bench-different-noise}
\vskip 0.2in
\centering
\begin{tabular}{lcc|cc|cc|cc}
\toprule
 \textbf{Noise} & \multicolumn{2}{c}{\textbf{Brier-CI90} \textcolor{green}{$\downarrow$}} & \multicolumn{2}{c}{\textbf{Brier-CI80} \textcolor{green}{$\downarrow$}} &  \multicolumn{2}{c}{\textbf{Brier-CI70} \textcolor{green}{$\downarrow$}} & \multicolumn{2}{c}{\textbf{Width CI90}}
 \\
& BDE & DE & BDE & DE & BDE & DE & BDE & DE  \\
& \multicolumn{2}{c|}{$\times 10^{-2}$} & \multicolumn{2}{c|}{$\times 10^{-2}$} & \multicolumn{2}{c|}{$\times 10^{-2}$}& &  \\
\midrule	
$\N{0}{1}$ & \textbf{6.1} & 9.7 &  \textbf{7.8} & 11 & \textbf{7.7} & 10 & 4.7 & 4.1 \\
$t(3)$ & \textbf{6.3} & 9.4 & \textbf{8.1} & 11 & \textbf{7.7} & 10 & 4.5 & 3.9 \\
$\Gamma(1/10, \sqrt{10})$ &  \textbf{9.6} & 12 & \textbf{12} & 13 & \textbf{11} & 12 & 8.1 & 7.6 \\
\bottomrule 
\end{tabular}
\end{table}

\FloatBarrier

\subsection{No regularisation}
\noindent To examine the effect of regularisation, we repeated a part of Experiment 1 without any regularisation. We note that the regularisation seems to have hardly any effect on the outcome. This could be the result of the fixed training time of 80 epochs and a relatively simple neural network architecture preventing overfitting.

\begin{table*}[h]
\caption{Results of the comparison of our method Bootstrapped Deep Ensembles (BDE) to Deep Ensembles (DE) without any regularisation. A total of 100 simulated data sets were used to calculate the metrics. On each new data set, new ensembles were trained, and new confidence and prediction intervals were constructed. Brier-CI90 denotes the Brier score of the CICF of a 90$\%$ confidence interval. Brier-PI90 denotes the equivalent quantity for the PICF. RMSE gives the root mean squared error of the predictions with respect to the targets. Since the predictor is identical for both methods, there is only one value.}
\label{bench-regularisation}
\begin{center}
\begin{tabular}{lcc|cc|c|cc|cc}
\toprule
 \textbf{SIMULATION} & \multicolumn{2}{c}{\textbf{Brier-CI90} \textcolor{green}{$\downarrow$}} & \multicolumn{2}{c}{\textbf{Brier-PI90} \textcolor{green}{$\downarrow$}} &  \multicolumn{1}{c}{\textbf{RMSE}} & \multicolumn{2}{c}{\textbf{Width CI90}} & \multicolumn{2}{c}{\textbf{Width PI90}}  \\
& BDE & DE & BDE & DE & BDE/DE & BDE & DE & BDE & DE \\
& \multicolumn{2}{c|}{$\times 10^{-2}$} & \multicolumn{2}{c|}{$\times 10^{-3}$} & & & & &  \\
\midrule	
Boston  & \textbf{8.0}  & 11 & \textbf{8.9} & 13 & 3.82 & 4.32 & 3.88 & 11.2 & 10.7 \\
Concrete & \textbf{4.1} & 8.6 & \textbf{3.9} & 6.8 & 10.3 & 10.7 & 8.13 & 30.6 & 29.0 \\
Energy & \textbf{9.0} & 12 & \textbf{6.2} & 8.3 & 2.87 & 2.98 & 2.43 & 4.98 & 4.73 \\
\bottomrule 
\end{tabular}
\end{center}
\end{table*}

\subsection{Different simulation method}
\noindent Instead of a random forest, we used a neural network in order to simulate data (see Algorithm \ref{alg: nnalgo}). The network has the same architecture and training procedure as the ones used for the experiment.

We see in \cref{bench-neural-network} that we get better results for both BDE and DE, although BDE still perform better. A likely explanation is that this task is easier, as is indicated by the significantly lower RMSE. The model we are simulating targets from is identical to the model we are using for the experiment. 

\begin{algorithm}[h]
      \caption{Pseudo-code to simulate data based on a real-world data set $\mathcal{D}$ using a neural network.}
   \label{alg: nnalgo}
\begin{algorithmic}[1]
   \STATE Train a neural network on $\mathcal{D}$ that outputs $f(\bm{x})$ and $\sigma^{2}(\bm{x})$
   \STATE Simulate new targets: $y_{\text{new}} \sim \N{f(\bm{x})}{\sigma^{2}(\bm{x})}$
   \STATE \textbf{Return:} $\mathcal{D}_{\text{new}} = (X, Y_{\text{new}})$
\end{algorithmic}
\end{algorithm}

\begin{table*}[h]
\caption{Results of the comparison of our method Bootstrapped Deep Ensembles (BDE) to Deep Ensembles (DE) when using a neural network to simulate data. A total of 100 simulated data sets were used to calculate the metrics. On each new data set, new ensembles were trained, and new confidence and prediction intervals were constructed. Brier-CI90 denotes the Brier score of the CICF of a 90$\%$ confidence interval. Brier-PI90 denotes the equivalent quantity for the PICF. RMSE gives the root mean squared error of the predictions with respect to the targets. Since the predictor is identical for both methods}
\label{bench-neural-network}
\begin{center}
\begin{tabular}{lcc|cc|c|cc|cc}
\toprule
 \textbf{SIMULATION} & \multicolumn{2}{c}{\textbf{Brier-CI90} \textcolor{green}{$\downarrow$}} & \multicolumn{2}{c}{\textbf{Brier-PI90} \textcolor{green}{$\downarrow$}} &  \multicolumn{1}{c}{\textbf{RMSE}} & \multicolumn{2}{c}{\textbf{Width CI90}} & \multicolumn{2}{c}{\textbf{Width PI90}}  \\
& BDE & DE & BDE & DE & BDE/DE & BDE & DE & BDE & DE \\
& \multicolumn{2}{c|}{$\times 10^{-2}$} & \multicolumn{2}{c|}{$\times 10^{-3}$} & & & & &  \\
\midrule	
Boston  & \textbf{2.8} & 3.1 & \textbf{6.3}& 8.8  & 2.92 & 3.31 & 3.27 & 7.86 &7.50 \\
Concrete  & \textbf{2.4} & 3.3 & \textbf{4.9} & 6.9 & 6.00  & 7.81 & 7.12 & 17.7 &16.6 \\
Energy & \textbf{3.7} & 3.8 & \textbf{8.5} & 8.9 & 2.64 & 3.10 & 1.99 & 5.82 & 5.53 \\
\bottomrule 
\end{tabular}
\end{center}
\end{table*}

\subsection{Different retraining fractions}
\noindent As expected we observe from \cref{bench-retrainingfractions} that the widths of the confidence intervals get larger with an increasing retraining fraction. Trivially when we setting the retraining fraction to 0 would yield zero variance and setting it to 1 would also capture the uncertainty due to random training.
\begin{table}[h]
\caption{The effect of the training fraction on bootstrapped DE. A total of 100 simulated data sets were used to calculate the metrics. On each new data set, new ensembles were trained, and new confidence and prediction intervals were constructed. Brier-CI80 denotes the Brier score of the CICF of an 80$\%$ confidence interval. Brier-PI80 denotes the equivalent quantity for the PICF}
\label{bench-retrainingfractions}
\vskip 0.15in
\begin{center}
\begin{tabular}{lcccr}
\toprule
\textbf{Retraining fraction} &\textbf{Brier-CI80} \textcolor{green}{$\downarrow$}& \textbf{Brier-PI80} \textcolor{green}{$\downarrow$} & \textbf{Width CI80} & \textbf{Width PI80}  \\
 & $\times 10^{-2}$ &$\times 10^{-2}$&& \\
\midrule
0.1  & 11.5  & 3.0  & 2.84  & 7.37  \\
0.2  & 9.1  & 2.6 & 3.16 & 7.54 \\
0.3  & 7.8 & 2.5 & 3.4 & 7.64 \\
0.4  & 6.1 & 2.1 & 3.7 & 7.89 \\
\bottomrule
\end{tabular}
\end{center}
\vskip -0.1in
\end{table}

\subsection{RMSE values of Experiment 1}
\cref{RMSEvalues} gives the RMSE values of all methods during Experiment 1 of the main text.
\begin{table}[h!]
\caption{The RMSE values of the methods during the simulations of Experiment 1. Each value is calculated with respect to the targets of the test set and are averaged over the 100 simulations that were used for each data set. Bootstrapped Deep Ensembles and Deep Ensembles have the same score since they use they exact same predictor. Concrete Dropout has a very comparable score. The Naive Bootstrap has significantly larger errors. This is to be expected since each ensemble member is effectively being trained on less data due to the resampling.}
\label{RMSEvalues}
\vskip 0.1in
\begin{center}
\begin{tabular}{lccc}
\toprule
\textbf{SIMULATION}  & (Bootstrapped) Deep Ensembles  & Naive Bootstrap & Concrete Dropout \\
\midrule	
Boston    & 3.92 & 4.03& 3.95 \\
Concrete  & 10.4& 12.5& 10.8\\
Energy & 2.73& 3.38& 2.67 \\
Kin8nm &  0.22& 0.29& 0.24 \\
Naval  &0.013& 0.018& 0.013\\
Power-Plant & 5.01& 5.66& 5.08\\
Yacht & 3.26&3.86&2.68\\
Wine & 0.65& 0.68& 0.71 \\
\bottomrule 
\end{tabular}
\end{center}
\vskip -0.1in
\end{table}

\subsection{Additional plots Experiment 1} \label{additionalviolins}
\begin{figure}[h]
\centering
\subfigure[Boston, $\alpha=0.05$]{\includegraphics[width=.24\textwidth]{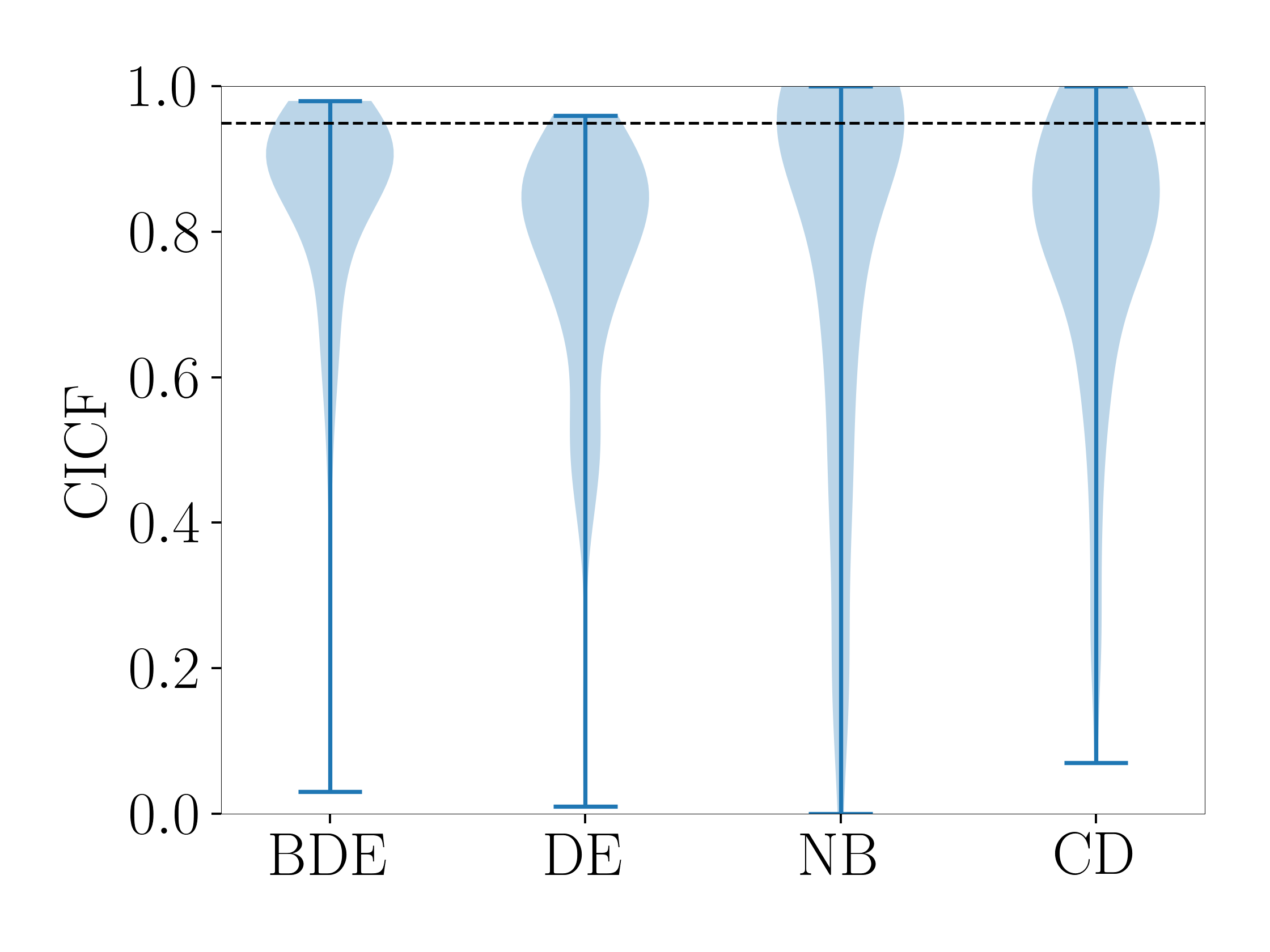}}
\subfigure[Boston, $\alpha=0.2$]{\includegraphics[width=.24\textwidth]{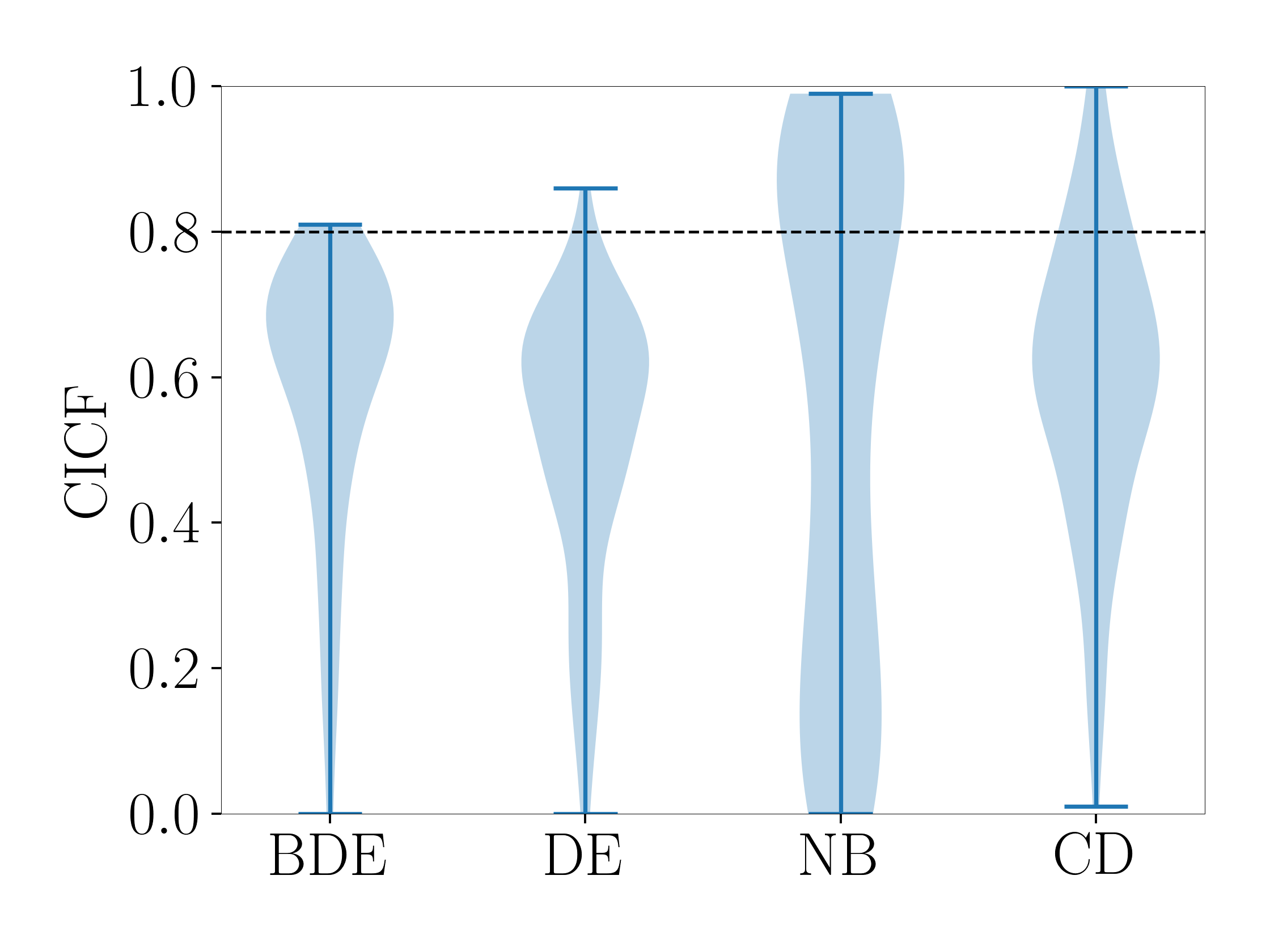}}
\subfigure[Concrete, $\alpha=0.05$]{\includegraphics[width=.24\textwidth]{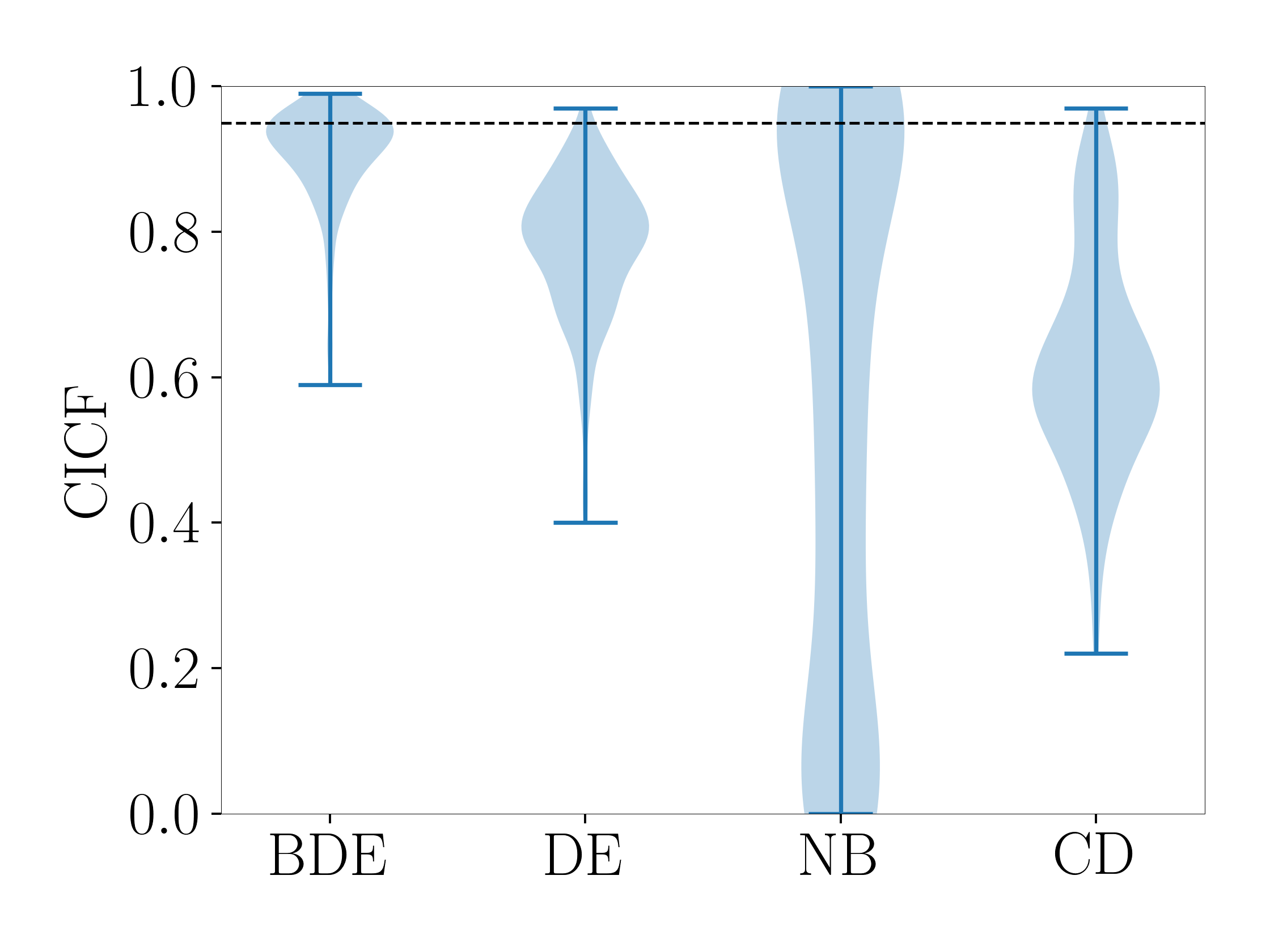}}
\subfigure[Concrete, $\alpha=0.2$]{\includegraphics[width=.24\textwidth]{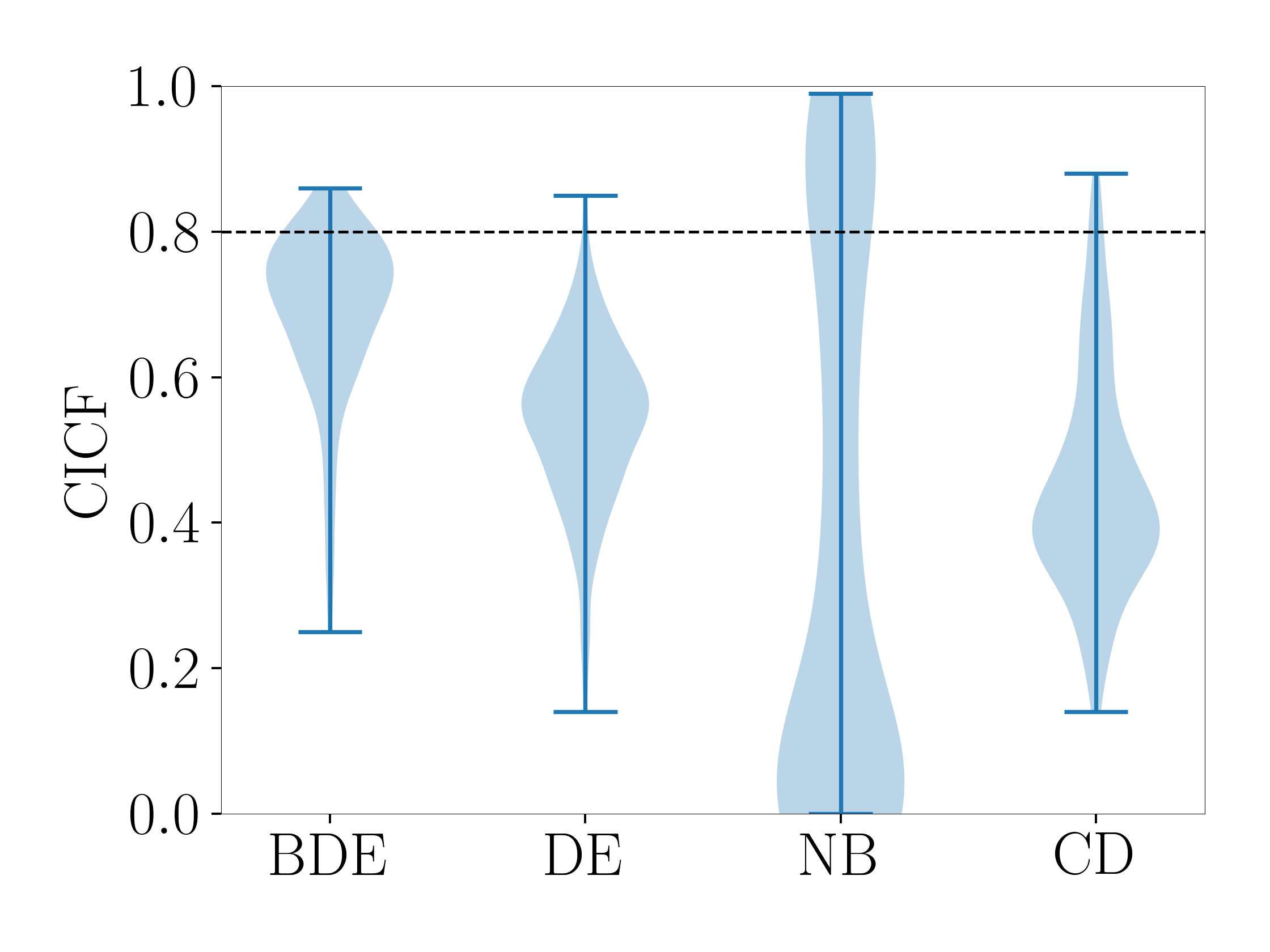}}
\subfigure[Energy, $\alpha=0.05$]{\includegraphics[width=.24\textwidth]{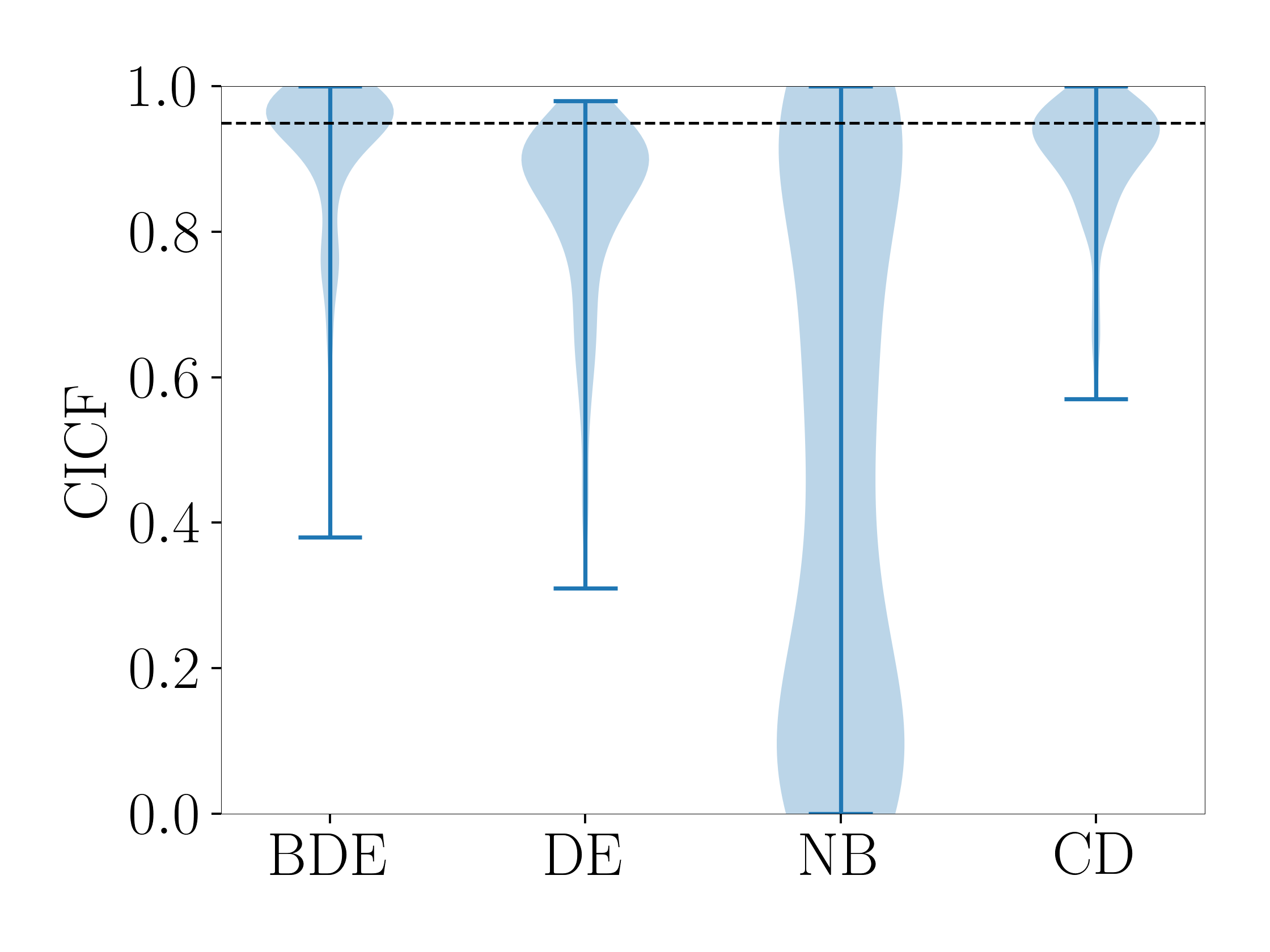}}
\subfigure[Energy, $\alpha=0.2$]{\includegraphics[width=.24\textwidth]{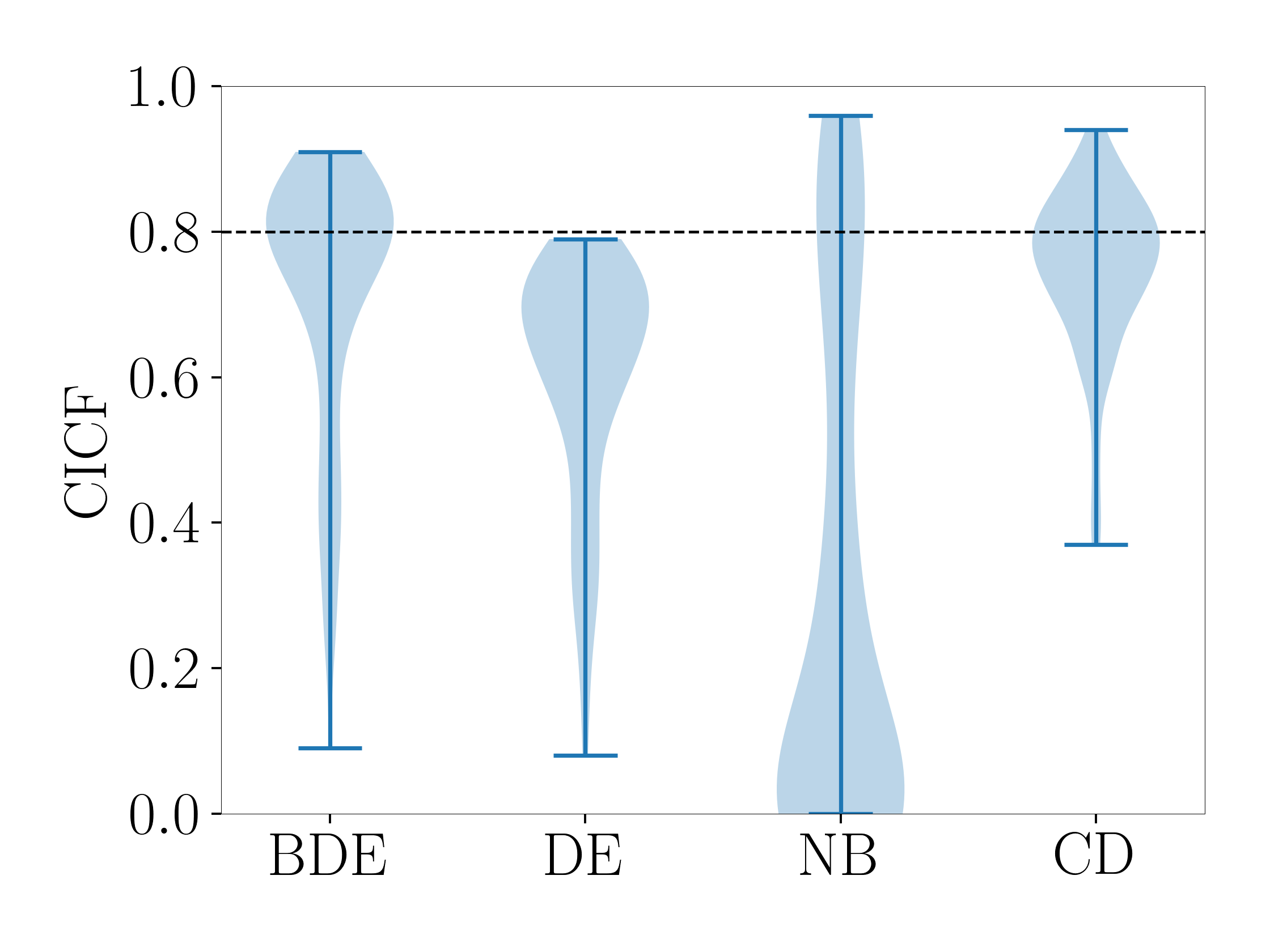}}
\subfigure[kin8nm, $\alpha=0.05$]{\includegraphics[width=.24\textwidth]{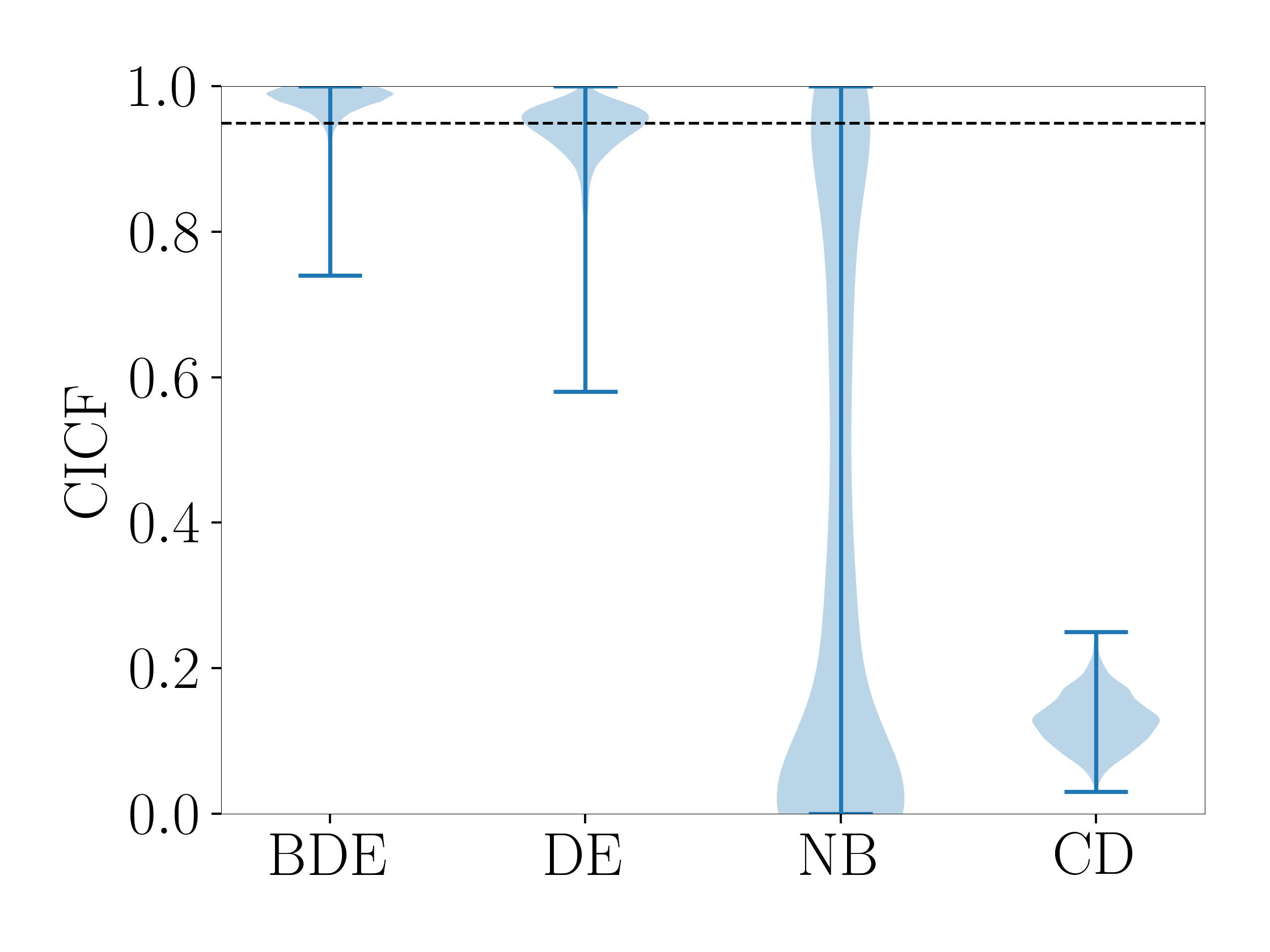}}
\subfigure[kin8nm, $\alpha=0.2$]{\includegraphics[width=.24\textwidth]{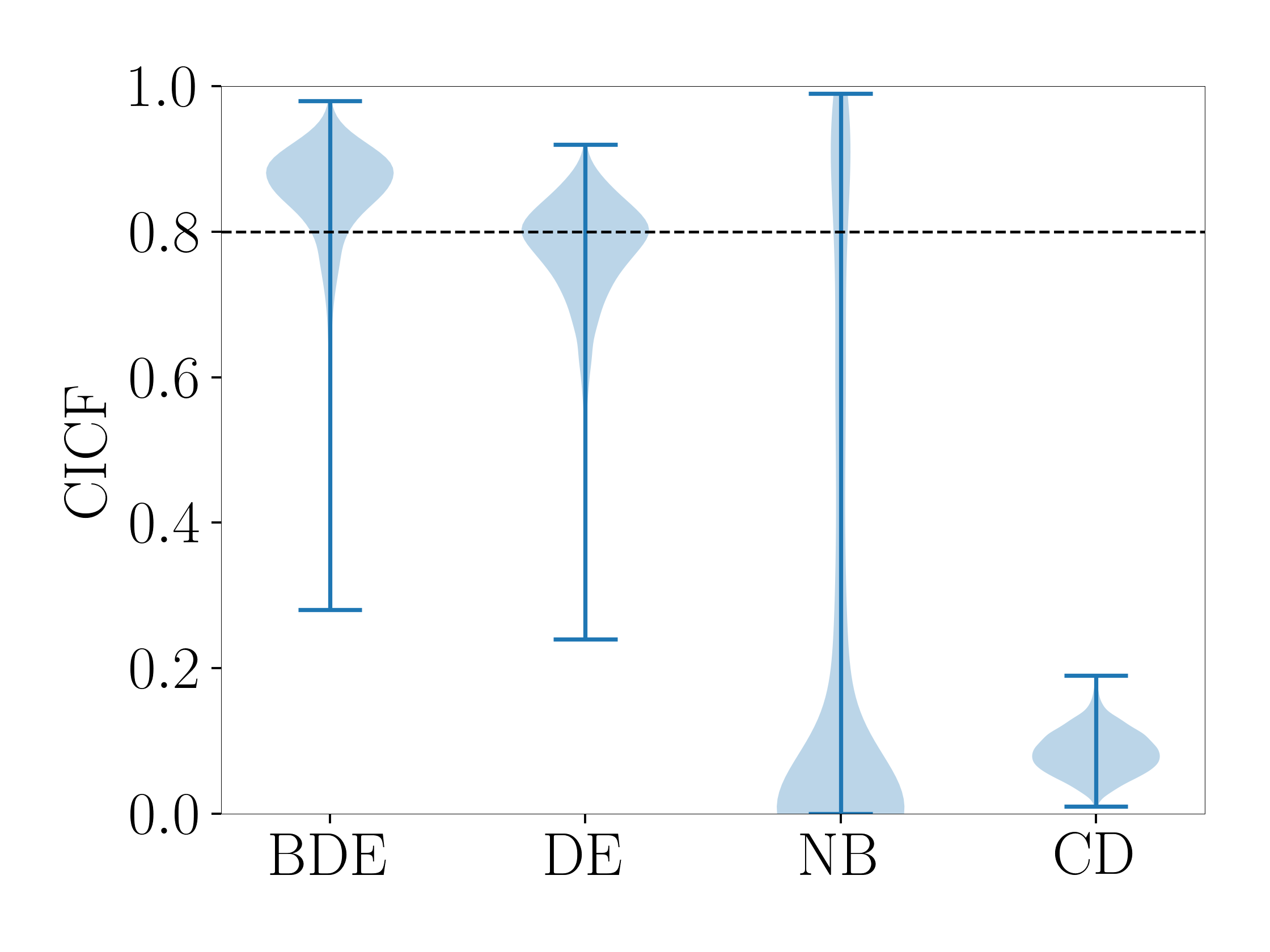}}
\subfigure[Naval, $\alpha=0.05$]{\includegraphics[width=.24\textwidth]{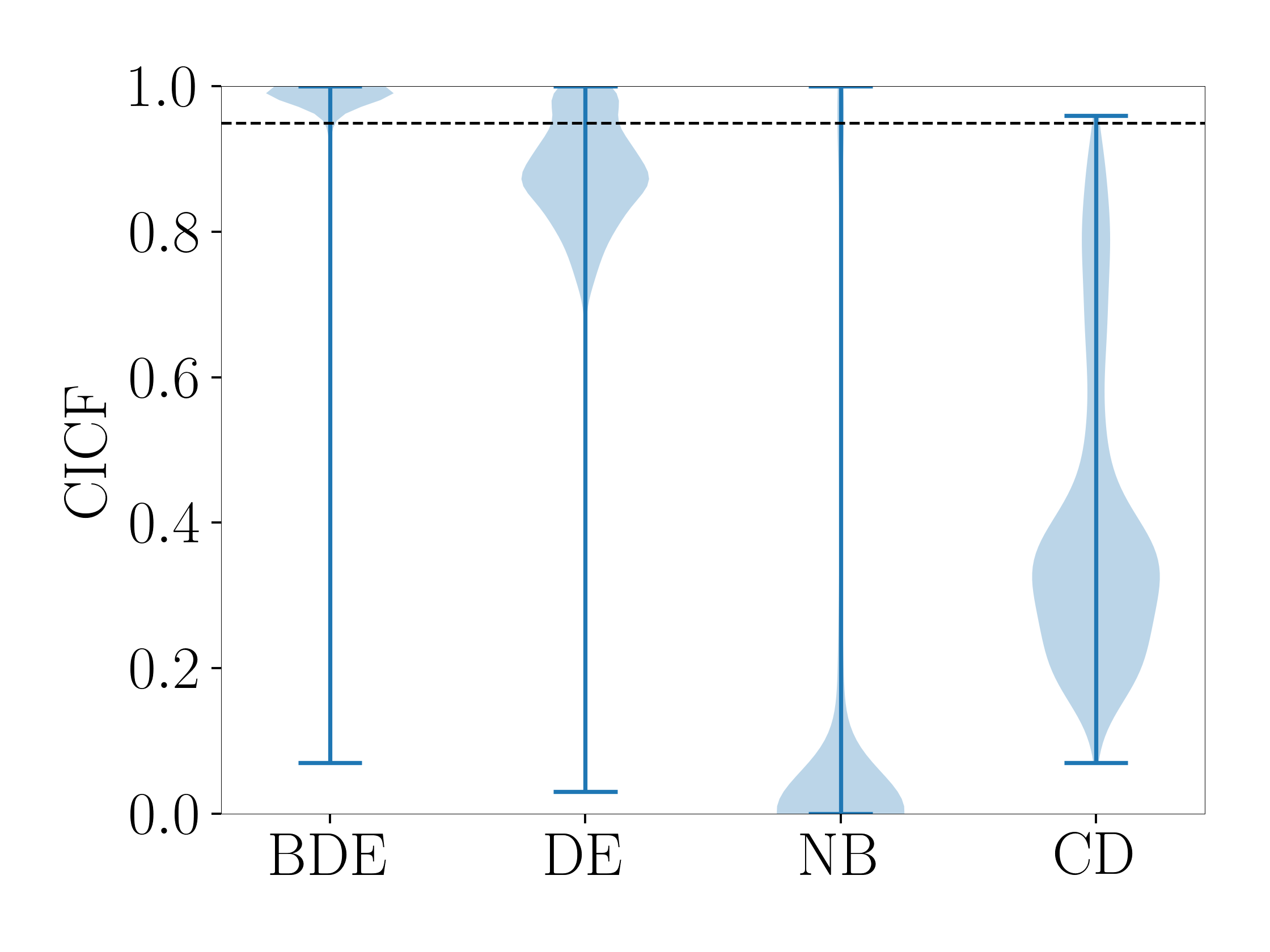}}
\subfigure[Naval, $\alpha=0.2$]{\includegraphics[width=.24\textwidth]{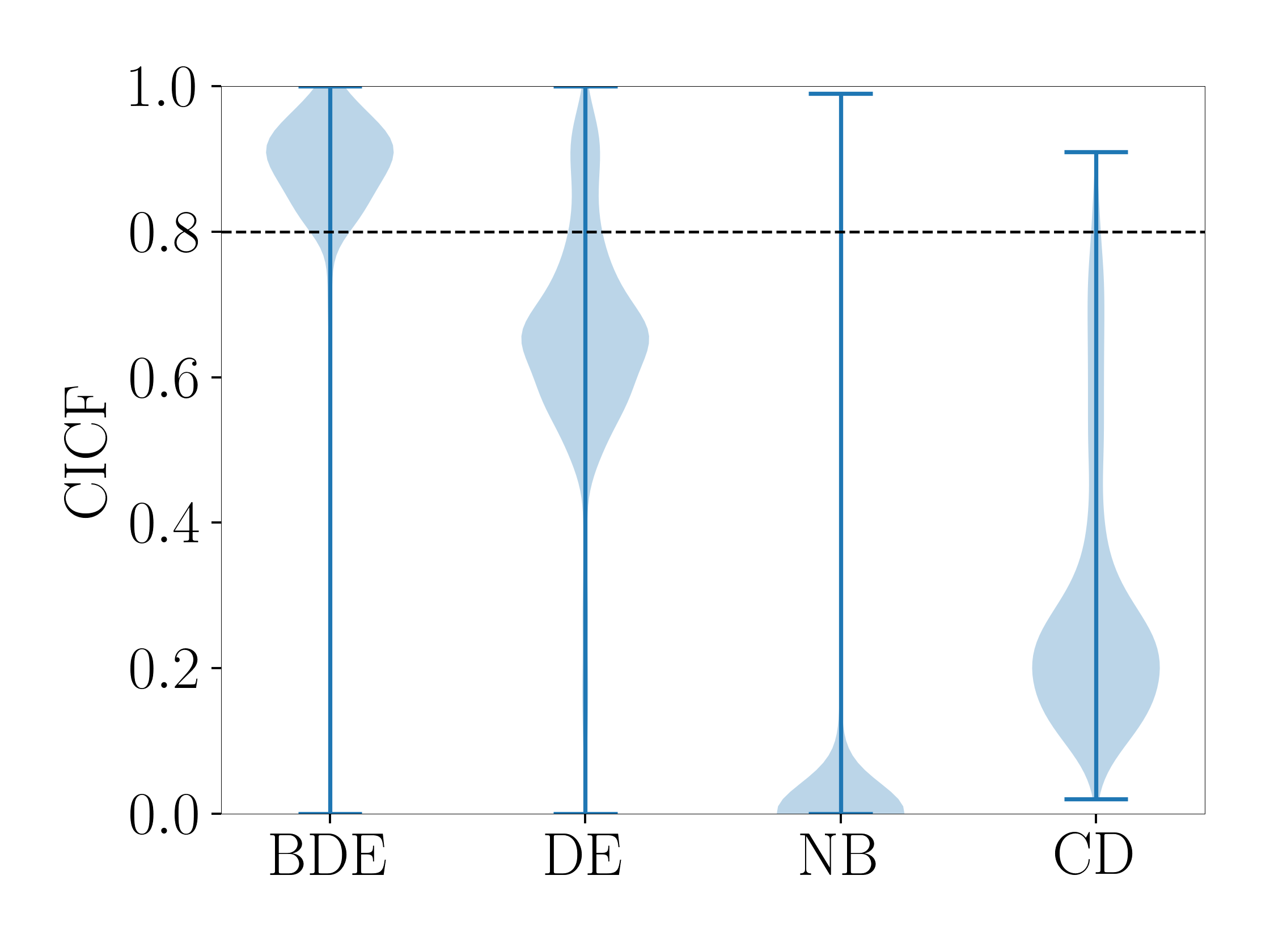}}
\subfigure[Power-plant, $\alpha=0.05$]{\includegraphics[width=.24\textwidth]{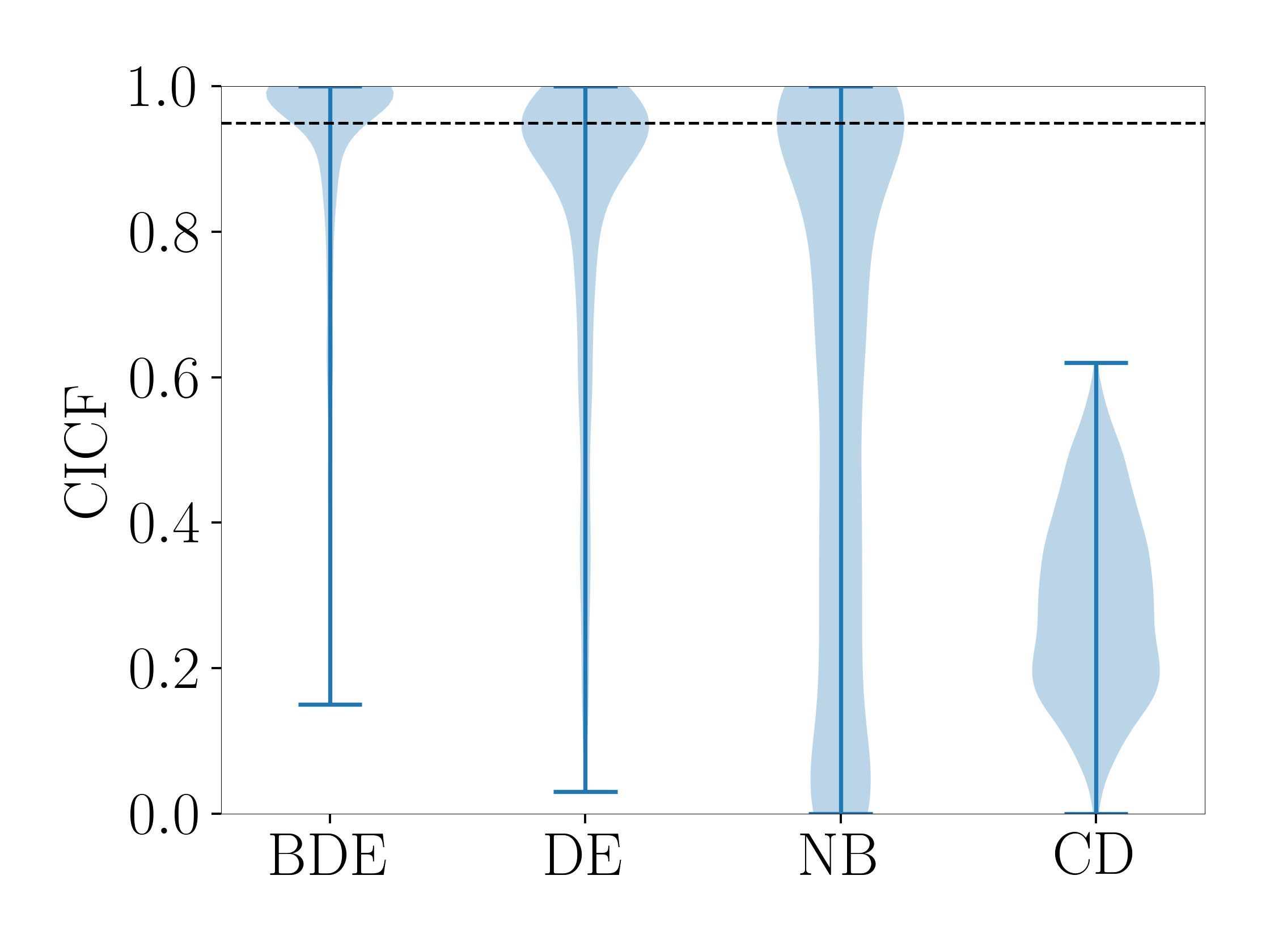}}
\subfigure[Power-plant, $\alpha=0.2$]{\includegraphics[width=.24\textwidth]{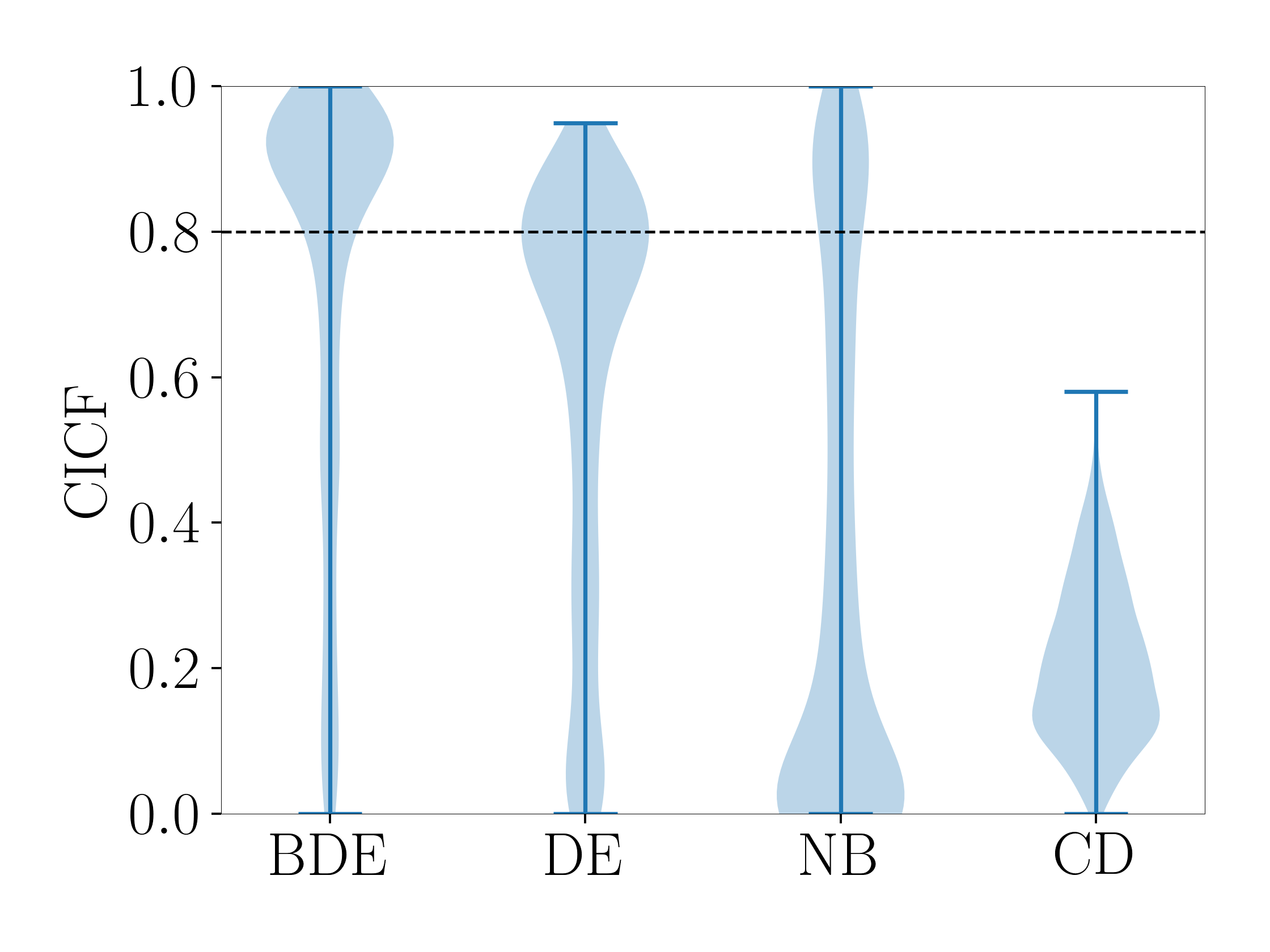}}
\subfigure[Wine, $\alpha=0.05$]{\includegraphics[width=.24\textwidth]{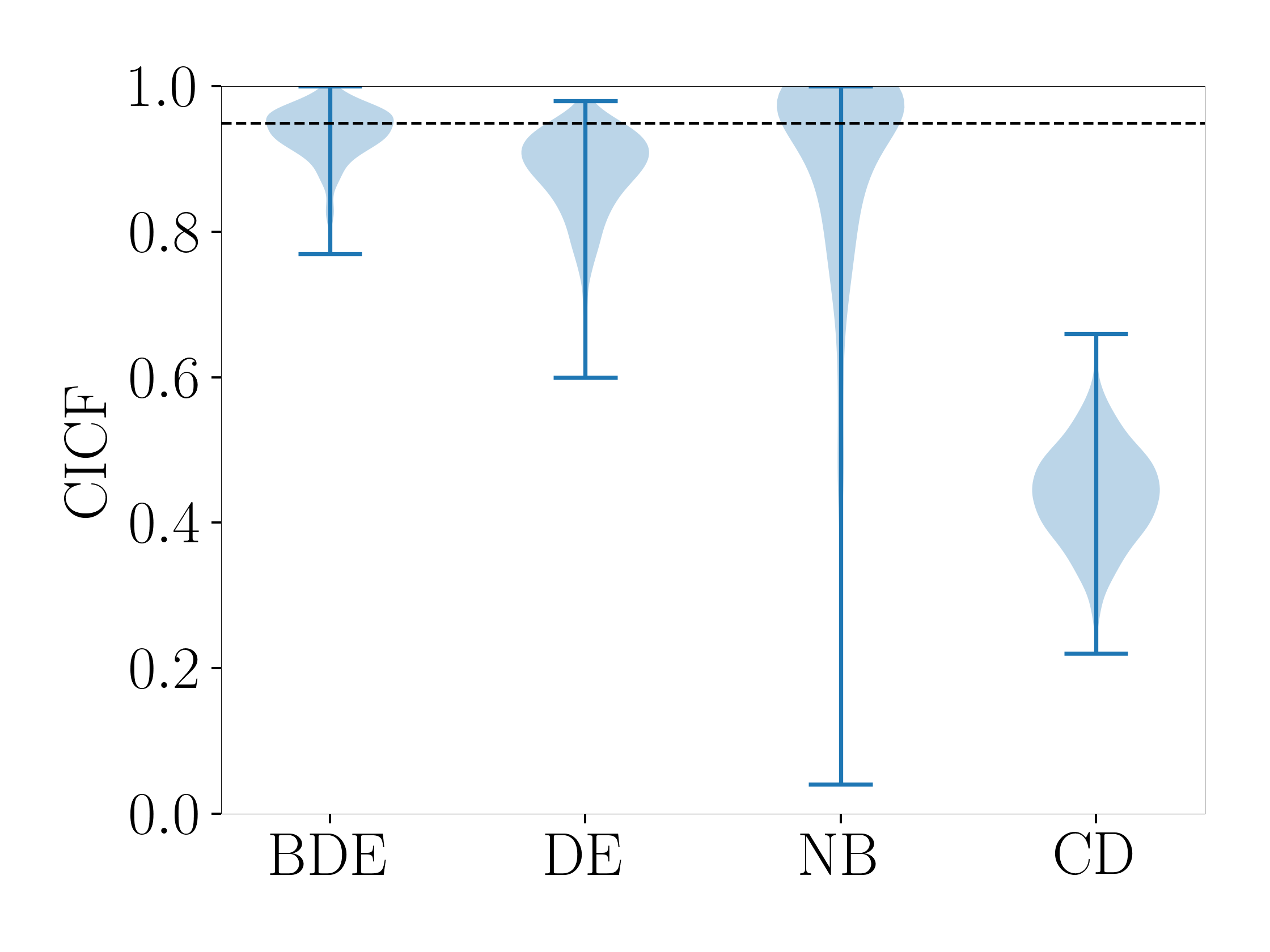}}
\subfigure[Wine, $\alpha=0.2$]{\includegraphics[width=.24\textwidth]{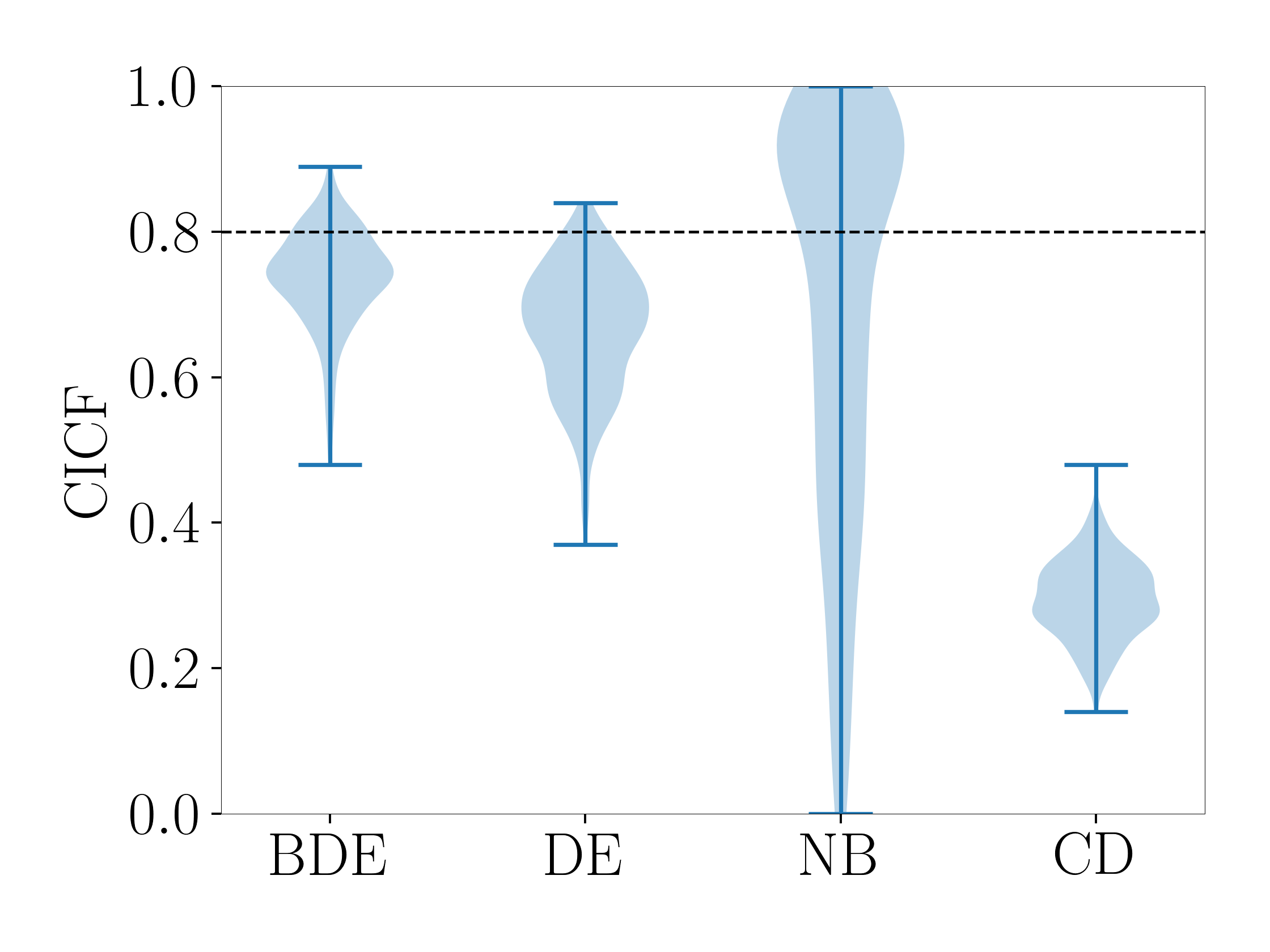}}
\subfigure[Yacht, $\alpha=0.05$]{\includegraphics[width=.24\textwidth]{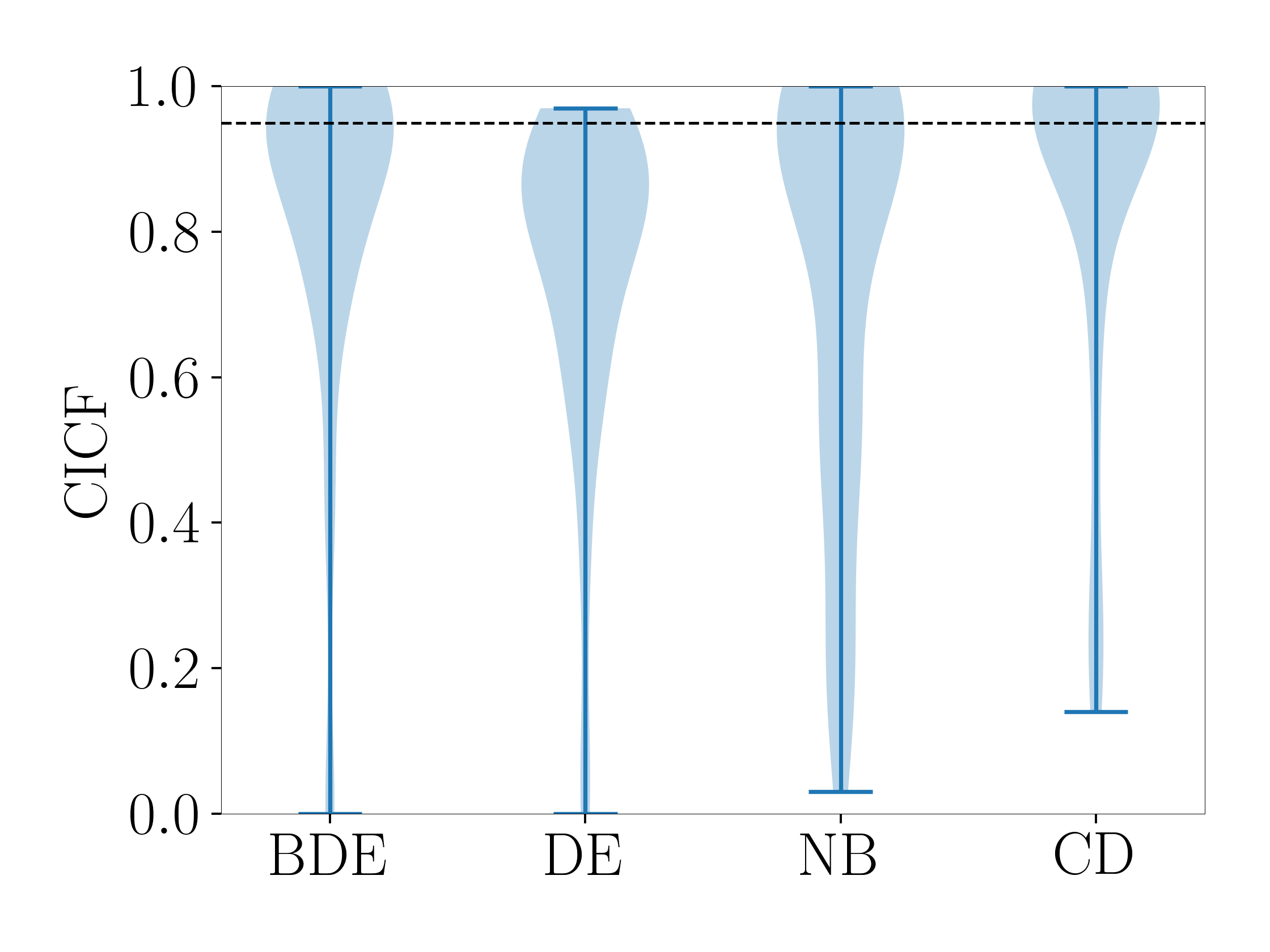}}
\subfigure[Yacht, $\alpha=0.2$]{\includegraphics[width=.24\textwidth]{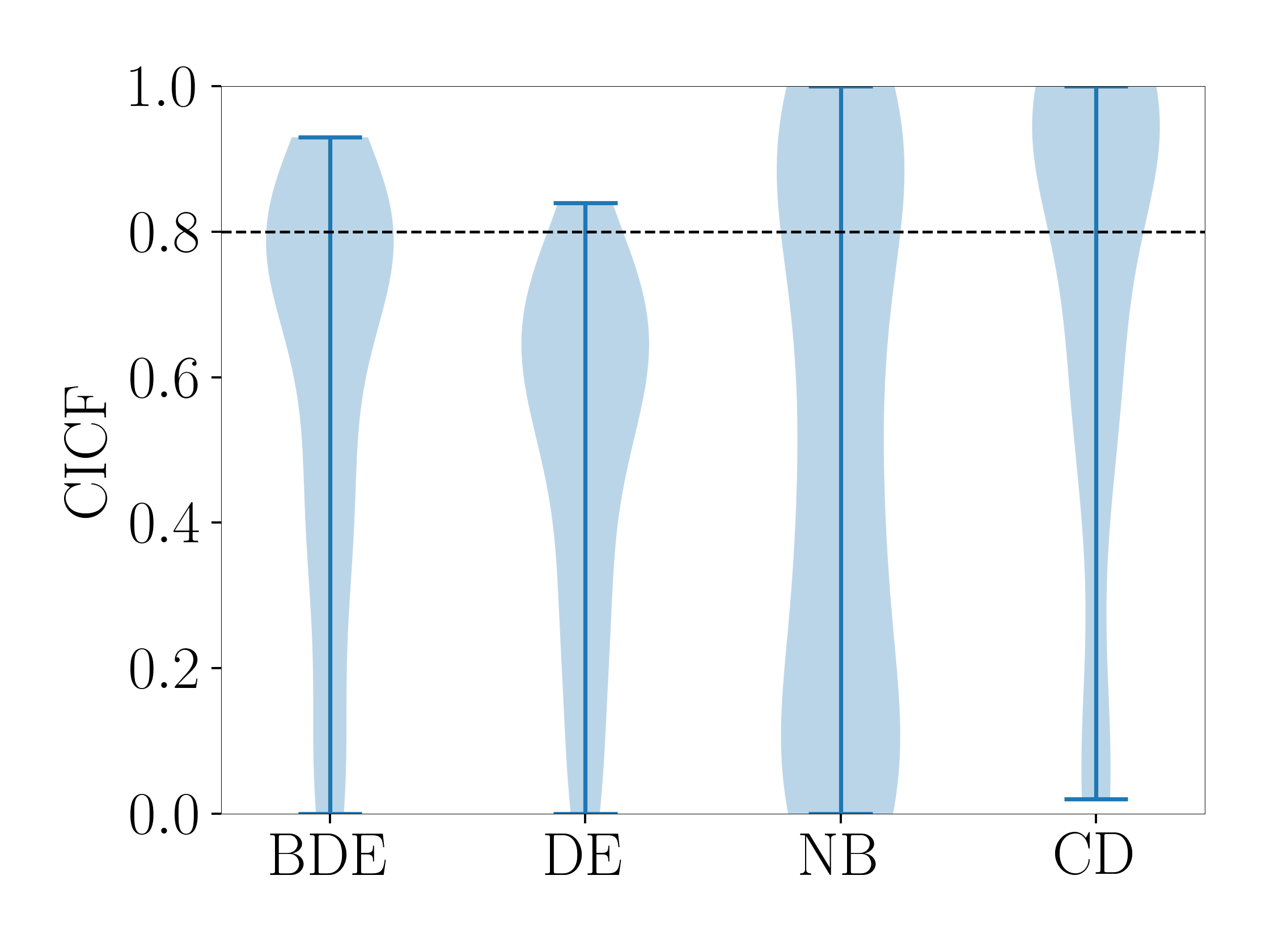}}

\caption{Violin plots of the CICF values for all 8 simulations of experiment 1 in the main text. For each simulation we give the CICF values for the $95\%$ and $80\%$ confidence intervals. Each plot is made using the CICF scores of each data point in the test sets. The CICF scores are calculated using 100 simulations. The confidence intervals of Bootstrapped Deep Ensembles have better coverage than the other methods in most simulations.}
\label{fig: violinsCICF}
\end{figure}

\begin{figure}[h!]
\centering
\vskip 0.2in
\subfigure[Boston, $\alpha=0.05$]{\includegraphics[width=.24\textwidth]{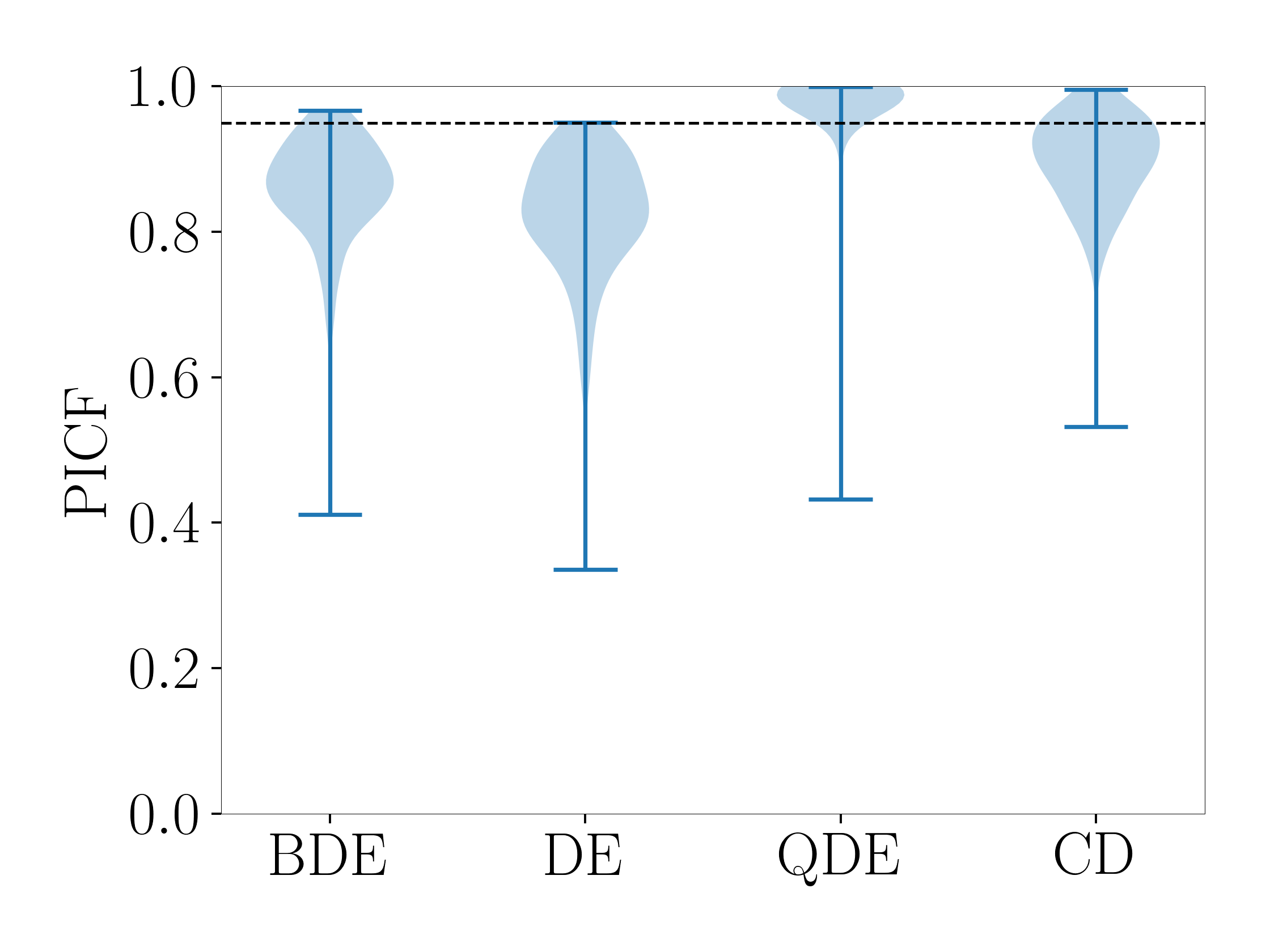}}
\subfigure[Boston, $\alpha=0.2$]{\includegraphics[width=.24\textwidth]{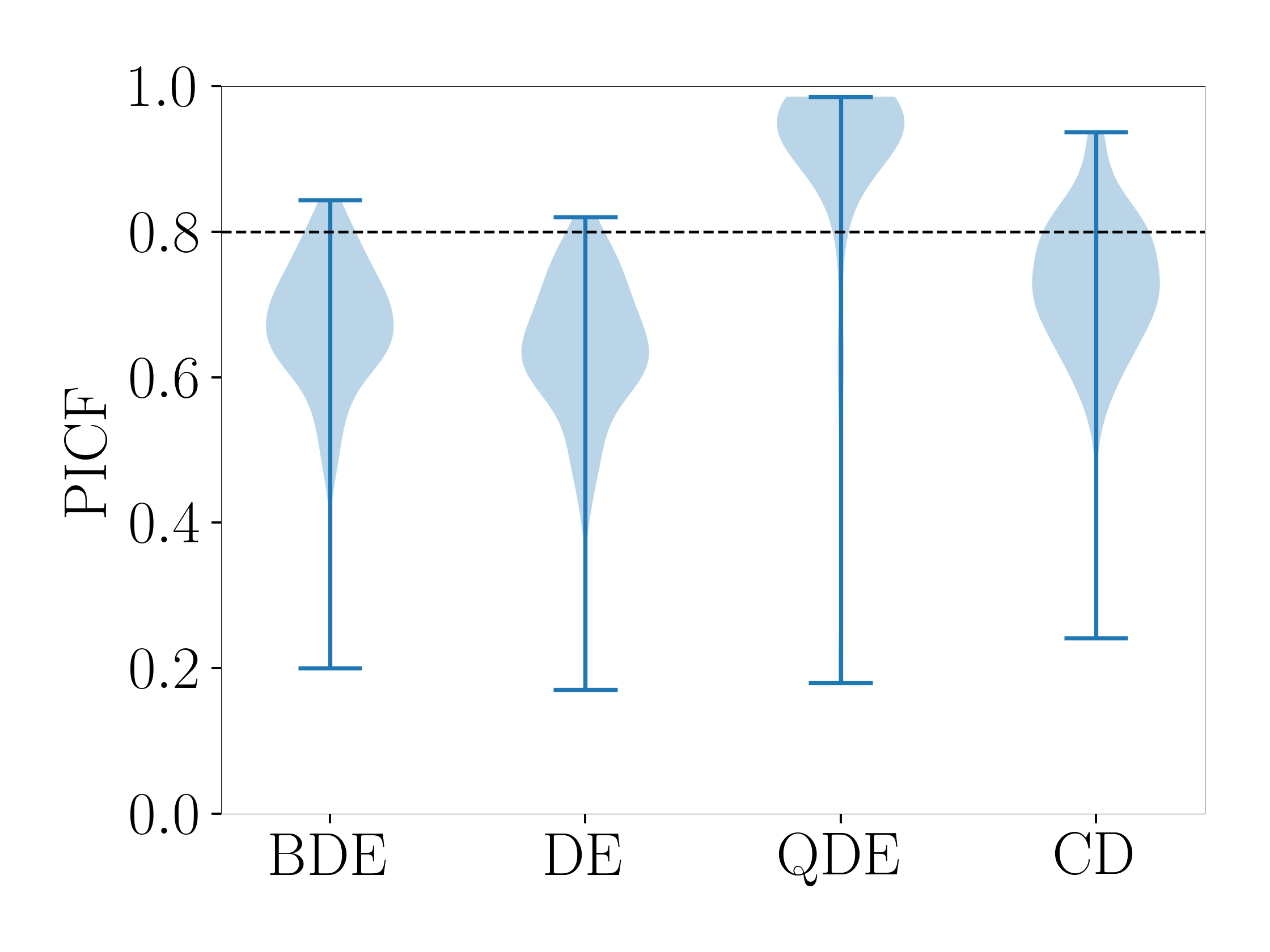}}
\subfigure[Concrete, $\alpha=0.05$]{\includegraphics[width=.24\textwidth]{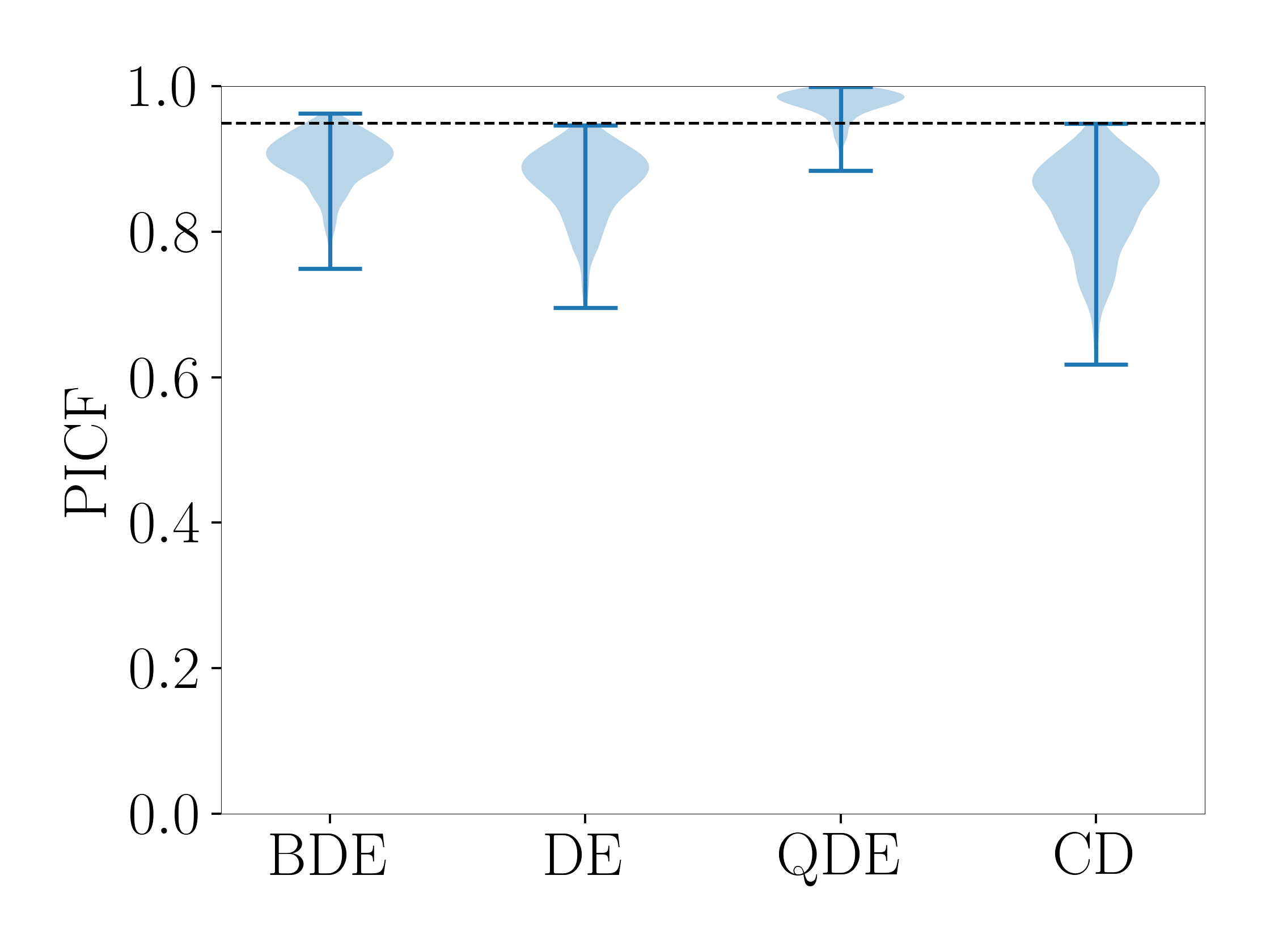}}
\subfigure[Concrete, $\alpha=0.2$]{\includegraphics[width=.24\textwidth]{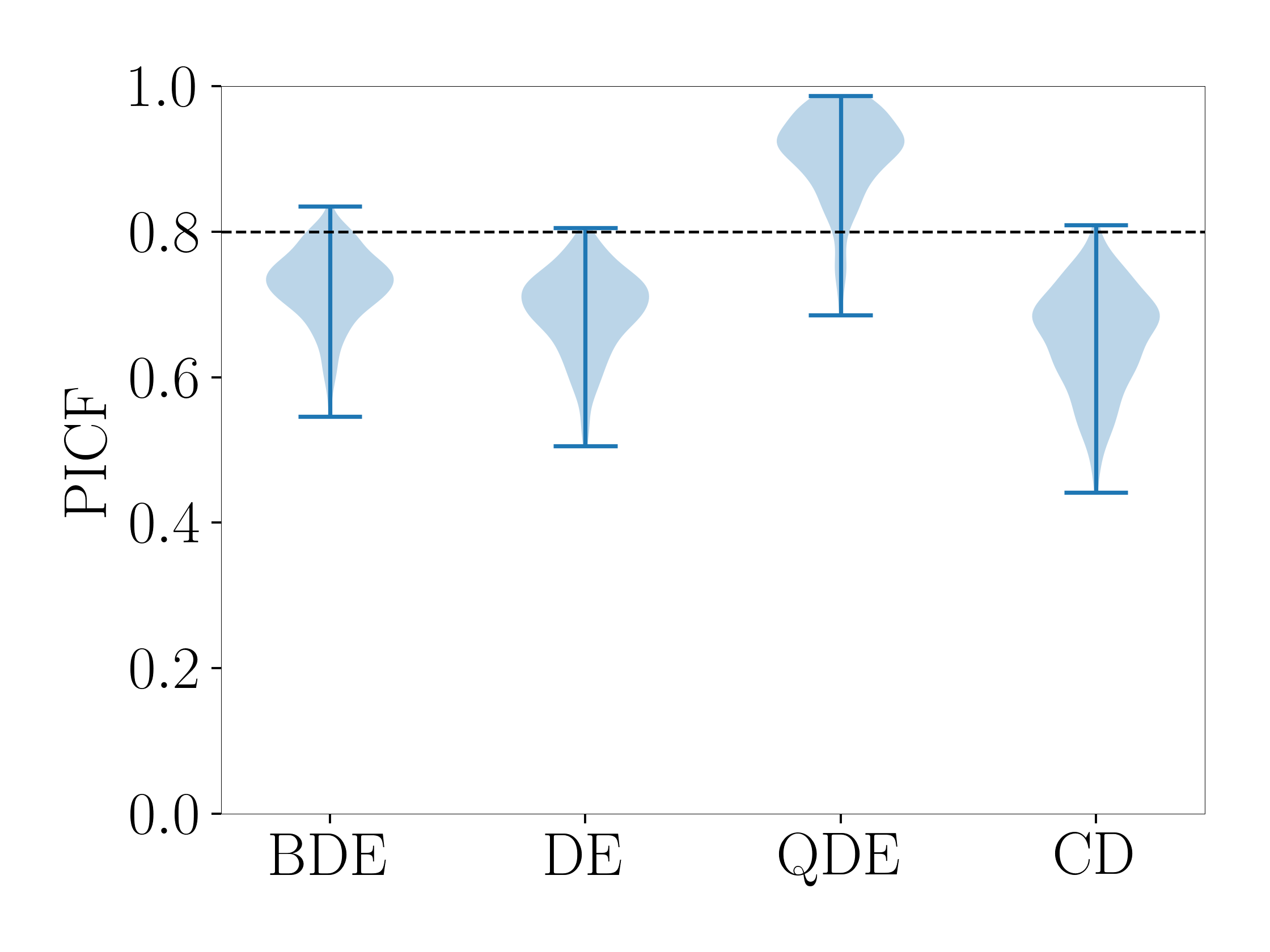}}
\subfigure[Energy, $\alpha=0.05$]{\includegraphics[width=.24\textwidth]{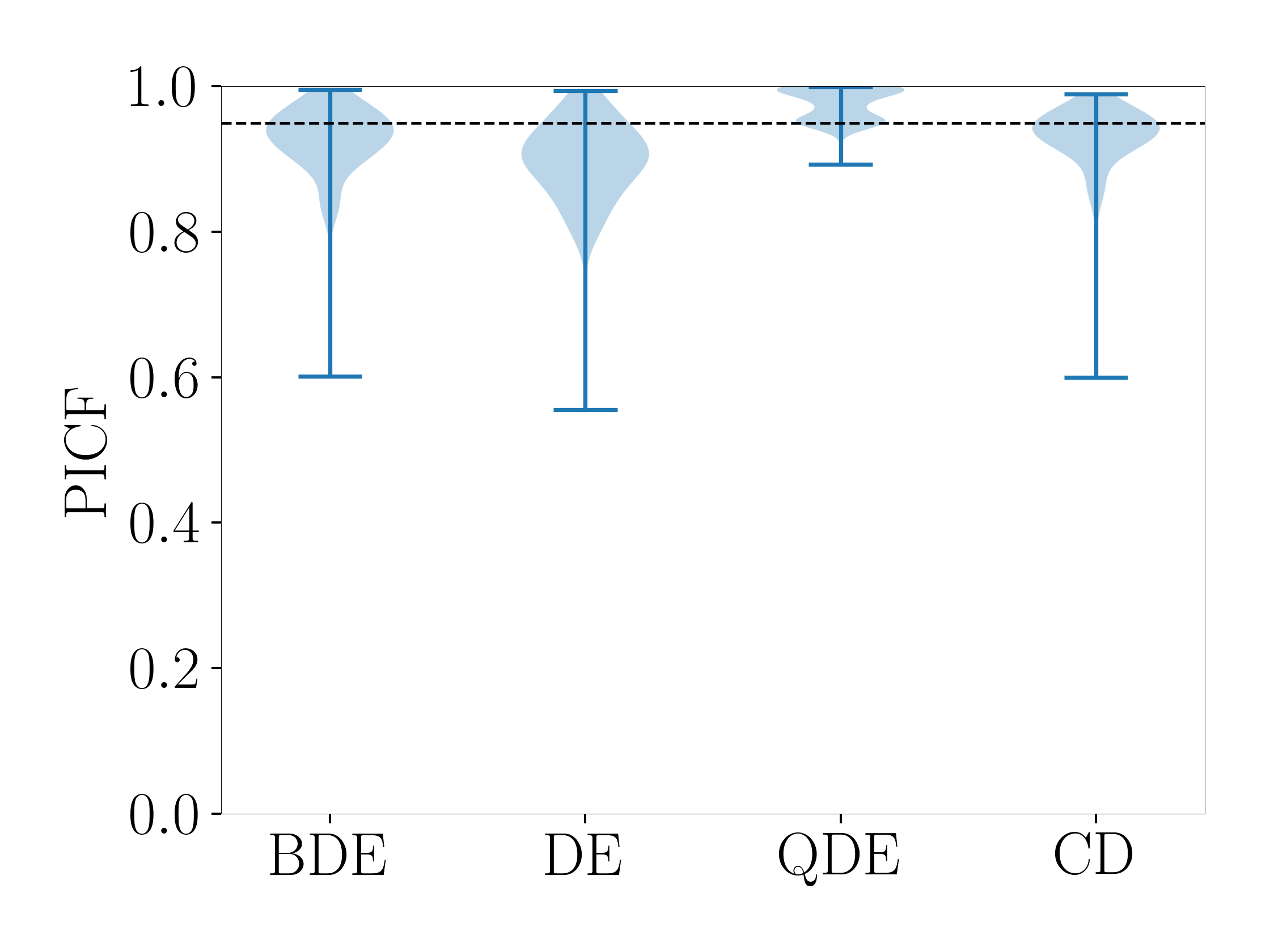}}
\subfigure[Energy, $\alpha=0.2$]{\includegraphics[width=.24\textwidth]{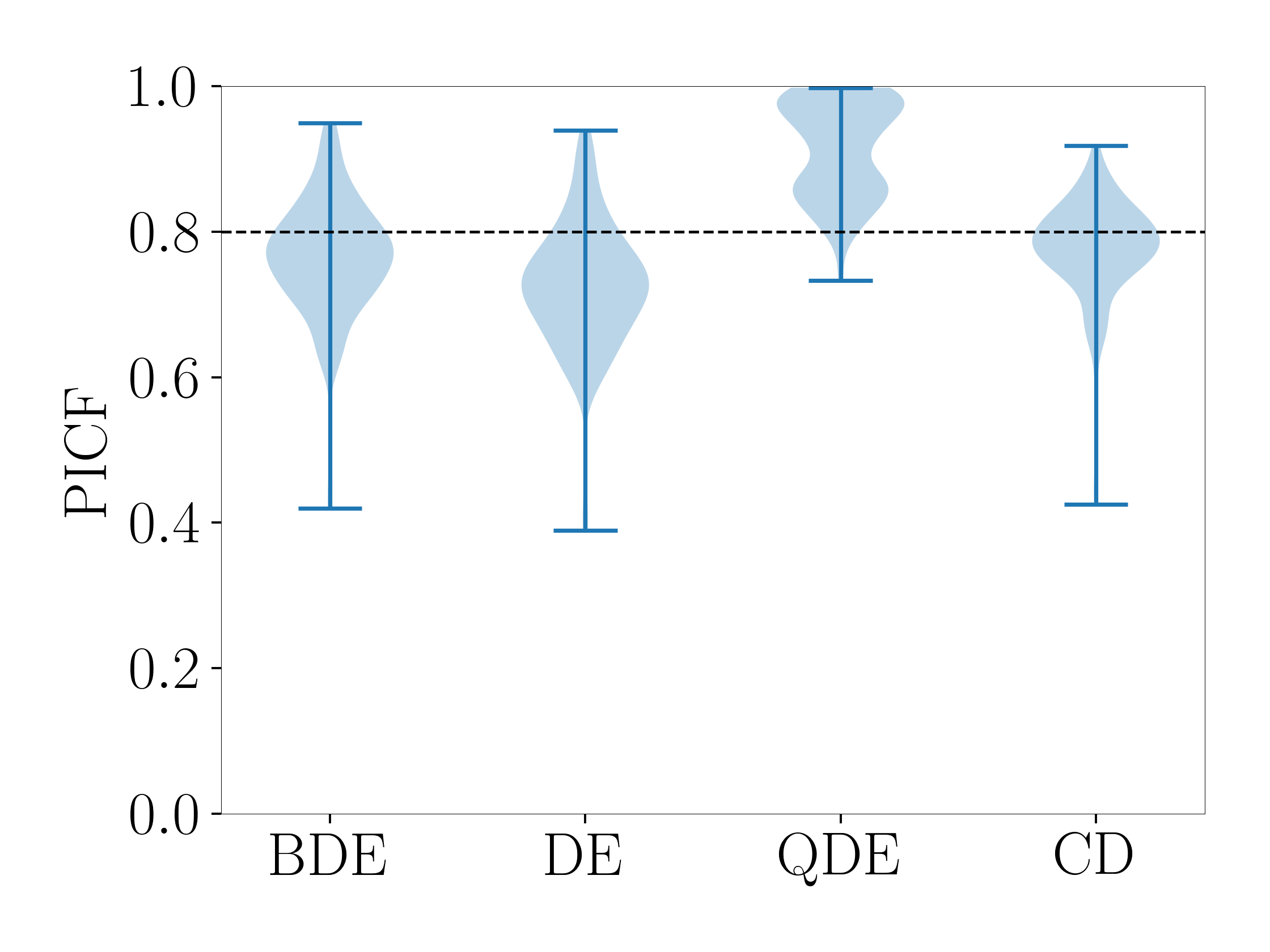}}
\subfigure[kin8nm, $\alpha=0.05$]{\includegraphics[width=.24\textwidth]{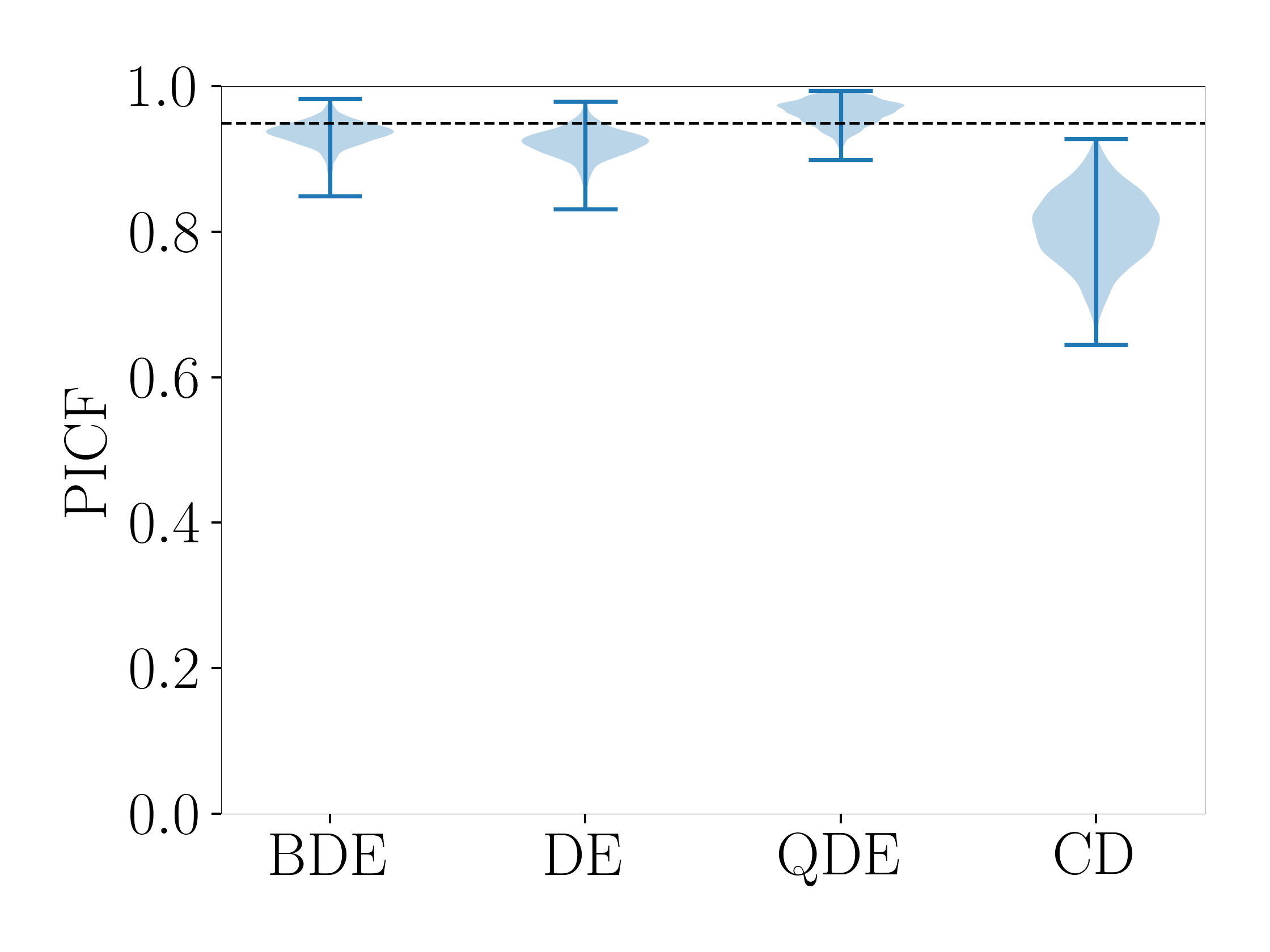}}
\subfigure[kin8nm, $\alpha=0.2$]{\includegraphics[width=.24\textwidth]{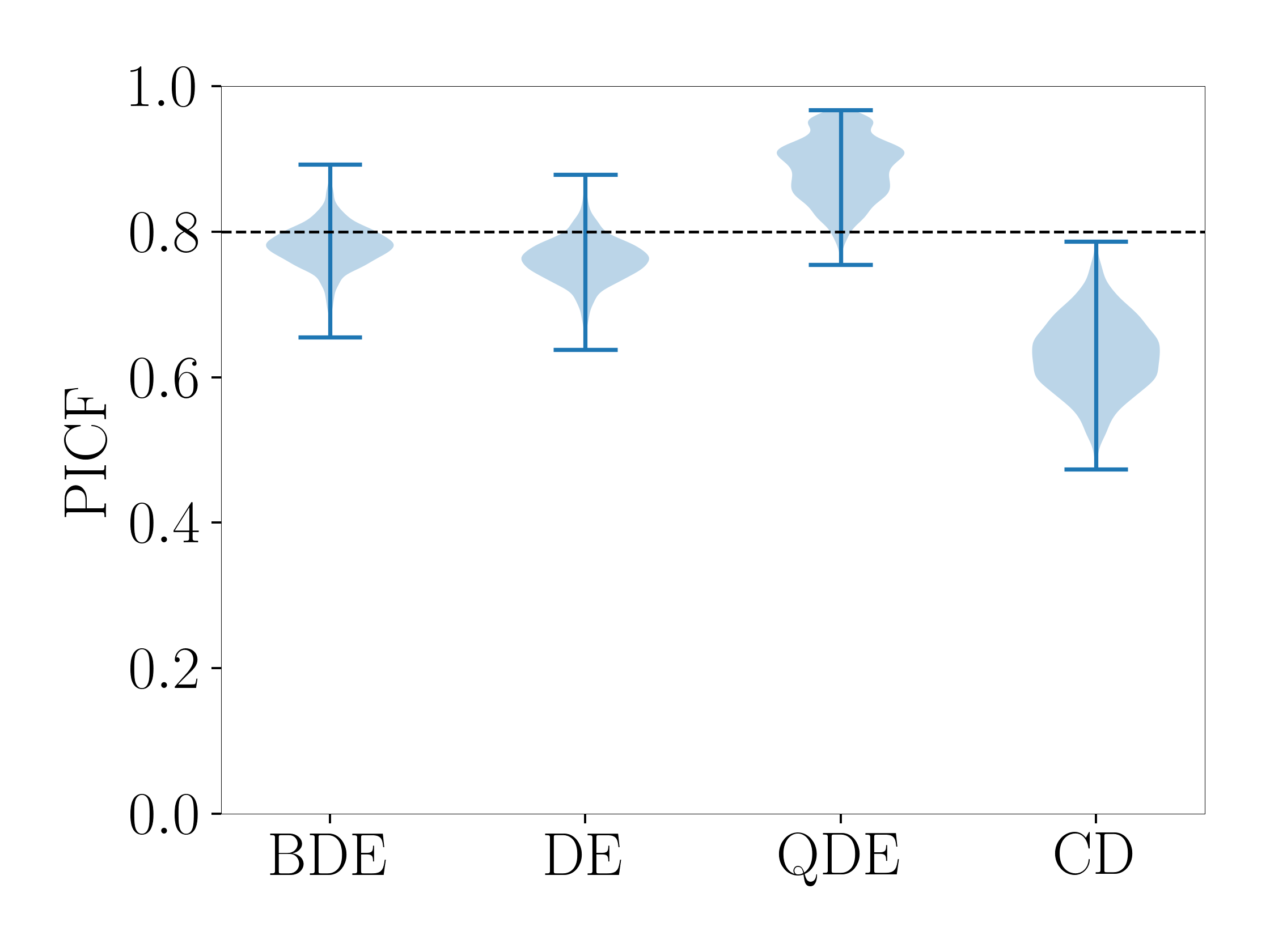}}
\subfigure[Naval, $\alpha=0.05$]{\includegraphics[width=.24\textwidth]{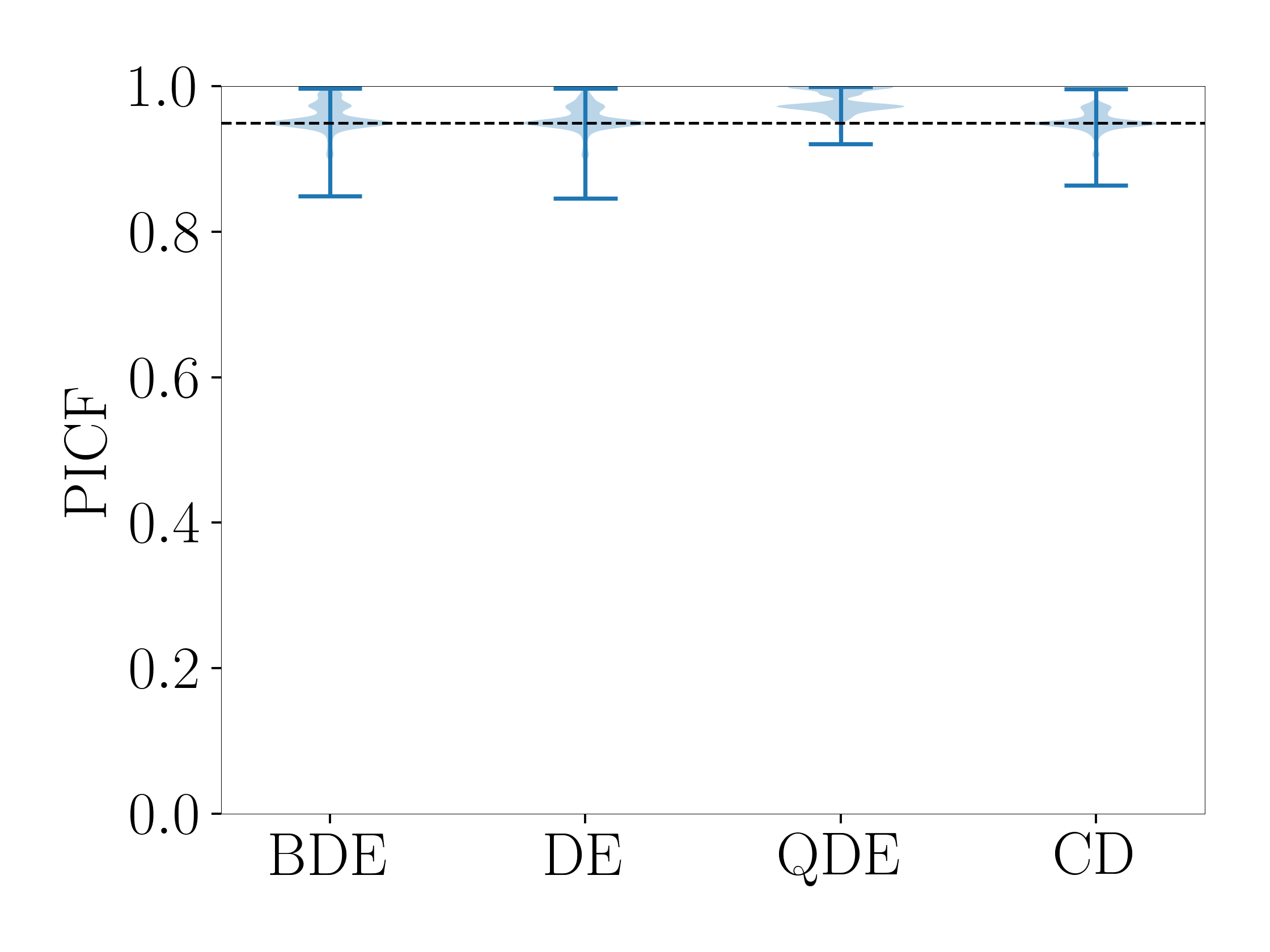}}
\subfigure[Naval, $\alpha=0.2$]{\includegraphics[width=.24\textwidth]{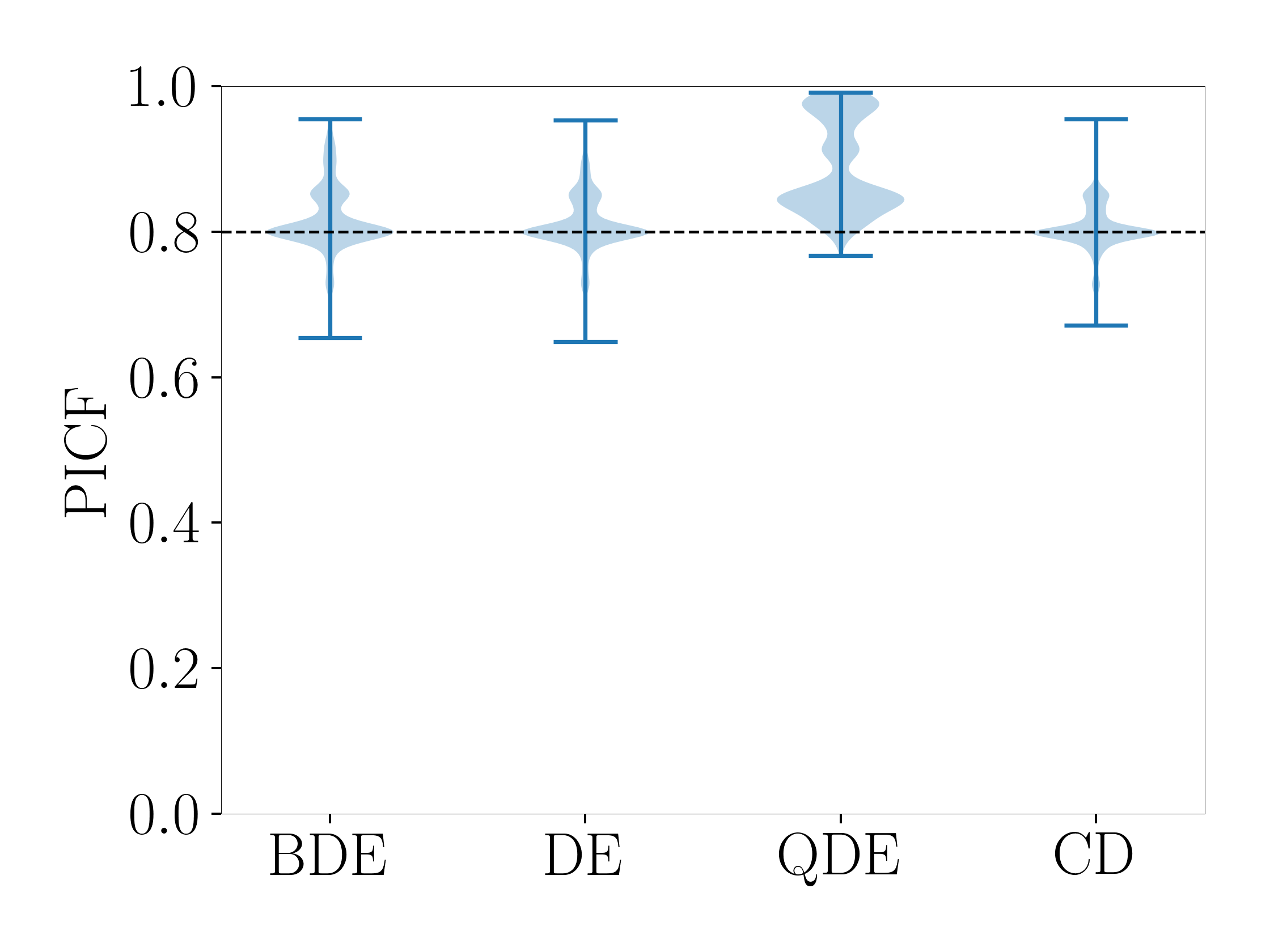}}
\subfigure[Power-plant, $\alpha=0.05$]{\includegraphics[width=.24\textwidth]{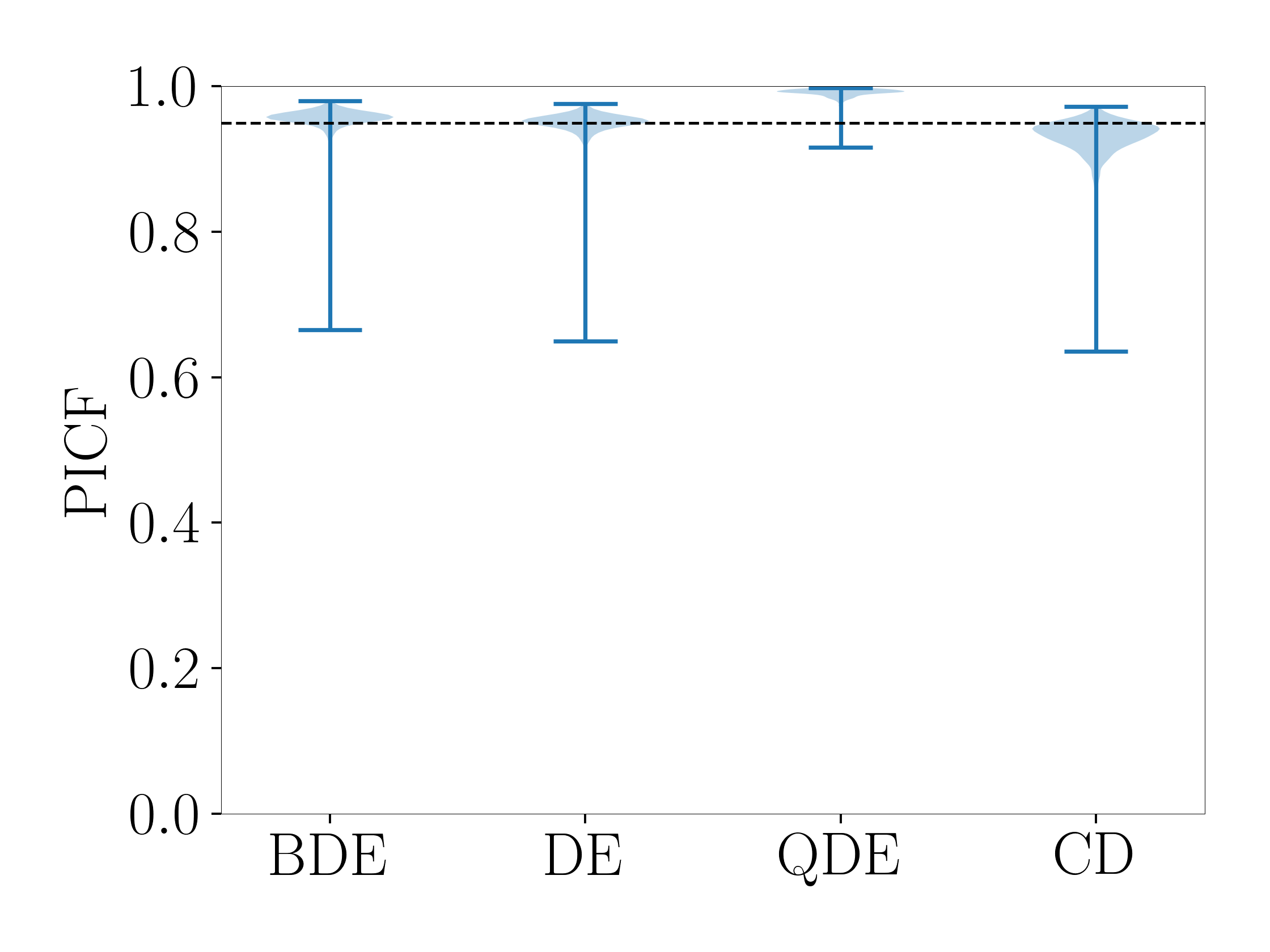}}
\subfigure[Power-plant, $\alpha=0.2$]{\includegraphics[width=.24\textwidth]{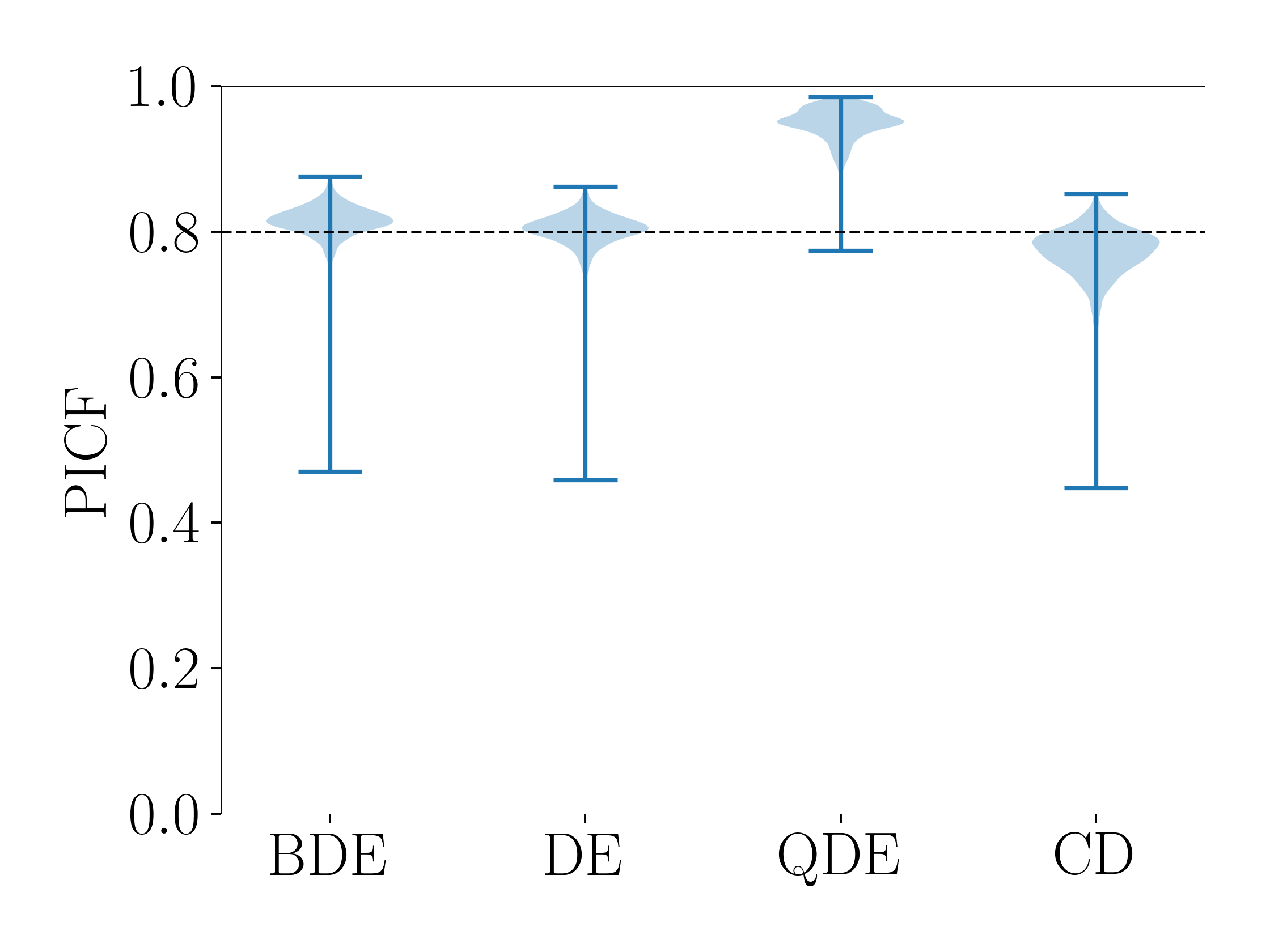}}
\subfigure[Wine, $\alpha=0.05$]{\includegraphics[width=.24\textwidth]{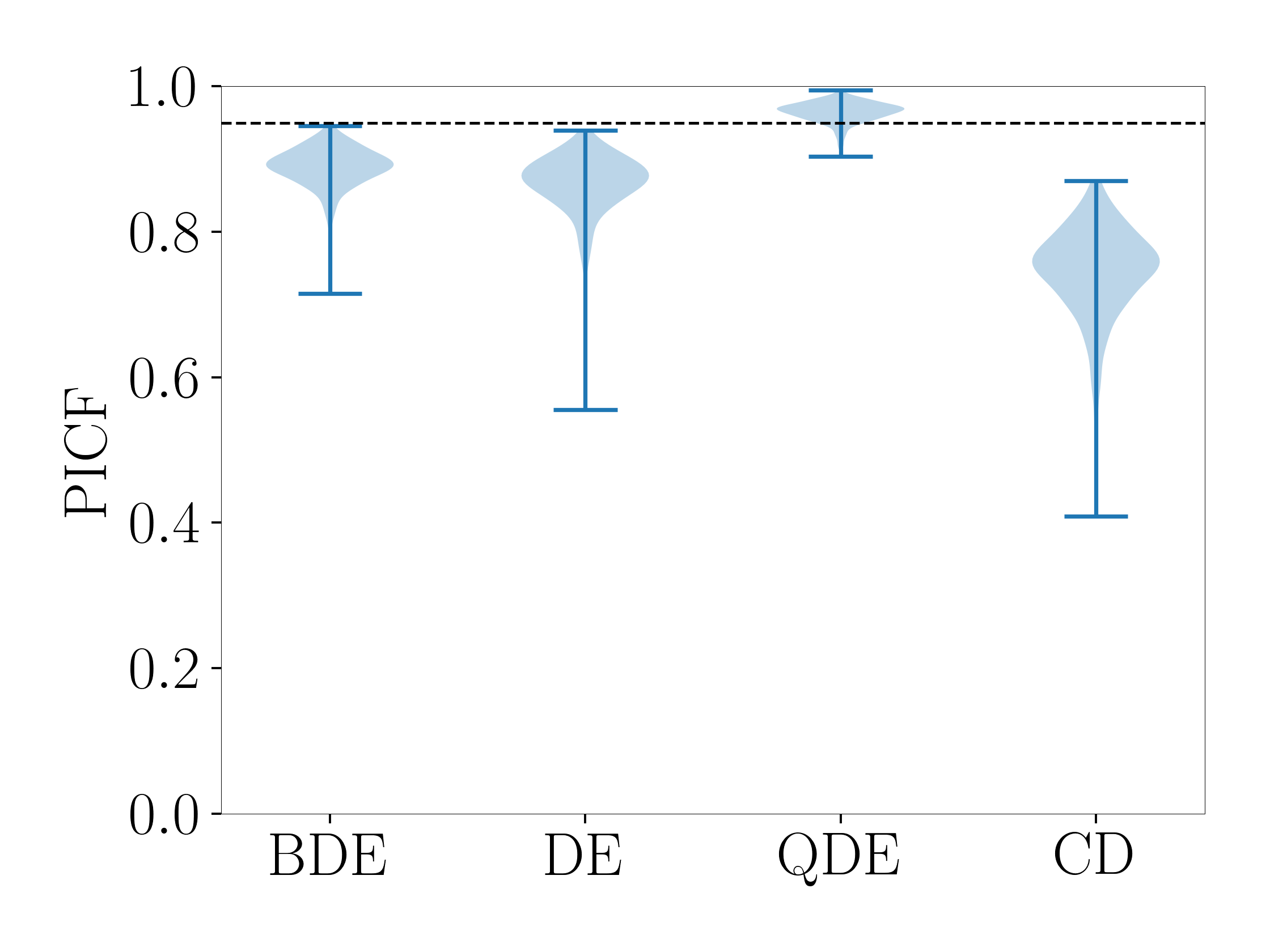}}
\subfigure[Wine, $\alpha=0.2$]{\includegraphics[width=.24\textwidth]{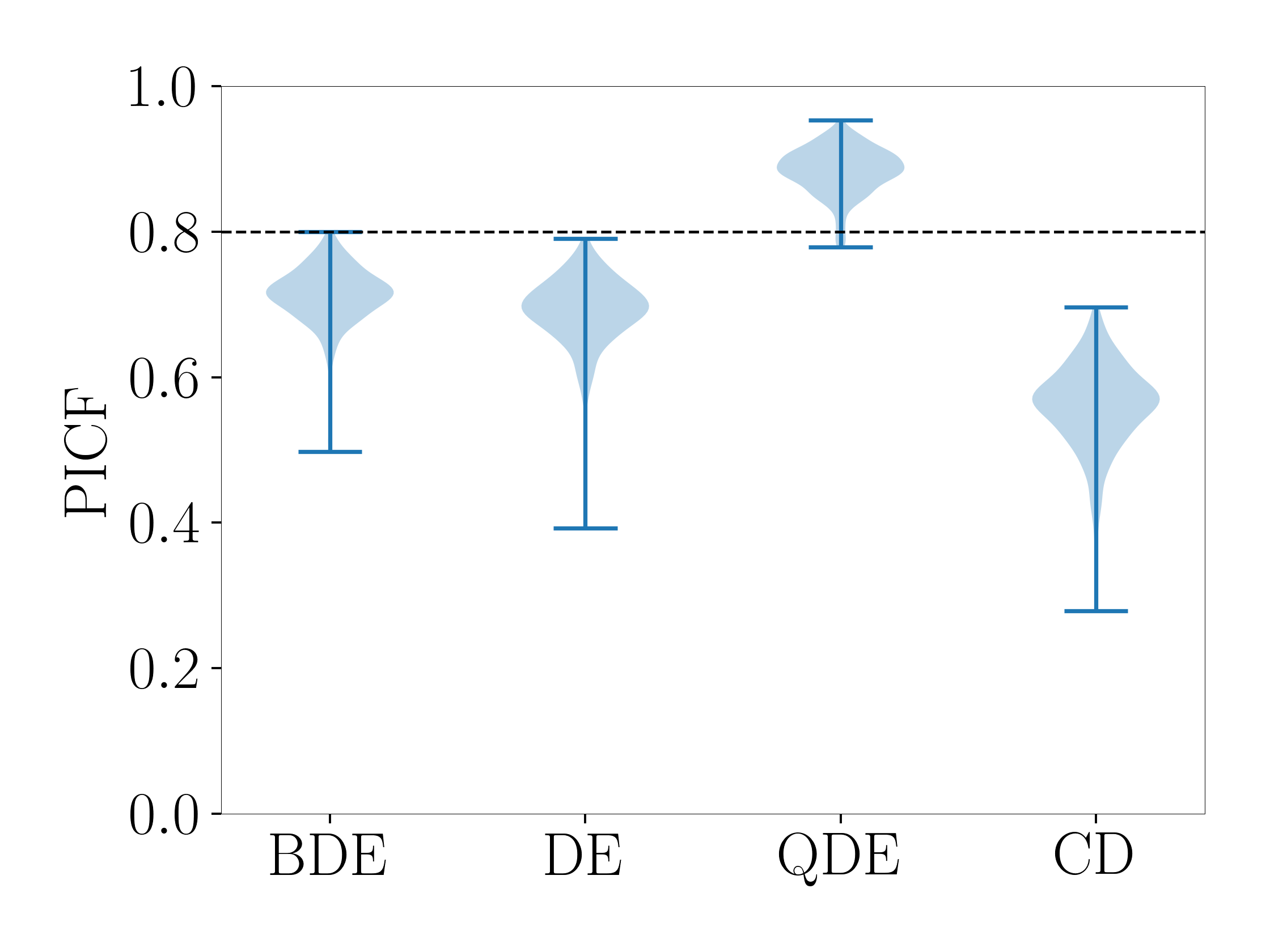}}
\subfigure[Yacht, $\alpha=0.05$]{\includegraphics[width=.24\textwidth]{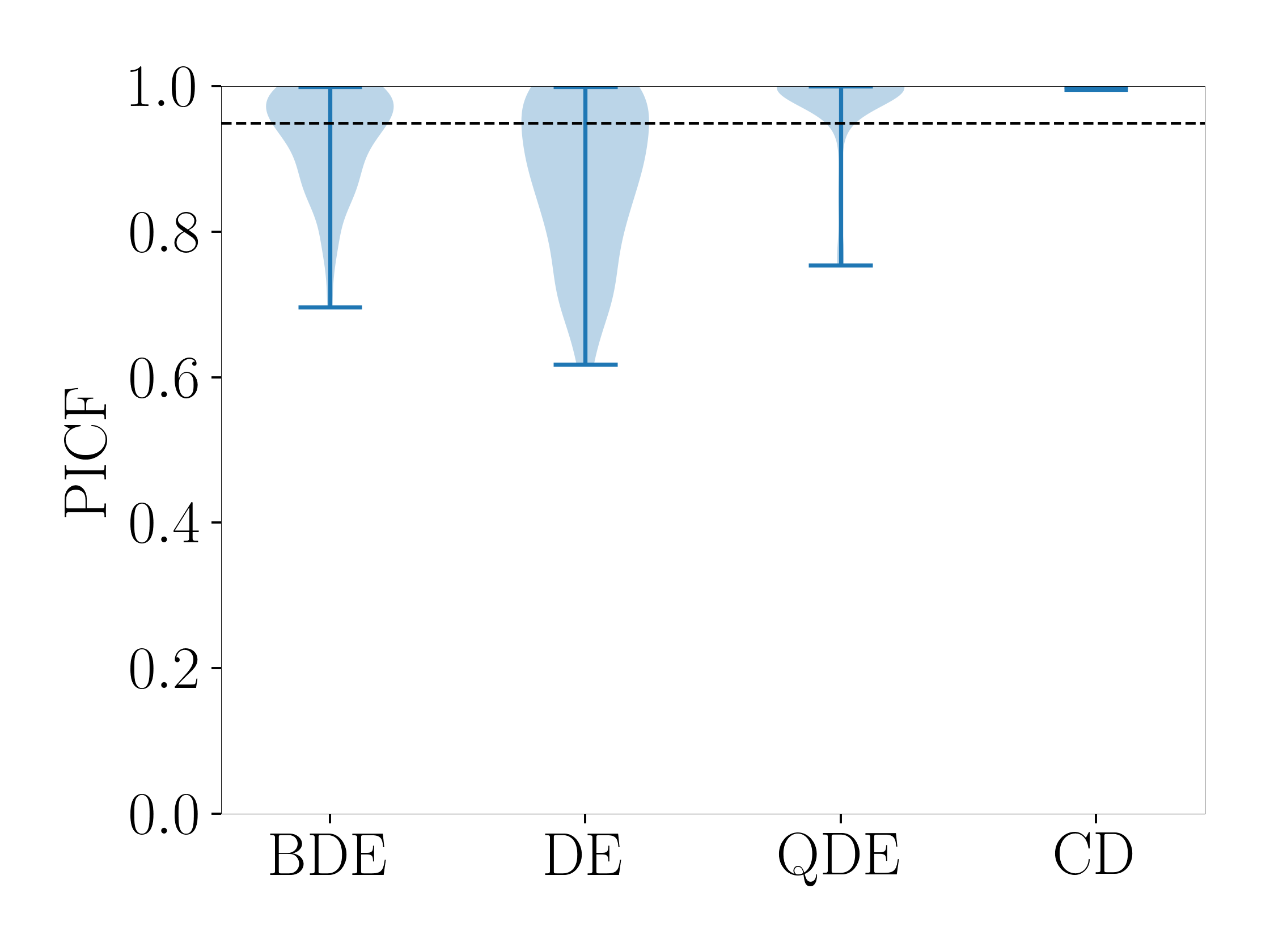}}
\subfigure[Yacht, $\alpha=0.2$]{\includegraphics[width=.24\textwidth]{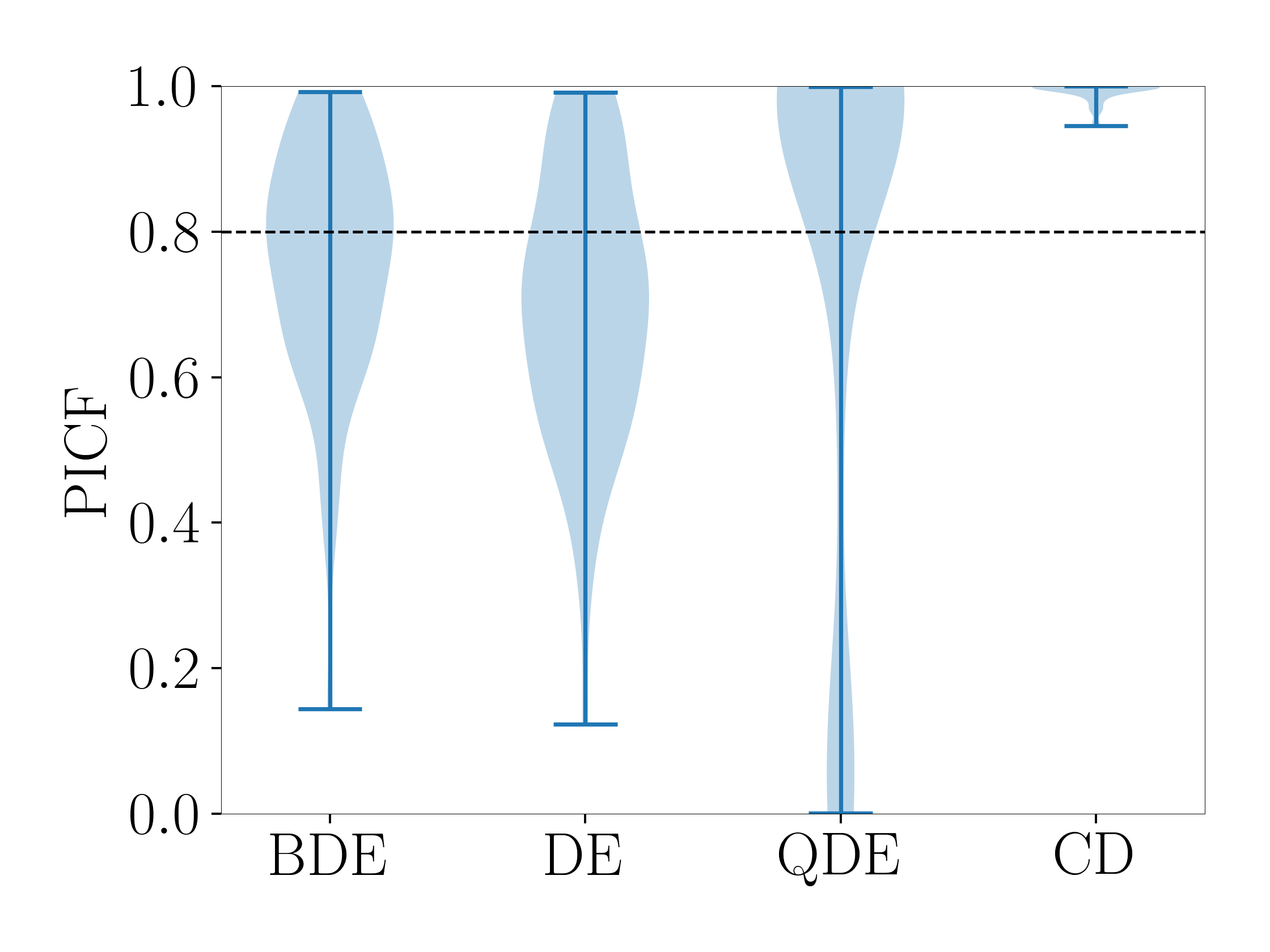}}
\caption{Violin plots of the PICF values for all 8 simulations of experiment 1 in the main text. For each simulation we give the PICF values for the $95\%$ and $80\%$ confidence intervals. The prediction intervals have better coverage than the confidence intervals. Quality-Driven Ensembles give prediction intervals that are too large in most simulations.}
\label{fig: violinsPICF}
\end{figure}

\noindent Figures \ref{fig: violinsCICF} and \ref{fig: violinsPICF} give the violin plots of the CICF and PICF values from Experiment 1 in the main text. These figures demonstrate that bootstrapped DE are able to provide reliable confidence and prediction intervals. We also note that high-quality prediction intervals do not guarantee high quality confidence intervals.

\FloatBarrier

\section{Detecting overfitting} \label{overfittingappendix}
\noindent While not the primary goal of our method, a potential added bonus is the possibility to detect overfitting. Suppose that we are in a situation where the networks are overfitting the noise. The ensemble members of standard DE, that are trained on the exact same targets, will tend to provide the same predictions when overfitting on the targets, yielding very small confidence intervals. The retrained ensemble members of Bootstrapped DE, on the other hand, are trained on different targets and hence will tend to provide quite different predictions from their original counterparts, leading to relatively large confidence intervals. With extreme overfitting, to the point that $\hat{\sigma}^2(\bm{x})$ gets close to zero, this advantage of bootstrapped DE over DE will vanish and both methods will fail to detect overfitting. 

We provide short a motivating example by comparing BDE and DE in a scenario in which we know that the network will overfit: a complex network with only 7 data points and no regularisation. The targets were simulated from a $\N{0}{0.2^{2}}$ distribution, pure noise. Each ensemble member had three hidden layers containing 400, 200, and 100 hidden layers and was trained for 80 epochs. 

\begin{figure}[h!]
\centering
\includegraphics[width=0.6\textwidth]{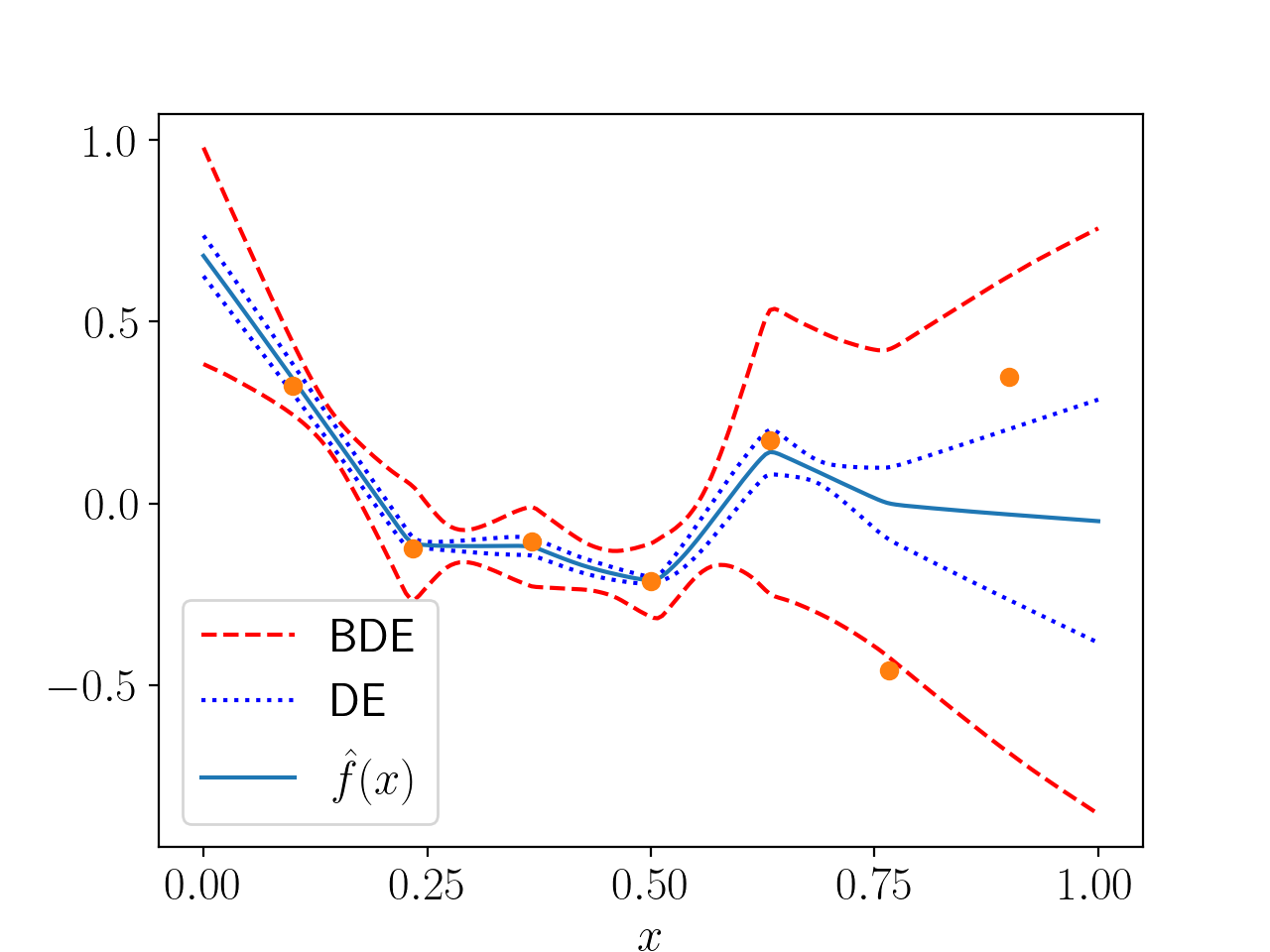}
\caption{The 90$\%$ confidence intervals of Bootstrapped Deep Ensembles (BDE) and Deep Ensembles (DE). The original ensemble members were trained long enough to overfit on the data. DE are unable to detect this since all ensemble members behave roughly the same. The variance of the networks after retraining on new targets, however, is much larger since each network will overfit on \textit{different} targets. BDE are therefore better able to detect overfitting.}
\label{fig: overfitting}
\end{figure}

Figure \ref{fig: overfitting} illustrates that BDE is better able to detect overfitting. The confidence intervals of our method increase at the location of the data points, indicating overfitting, whereas those of DE almost vanish. The reason is as follows. The original ensemble members $\hat{f}_{i}(\bm{x})$ are the same for both BDE and DE. In the retraining step of BDE, however, the ensemble members will overfit to new targets, resulting in a large estimate for $\sigmad(\bm{x})$. This overfitting is even more apparent by the fact that confidence intervals of BDE sometimes actually increase at the locations of the training data. We only investigated this briefly and it may be worthwhile to investigate this further.

\end{document}